\documentclass[journal]{IEEEtran}

\usepackage{multirow}
\usepackage{makebox}
\usepackage{mathrsfs}
\usepackage{bm}
\usepackage{tabularx}
\usepackage{color}
\usepackage{fmtcount}

\usepackage{cite}

\ifCLASSINFOpdf
   \usepackage[pdftex]{graphicx}
\else
   \usepackage[dvips]{graphicx}
\fi
\usepackage{epstopdf}
\usepackage[cmex10]{amsmath}

\usepackage{array}

\ifCLASSOPTIONcompsoc
  \usepackage[caption=false,font=normalsize,labelfont=sf,textfont=sf]{subfig}
\else
  \usepackage[caption=false,font=footnotesize]{subfig}
\fi

\usepackage{fixltx2e}

\def\mc{\mathcal}
\def\mb{\mathbf}

\ifCLASSOPTIONcaptionsoff
  \usepackage[nomarkers]{endfloat}
 \let\MYoriglatexcaption\caption
 \renewcommand{\caption}[2][\relax]{\MYoriglatexcaption[#2]{#2}}
\fi

\usepackage{url}

\begin{document}
\title{An Experimental Survey on Correlation Filter-based Tracking}

\author{Zhe~Chen,
        Zhibin~Hong,
        and~Dacheng~Tao,~\IEEEmembership{~Fellow,~IEEE}}% 

%\markboth{Journal of \LaTeX\ Class Files,~Vol.~13, No.~9, September~2014}%
%{Shell \MakeLowercase{\textit{et al.}}: An Experimental Survey on Correlation Filter-based Tracking Algorithms}

\maketitle

\begin{abstract}
Over these years, Correlation Filter-based Trackers (CFTs) have aroused increasing interests in the field of visual object tracking, and have achieved extremely compelling results in different competitions and benchmarks. In this paper, our goal is to review the developments of CFTs with extensive experimental results. 11 trackers are surveyed in our work, based on which a general framework is summarized. Furthermore, we investigate different training schemes for correlation filters, and also discuss various effective improvements that have been made recently. Comprehensive experiments have been conducted to evaluate the effectiveness and efficiency of the surveyed CFTs, and comparisons have been made with other competing trackers. The experimental results have shown that state-of-art performence,  in terms of robustness, speed and accuracy, can be achieved by several recent CFTs, such as MUSTer and SAMF. We find that further improvements for correlation filter-based tracking can be made on estimating scales, applying part-based tracking strategy and cooperating with long-term tracking methods.

%Over these years, numerous Correlation Filter-based Trackers (CFTs) have been proposed with extremely compelling performance.  in different competitions and benchmarks. In this paper, our goal is to review prevailing developments of CFTs and to provide a comprehensive survey with extensive experimental results. In specific, a general framework of CFTs is summarized first, and then several important improvements, which are based on training schemes, feature extraction methods, scale handling, part-based appearance models and the cooperation with long-term tracking, are discussed in details. In the conducted experiments, we fairly evaluated  ten representative CFTs and carefully compared them with other 29 popular trackers on a well-known large-scale benchmark. In the end, conclusions and some future research directions on CFTs are properly discussed.
\end{abstract}


\IEEEpeerreviewmaketitle

\section{Introduction}
\IEEEPARstart{V}{isual} object tracking is one of the most challenging tasks in the field of computer vision and is related to a wide range of applications like surveillance and robotics. Given the initial state of a target in the first frame, the goal of tracking is to predict states of the target in a video. However, designing a fast and robust tracker is difficult according to various critical issues in visual tracking, such as illumination variations, occlusions, deformations, rotations and so on. Over the past decade, various tracking algorithms have been proposed to cope with these challenges, some of which use generative models \cite{kwon2011tracking, sevilla2012distribution, zhang2012robust, liu2012visual, belagiannis2012segmentation, kwak2011learning} while the others use discriminative models \cite{hare2011struck, lucey2008enforcing, wang2010discriminative, kalal2010pn, zhang2012real, wang2015object}. Generative trackers perform tracking by searching the best-matching windows, and discriminative methods learn to distinguish the target from backgrounds. In \cite{wu2013online, wu2015object}, it has been found that background information is advantageous for effective tracking, which suggests that discriminative methods are more competing. In particular, the correlation filter-based discriminative trackers have made significant achievements recently, and have been paid more attention by corresponding  researchers. Therefore, summarizing  the developments of correlation filter-based tracking algorithms and comparing them with other popular trackers are supposed to be conducive for future researches.

Conventionally, correlation filters are designed to produce correlation peaks for each interested target in the scene while yielding low responses to background, which are usually used as detectors of expected patterns. Although localization tasks can be effectively performed by these filters, the required training needs used to make them inappropriate for online tracking. Only after the proposal of Minimum Output Sum of Squared Error (MOSSE) \cite{bolme2010visual} filter, this situation has been changed. Using an adaptive training scheme, MOSSE is considerably robust and efficient in tracking. Based on the basic framework of MOSSE filter, numerous improvements have been made later. For example, Henriques \textit{et al.} \cite{henriques2012exploiting} improved the MOSSE filter by introducing kernel methods, and Danelljan \textit{et al.} \cite{danelljan2014adaptive} applied color-attributes to better represent the input data. By further handling the scale changes, three Correlation Filter-based Trackers (CFTs), namely SAMF \cite{li2014scale}, DSST \cite{danelljan2014accurate} and an improved KCF \cite{henriques2014high}, have achieved state-of-art results and have beaten all other attended trackers in terms of accuracy in a recent competition\cite{lirisvisual}. With more CFTs developed recently \cite{hong2015multi, ma2015long, liu2015real, li2015reliable}, correlation filter-based tracking has proven its great strengths in efficiency and robustness, and has considerably accelerated the development of visual object tracking

Despite the various correlation filter-based tracking algorithms proposed these years, there is no work to review them with comprehensive evaluations. To facilitate other researchers for future contributions, our fundamental goals of this paper include: 1) formulating a general framework; 2) investigating the major developments of CFTs; 3) carrying out comprehensive evaluations on a large scale benchmark; 4) making appropriate comparisons; and 5) illustrating future research directions.

\begin{table*}[!h]
    \begin{center}
       \caption{Major Surveyed Papers}
	  \label{tab:sum}
	  \addtocounter{footnote}{1}
 		\begin{tabular}{|c|c|p{10cm}|}
		  \hline
		  Name & Published Year & Major Contribution\\
		  \hline
		  MOSSE and Regularized ASEF \cite{bolme2010visual} & 2010 & Pioneering work of introducing correlation filters for visual tracking \\
		  \hline
		  CSK \cite{henriques2012exploiting} & 2012 &  Introduced Ridge Regression problem with circulant matrix to apply kernel methods \\
		  \hline
		  STC \cite{zhang2014fast} &2014& Introduced spatio-temporal context information \\
		  \hline
		  KCF \cite{henriques2014high} &2014& Formulated the work of CSK and introduced multi-channel HOG feature.  \\
		  \hline
		  CN \cite{danelljan2014adaptive} &2014&Introduced color attributes as effective features\\
		  \hline
		  DSST \cite{danelljan2014accurate} &2014&Relieved the scaling issue using feature pyramid and 3-dimensional correlation filter\\
		  \hline
		  SAMF \cite{li2014scale} &2014& Integrated both color feature and HOG feature; Applied a scaling pool to handle scale variations\\
		  \hline
		  RPAC \cite{liu2015real}\footnotemark[1] &2015&Introduced part-based tracking strategy\\
		  \hline
		  RPT \cite{li2015reliable} &2015&Introduced reliable local patches to facilitate tracking\\
		  \hline
		  LCT \cite{ma2015long} &2015&Introduced online random fern classifier as re-detection component for long-term tracking\\
		  \hline
		  MUSTer \cite{hong2015multi} &2015& Proposed a biology-inspired  framework where short-term processing and long-term processing are cooperated with each other\\
		  \hline
		  \multicolumn{3}{l}{$^1$This abbreviation is taken from its title: \textit{Real-time Part-based visual tracking via Adaptive Correlation filters}.}
		  \end{tabular}
    \end{center}
\end{table*}

In this work, various important CFTs are surveyed and their contributions are discussed in detail. Brief introductions of these studies can be found in Table \ref{tab:sum}. In general, training schemes of filters are extremely crucial in correlation filter-based tracking, and CFTs can be further improved by introducing better training schemes, extracting powerful features, relieving scaling issue, applying part-based tracking strategy and cooperating with long-term tracking. To evaluate the tracking performance, we have collected source codes of 8 CFTs from the internet, and implemented two CFTs in a simple version. By running on a large scale benchmark \cite{wu2013online, wu2015object}, the performance of tested CFTs is compared with other popular competing trackers. The obtained experimental results are presented and analyzed, proving the efficiency and robustness of correlation filter-based tracking methods. According to the conducted experiments, latest CFTs are demonstrated to be state-of-art trackers. 

The rest of this paper is arranged as follows. In Section \ref{sec:overview}, we provide an overview of the basic framework of correlation filter-based tracking methods. Afterwards, theories and schemes for training correlation filters are introduced in Section \ref{sec:train}. For Section \ref{sec:improve}, numerous aspects of further improvements are reviewed and discussed in detail.  Afterwards, experimental results are presented and analyzed in Section \ref{sec:exp}. In the end, conclusions and future trends are summarized in Section \ref{sec:conclusion}. 

\section{Correlation Filter-based Tracking Framework}
\label{sec:overview}
\begin{figure*}[!t]
  \begin{center}
  \footnotesize
  \includegraphics[width=0.95\linewidth]{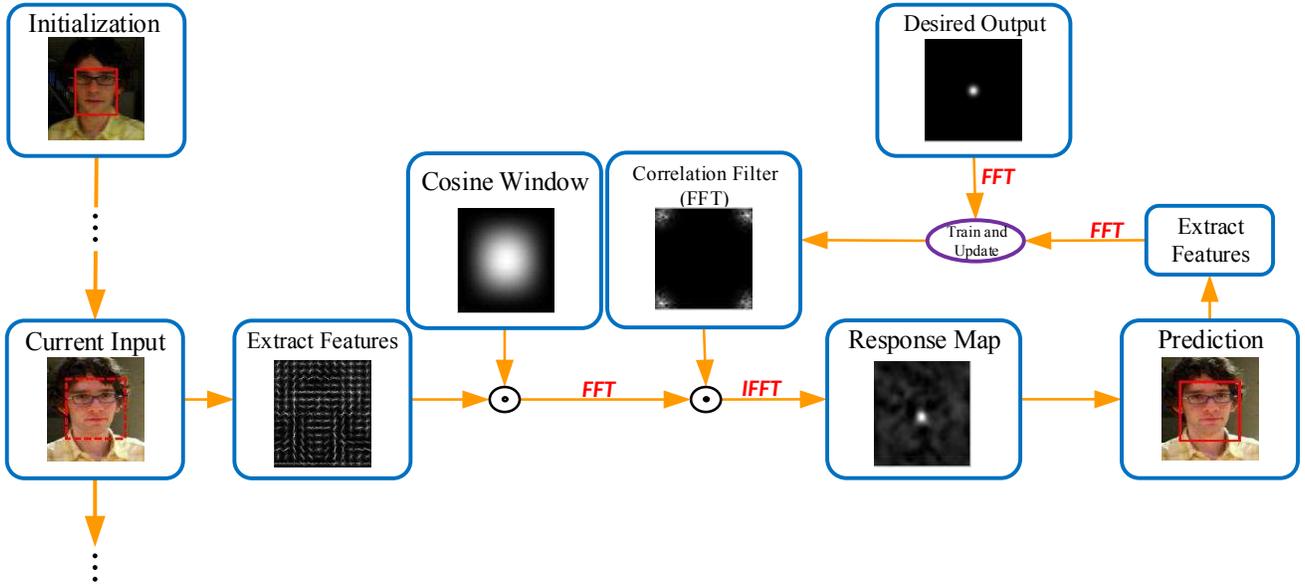}
  \end{center}
  \caption{General workflow for typical correlation filter-based tracking methods. At each frame after initialization, an image patch at previous estimated position is cropped as current input. Subsequently, visual features can be extracted for better describe the input, and a cosine window is usually applied for smoothing the discontinuities at window boundaries. Afterwards, correlation between current input and the learned filter is performed in frequency domain based on Convolution Theorem. The symbol $\odot$ in the figure denotes element-wise computation, and \textit{FFT} means Fast Fourier Transform. After the correlation, a spatial confidence map is obtained by Inverse FFT (\textit{IFFT}), whose peak can be predicted as the new position of target. Lastly, appearance at the newly estimated position is extracted for training and updating the correlation filter with a desired output. }
\label{fig:flow}
\end{figure*}

According to the existing correlation filter-based tracking methods, the general working framework can be summarized as follows. Initially, correlation filter is trained with image patch cropped from a given position of the target at first frame. Then in each subsequent time step, the patch at previous predicted position is cropped for detection. Afterwards, as shown in Figure \ref{fig:flow}, various features can be extracted from the raw input data, and a cosine window is usually applied for smoothing the boundary effects. Subsequently, efficient correlation operations are performed by replacing the exhausted convolutions with element-wise multiplications using Discrete Fourier Transform (DFT). In practice, the DFT of a vector is computed by the efficient Fast Fourier Transform (FFT) algorithm. Following the correlation procedure, a spatial confidence map, or response map, can be obtained using inverse FFT. The position with a maximum value in this map is then predicted as the new state of target. Next, appearance at the estimated position is extracted for training and updating the correlation filter. Because only the DFT of correlation filter is required for detection, training and updating procedures are all performed in frequency domain.

To describe the workflow mathematically, let $\mb{x}$ be the input of detection stage and $\mb{h}$ be the correlation filter. In practice, $\mb{x}$ can be either raw image patch or extracted features. Suppose the symbol $\hat{}$ represents the Fourier transform of a vector. According to Convolution Theorem, circulant convolution equals to element-wise multiplication in frequency domain
\begin{equation}
    \mb{x}\otimes \mb{h} = \mc{F}^{-1}\left(\hat{\mb{x}} \odot \hat{\mb{h}}^*\right)
\label{eq:1}
\end{equation}
where $\mc{F}^{-1}$ is inverse Fourier transform operation, $\odot$ denotes element-wise multiplication and $*$ means the complex conjugate. The results of (\ref{eq:1}) are the expected correlation output between $\mb{x}$ and $\mb{h}$, which also form the mentioned confidence map.

For training the filter, let us first define a desired correlation output $\mb{y}$. Using the new instance $\mb{x}'$ of target, correlation filter $\mb{h}$ should satisfy:
\begin{equation}
    \mb{y} =\mathcal{F}^{-1}\left(\hat{\mb{x}}' \odot \hat{\mb{h}}^*\right)
\label{eq:2}
\end{equation}
and thus:
\begin{equation}
    \hat{\mb{h}}^* =\frac{\hat{\mb{y}}}{\hat{\mb{x}}'}
\label{eq:3}
\end{equation}
where $\hat{\mb{y}}$ is the DFT of $\mb{y}$ and the division is computed element-wise.

In terms of computation cost, the complexity of circular convolution for an image of size $n\times n$ is $\mc{O}(n^4)$ , while the element-wise multiplications using FFT only require $\mc{O}(n^2\log n)$. Therefore the acceleration brought by FFT is significant.  

However, there are some issues that should be handled well when using the correlation filter-based tracking framework. First, training schemes are extremely crucial for CFTs. Since the target may change its appearance continuously, correlation filters should be adaptively trained and updated on-the-fly to adapt to the new appearance of target. Second, feature representing methods also greatly influence the performance. Although raw pixels can be directly used for detection, the tracker may be affected by various noises like illumination changes and motion blurs. More powerful features are supposed to be helpful. Moreover, how to adapt to the scales of target is another challenging problem for CFTs. Since the sizes of correlation filters are usually fixed in tracking, scale variations of the target cannot be handled well in these trackers. As a result, an effective scale estimation approach is supposed to complement this shortage of correlation filter-based tracking. Furthermore, long-term tracking is believed to be the weakness of many CFTs since they commonly lack the ability to re-locate the target after drifting. By cooperating with long-term tracking methods, CFTs can be much more robust in tracking.

\section{Training Schemes for correlation filters}
\label{sec:train}
The behaviors of correlation filters can be diversified if different methods are used for training.To train a robust correlation filter for online visual tracking, numerous studies have been proposed.

\subsection{Traditional Training Methods} 

For the simplest case, template cropped from an image can be used to produce peaks for the target. However, their responses to background patterns are also relatively high. To overcome this issue, a variety of correlation filters \cite{hester1980multivariant, refregier1991optimal, mahalanobis1987minimum, savvides2003efficient, mahalanobis1994unconstrained} were trained by suppressing responses to negative training samples while maintaining high response to the target. The main difference among these filters is the methods they are constructed with the collected training samples. For example, Synthetic Discriminant Functions (SDF) \cite{casasent1984unified, hester1980multivariant}, Optimal Tradeoff Filters (OTF) \cite{refregier1991optimal} and Minimum Average Correlation Energy (MACE) \cite{mahalanobis1987minimum} are trained with enforced hard constraints so that peaks would always be produced in the same height. On the contrary, hard constraints are believed to be unnecessary in other filters, such as Maximum Average Correlation Height (MACH) \cite{savvides2003efficient} and Unconstrained MACE (UMACE) \cite{mahalanobis1994unconstrained}. These filters are trained by relaxing the hard constraints. More details about developments of correlation filters can be found in the survey \cite{kumar2014recent}. Recently, a correlation filter, which is named as Average of Synthetic Exact Filters (ASEF) \cite{bolme2009average}, averages all the trained exact filters to obtain a general one. The resulted filter has shown to perform well in eye localization \cite{bolme2009average} and pedestrian detection\cite{bolme2009simple}.  Although ASEF may be robust enough to be applied in visual tracking, a large number of samples are required for training, which makes it too slow for online tasks. 

\subsection{Adaptive Correlation Filters}
To train correlation filters more efficiently, a novel filter termed as Minimum Output Sum of Squared Error (MOSSE) was developed by Bolme \textit{et al.} \cite{bolme2010visual}, together with an improved version of ASEF.

\subsubsection{MOSSE}
According to (\ref{eq:2}) and (\ref{eq:3}), a simple filter can be obtained on sample $\mb{x}$ with the corresponding desired output $\mb{y}$. However, more samples are needed to improve the robustness of correlation filters. To properly map these input samples to desired outputs, MOSSE finds a filter $\mb{h}$ by minimizing the sum of squared error between actual correlation outputs and desired correlation outputs. By computing in frequency domain, this minimization problem can be expressed by:
\begin{equation}
    \min_{\hat{\mb{h}}^*}\sum_{i} {\|\hat{\mb{x}}_i \odot \hat{\mb{h}}^*- \hat{\mb{y}}_i\|}^2
\label{eq:4}
\end{equation}
where $i$ indexes each training image. Then the solution of $\hat{\mb{h}}^*$ is given by:
\begin{equation}
    \hat{\mb{h}}^* = \frac{  \sum_{i}{ \hat{\mb{y}}_i \odot \hat{\mb{x}}_i^*} }
{\sum_{i}{\hat{\mb{x}}_i \odot \hat{\mb{x}}_i^*}}
\label{eq:5}
\end{equation}
whose detailed derivations can be found in \cite{bolme2010visual}. 

In general, the desired output $\mb{y}$ can take any shape. In MOSSE, it is generated from ground truth with a compact 2D Gaussian shaped distribution whose peak is at the center. If Kronecker delta function is used for defining $\mb{y}$, whose value at target center is one and values elsewhere are zero, the resulted filter is theoretically a UMACE filter mentioned above. Thus UMACE is a special case of MOSSE.

\subsubsection{Regularized ASEF}
By slightly modifying the original form, ASEF is also capable of efficient tracking. Using one sample at a time, a filter called exact filter can be found by solving (\ref{eq:4}):
\begin{equation}
    \hat{\mb{h}}_i^* = \frac{  \hat{\mb{y}}_i \odot \hat{\mb{x}}_i^* }
{\hat{\mb{x}}_i \odot \hat{\mb{x}}_i^*}
\label{eq:6}
\end{equation}

Then a more general filter can be produced by averaging all the computed exact filters:
\begin{equation}
    \hat{\mb{h}}_i^* = \frac{1}{N} \sum_i{  \frac{\hat{\mb{y}}_i \odot \hat{\mb{x}}_i^* }
{\hat{\mb{x}}_i \odot \hat{\mb{x}}_i^*} }
\label{eq:7}
\end{equation}

However, original ASEF can be much unstable because the denominator in (\ref{eq:7}) may be extremely small. To help produce a more stable filter, a regularization parameter $\epsilon$ can be introduced in the denominator to prevent it from being a close-to-zero number, which has shown to be effective for stabilization.

\subsection{Kernelized Correlation Filters}
\label{sec:KCF}
After the success of \cite{bolme2009average, bolme2010visual}, correlation filter-based tracking framework has shown to be significantly efficient for robust tracking. However, the overall performance may be limited because the ASEF and MOSSE filters can be viewed as simple linear classifiers. By taking advantage of kernel trick, correlation filters are supposed to be more powerful.

There have already been some studies \cite{patnaik2009fast, casasent2007analysis, jeong2006kernel} to apply kernel methods in correlation filters. According to \cite{patnaik2009fast, casasent2007analysis}, it has been found that filters which do not use the power spectrum or image translations are easier to be kernelized. Different from these studies, Henriques \textit{et al.} \cite{henriques2012exploiting, henriques2014high} proposed that correlation filters can be effectively kernelized with the introduction of Ridge Regression problem and circulant matrix. 

 \subsubsection{Ridge Regression Problem}
By considering correlation filters as classifiers, they can be trained by finding the relation between $i$-th input $\mb{x}_i$ and its label $y_i$ from a training set. Suppose the relation takes the form $f(\mb{x}_i) = y_i$, training problem can be viewed as minimizing the objective function:
\begin{equation}
    \min_{\mathbf{w}} \sum_{i} L\left(f\left(\mathbf{w},\mathbf{x}_i\right),y_i\right) + \lambda \|\mathbf{w}\|^2
\label{eq:8}
\end{equation}
where $\mathbf{w}$ denotes the parameters, $\lambda$ is regularization parameter to prevent overfitting, and $L(\cdot)$ is loss function. In SVM, $L(\cdot)$ is defined by hinge loss $L(f(\mathbf{w},\mathbf{x}_i),y_i )=\max ⁡(0,1-y_i f(\mathbf{w},\mathbf{x}_i))$, while Regularized Least Squares (RLS) which uses quadratic loss $L(f(\mathbf{w},\mathbf{x}_i),y_i )=(y_i-f(\mathbf{w},\mathbf{x}_i))^2$ can be alternatively applied for training filters.  It has been shown that training by RLS can deliver equivalent performance with hinge loss \cite{rifkin2003regularized}. The RLS is also known as Ridge Regression.

For the function $f(\mb{x}_i)$, it can be a linear operation $f(\mathbf{x}_i)= \langle \mathbf{w},\mathbf{x}_i \rangle+b$ where $\langle \cdot,\cdot  \rangle$ is dot product and $b$ is constant offset. By solving (\ref{eq:8}), the parameter $\mb{w}$ can be given in a closed form  \cite{rifkin2003regularized}:
\begin{equation}
	\mathbf{w}=\left(X^T X+\lambda I\right)^{-1} X^T \mb{y}		
\label{eq:9}
\end{equation}
where $X$ is a matrix whose rows are training samples, $\mb{y}$ is a vector of corresponding labels, and $I$ is identity matrix. It is worth noting that if the computation is performed in frequency domain, $X^T$ should be replaced by the Hermitian transpose of $X$ in (\ref{eq:9}), which is $X^H = (X^*)^T$.

 To introduce the kernel functions for improving performance, input data $\mb{x}$ can be mapped to a non-linear-feature space with $\varphi(\mathbf{x})$, and $\mb{w}$ can be expressed by linear combination of the inputs $\mathbf{w}=\sum_{i} \alpha_i \varphi(\mathbf{x}_i)$. Then $f(\mb{x}_i)$ takes the form:
\begin{equation}
  f(\mb{x}_i) = \sum_{j=1}^n { \alpha_i \kappa(\mb{x}_i, \mb{x}_j)}
  \label{eq:10}
\end{equation}
where $\kappa(\mb{x}_i, \mb{x}_j) = \langle \varphi (\mathbf{x}_i),\varphi (\mathbf{x}_j) \rangle$ is the kernel function. Suppose $K$ is the kernel matrix with its elements $K_{ij} = \kappa(\mb{x}_i,\mb{x}_j)$. The solution of (\ref{eq:8}) using kernel functions can be given by \cite{rifkin2003regularized}:
\begin{equation}
	\bm{\alpha} =\left(K+\lambda I\right)^{-1}\mb{y}	
\label{eq:11}
\end{equation}
 where $I$ is identity matrix. To avoid difficulty in computing inverse matrix of (\ref{eq:11}), circulant matrix can be introduced.
 
\subsubsection{Circulant Matrix}
Generally, samples are obtained by random sampling \cite{hare2011struck, babenko2011robust, grabner2008semi, avidan2004support, saffari2009line}. With the help of circulant matrix, however, all the translated samples around the target can be collected for training without sacrificing much speed. 

With a base sample $\mb{x} = (x_0,\ldots,x_{n-1})$, a circulant matrix $X$ has the following form:
\begin{equation}
	X = C(\mathbf{x}) = \left( \begin{array}{cccc}
			x_0 & x_1 & \ldots & x_{n-1} \\
			x_{n-1} & x_0 & \ldots & x_{n-2} \\
			\vdots & \vdots & \ddots & \vdots \\
			x_1 & x_2 & \ldots & x_0  \end{array} \right)
\label{eq:12}
\end{equation}
 
There are various interesting properties of circulant matrices. For example, their sums, products and inverses are also circulant. In addition, a circulant matrix can be made diagonal with the DFT of its base vector $\mb{x}$ \cite{gray2006toeplitz}:
\begin{equation}
    X = F \mbox{diag}(\mathbf{\hat{x}})F^H
\label{eq:13}
\end{equation}
where $F$ is DFT matrix, which is used for computing the DFT of an vector $\mathcal{F}(\mb{z}) =\sqrt{n} F\mb{z}$. Then the solution of $\mb{w}$ can be expressed in the form:
\begin{equation}
	\mb{w} =F\mbox{diag}\left(\frac{\mathbf{\hat{x}}}{\mathbf{\hat{x}}^* \odot \mathbf{\hat{x}}+\lambda}\right)F^H\mb{y}
\label{eq:14}
\end{equation}
which is equivalent to a simpler form in frequency domain:
\begin{equation}
	\hat{\mb{w}} = \frac{\hat{\mb{x}}^* \odot \hat{\mb{y}}}{\hat{\mathbf{x}}^* \odot \hat{\mathbf{x}}+\lambda}
\label{eq:15}
\end{equation}
where the division is performed element-wise . Similarly, $\bm{\alpha}$ can also be computed efficiently if the kernel matrix $K$ is circulant:
\begin{eqnarray}
   \bm{\alpha} = F\left(\mbox{diag}(\hat{\mathbf{k}} + \lambda)\right)^{-1} F^H  \bm{g}
\label{eq:16}
\end{eqnarray}
where $\mb{k}$ is the base vector of circulant matrix $K$, and further:
 \begin{equation}
	\hat{\bm{\alpha}} = \frac{\hat{\mb{y}}}{\hat{\mathbf{k}}+\lambda}
\label{eq:17}
\end{equation}
where the division is also element-wise. 

It has been proven that the kernel function of a circulant kernel matrix should be unitarily invariant (detailed proof can be found in \cite{henriques2012exploiting, henriques2014high}). Since dot-product and radial basis kernel functions are found to satisfy this condition, polynomial kernels and Gaussian kernels are usually applied. 

If the kernel $\mb{k}$ is computed between $\mb{x}$ and $\mb{x}'$, a polynomial kernel $\mb{k}^{\mb{xx'}} = (\mathbf{x}^T\mathbf{x'}+a)^b$ can be expressed as:
\begin{equation}
	\mathbf{k}^{\mathbf{xx'}}  =  \left(\mathcal{F}^{-1}(\hat{\mathbf{x}}^* \odot \hat{\mathbf{x}}')+a\right)^b
\label{eq:18}
\end{equation}
and the Gaussian kernel $\mb{k}^{\mb{xx'}} =\exp\left(-\frac{1}{\sigma^2} (\|\mathbf{x}-\mathbf{x}' \|^2)\right)$ can be computed by:
\begin{equation}
	\mathbf{k}^{\mathbf{xx}'}  =  \exp\left(-\frac{1}{\sigma^2}\left(\|\mathbf{x}\|^2 + \|\mathbf{x}'\|^2 - 2\mathcal{F}^{-1}\left(\hat{\mathbf{x}}^*\odot \hat{\mathbf{x}}' \right)\right)\right)
\label{eq:19}
\end{equation}

All the derivations of equations (\ref{eq:8}) to (\ref{eq:19}) can be found in \cite{henriques2012exploiting, henriques2014high}. 

\subsubsection{Detection}
In a new frame, the target can be detected by the trained parameter $\bm{\alpha}$ and a maintained base sample $\mb{x}$. If the new sample is $\mb{z}$, a confidence map $\mb{y}$ can be obtained by:
\begin{equation}
	\mb{y} = C(\mathbf{k}^{\mathbf{xz}})\bm{\alpha} = \mathcal{F}^{-1}\left( \hat{\mathbf{k}}^{\mathbf{xz}} \odot \hat{\bm{\alpha}} \right)
\label{eq:20}
\end{equation}

Similar with ASEF and MOSSE, the position with a maximum value in $\mb{y}$ can be predicted as new position of the target. 

\subsection{Dense Spatio-Temporal Context Tracker}
\label{sec:STC}
Spatio-Temporal Context (STC) tracker proposed by \cite{zhang2014fast} was developed to exploit the use of context information. We consider it as another CFT since it follows a similar workflow described in Section \ref{sec:overview}.
 
Context information has already been considered in various trackers \cite{dinh2011context, grabner2010tracking, wen2012online, yang2009context}. In the majority of these studies, key points around the target are first extracted and then descriptors like SURF and SIFT are introduced to describe these consistent regions. However, crucial information can be ignored sometimes by these methods and they are also quite time-consuming. Therefore the fundamental goal of STC is to use context information more efficiently.

Instead of training by optimizing, STC is designed to learn a likelihood distribution, which is defined as the prior possibility of object locating in position $\mb{p}$ ($ \mb{p} \in \mb{R}^2$):
\begin{equation}
	\ell(\mb{p}) = P(\mb{p}|o)
\label{eq:21}
\end{equation}
where $\ell(\cdot)$ means likelihood and $o$ is the object present in the scene. 

Let $\mb{p}_o$ denote the position of targets center, and $\varOmega_c(\mb{p})$ denote the neighboring coordinates around $\mb{p}_o$. Then a context feature set can be defined by $P^c=\{\mathbf{c}(\mb{p}')=\left(I(\mb{p}'),\mb{p}')\right|\mb{p}'\in \varOmega_c (\mb{p}_o)\}$ where $I(\mb{p}')$ represents the image intensity at position $\mb{p}_o$. By marginalizing the likelihood distribution of $\mathbf{c}(\mb{p}')$ given $o$:
\begin{eqnarray}
    \ell(\mb{p}) &=& P(\mb{p}|o) \nonumber\\
    	&=& \sum_{\mathbf{c}(\mb{p}') \in P^c} P\left(\mb{p},\mathbf{c}(\mb{p}')|o\right) \nonumber \\
    	&=& \sum_{\mathbf{c}(\mb{p}') \in P^c} P\left(\mb{p}|\mathbf{c}(\mb{p}'),o\right)P\left(\mathbf{c}(\mb{p}')|o\right) 
\label{eq:30}
\end{eqnarray}
where $P\left(\mb{p}|\mathbf{c}(\mb{p}'),o\right)$ models the relationship between spatial context information and target location, and $P\left(\mathbf{c}(\mb{p}')|o\right)$ models the appearance of object. 

Since there is no direct expression of $P\left(\mb{p}|\mathbf{c}(\mb{p}'),o\right)$, let us define a function to describe it:
\begin{equation}
	P\left(\mb{p}|\mathbf{c}(\mb{p}'),o\right)=h(\mb{p}-\mathbf{p}')	
\label{eq:23}
\end{equation}
where $h$ can be some operations which take the difference of two vectors $\mb{p}$ and $\mb{p}'$ as its input. To relieve the ambiguities caused by similar objects in the neighborhood, $h$ should not be radially symmetric. In other words, $h(\mb{p}-\mb{p}')$ and $h(|\mb{p}-\mb{p}'|)$ should not equal to each other. 

For $P\left(\mathbf{c}(\mb{p}')|o\right)$, it can be defined as:
\begin{equation}
	P\left(\mathbf{c}(\mb{p}')|o\right) = I(\mb{p}')\omega_{\sigma}(\mb{p}'-\mb{p}_o)
\label{eq:24}
\end{equation}
where $I(\cdot)$ is image intensity and $\omega_{\sigma}(\cdot)$ denotes weighted Gaussian function defined by:
\begin{equation}
\omega_{\sigma}(\mb{p}'-\mb{p}_o) = a \exp\left(-\frac{1}{\sigma^2}\|\mb{p}'-\mb{p}_o\|^2\right)
\label{eq:25}
\end{equation}
where $a$ is a normalization parameter. Given this Gaussian distributed weights, contexts closer to the center of object are assigned with larger values while further contexts are assigned with smaller values. Therefore the tracker pays more attention on the central area. 

For training the $h(\mb{p}-\mb{p}')$, a desired output distribution $\ell(\mb{p})$ can be designed by hand. If the object is known to be at the center of scene, $\ell(\mb{p})$ can be defined by:
\begin{equation}
	\ell(\mb{p}) = P(\mb{p}|o) = b \exp\left(-\|\frac{\mb{p}-\mb{p}_o}{\alpha}\|^\beta\right)
\label{eq:26}
\end{equation}
where $b$ is also a normalization parameter, $\alpha$ is scale parameter and $\beta$ controls the shape of this distribution. Subsequently, we have:
\begin{eqnarray}
	\ell(\mb{p}) & = & b \exp\left(-\|\frac{\mb{p}-\mb{p}_o}{\alpha}\|^\beta\right) \nonumber \\
	& = & \sum_{\mathbf{c}(\mb{p}') \in P^c} h(\mb{p}-\mb{p}')I(\mb{p}')\omega_{\sigma}(\mb{p}'-\mb{p}_o) \nonumber\\
	& = & h(\mb{p}) \otimes \left(I(\mb{p})\omega_\sigma(\mb{p}-\mb{p}_o)\right)
\label{eq:27}
\end{eqnarray}

By introducing Convolution Theorem, we have:
\begin{equation}
    h(\mb{p}) = \mathcal{F}^{-1}\left(
    \frac{ 
    \mathcal{F}\left(b \exp\left(-\|\frac{\mb{p}-\mb{p}_o}{\alpha}\|^\beta\right)\right)
    } 
    {
    \mathcal{F}\left(I(\mb{p})\omega_{\sigma}(\mb{p}-\mb{p}_o)\right)
    }
    \right)
\label{eq:28}
\end{equation}
where division is performed element-wise. With a trained $h(\mb{p})$, $\ell(\mb{p})$ of the new frame can be calculated by:
\begin{equation}
    \ell(\mb{p}) = \mathcal{F}^{-1}\left(
    	\mathcal{F}\left(h(\mb{p})^{t-1}\right) \odot
	\mathcal{F}\left(I(\mb{p})^t \omega_\sigma(\mb{p}-\mb{p}_o^{t-1})\right)
    \right)
\label{eq:29}
\end{equation}
where $t$ represents current time step. 

Similarly, a position $\mb{p}$ with the maximum value in $\ell(\mb{p})$ can be viewed as the new position of the object.   

\subsection{Updating Scheme}
According to the introduced training schemes, each frame can produce a correlation filter, thus the strategy of combining it to existing trained filter is crucial for constructing a robust appearance model.

In CFTs, running average is usually applied for updating, though different algorithms may average over different components. For regularized ASEF, a general correlation filter is updated by averaging every learned exact filter:
\begin{equation}
	\hat{h}^*_t = \eta \frac{ \hat{\mb{y}_t} \odot \hat{\mb{x}}^*_t }{\hat{\mb{x}}_t \odot \hat{\mb{x}}^*_t + \epsilon} + (1-\eta)\hat{h}^*_{t-1}
\label{eq:30}
\end{equation}
where $t$ denotes the $t$-th frame and $\eta$ is learning rate. STC also updates its filter based on the form of (\ref{eq:30}).

Instead, MOSSE respectively averages the numerator and the denominator of (\ref{eq:9}):
\begin{equation}
\left\{
 \begin{array}{lll}
\hat{\mb{h}}^*_t &=& \frac{A_t}{B_t} \\
A_t &=& \eta(\hat{\mb{y}}_t \odot \hat{\mb{x}}_t^* ) + (1-\eta)A_{t-1} \\
B_t &=& \eta(\hat{\mb{x}}_t \odot \hat{\mb{x}}_t^*) + (1-\eta)B_{t-1} 
\end{array}
\right. 
\label{eq:31}
\end{equation}

For KCF, the dual space coefficients $\bm{\alpha}$ can be updated in frequency domain:
\begin{equation}
\hat{\bm{\alpha}}_t = \eta \frac{\hat{\mb{y}}}{\hat{\mathbf{k}}_t+\lambda} + (1-\eta)\hat{\bm{\alpha}}_{t-1} \\
\label{eq:32}
\end{equation}
whose $\hat{\mathbf{k}}_t$ is averaged by:
\begin{equation}
\hat{\mb{k}}_t = \eta \hat{\mb{k}}^{\mb{z}\mb{z}}+ (1-\eta)\hat{\mb{k}}_{t-1}
\label{eq:33}
\end{equation}
where $\mb{z}$ is the new sample extracted from currently predicted position.

Generally, CFTs use similar updating schemes described above, and sometimes slight modifications can be made to improve the performance. Given an example, Danelljan \textit{et al.} \cite{danelljan2014adaptive} modified the updating scheme of CSK tracker \cite{henriques2012exploiting} (original version of KCF) by using the cost function with weighted average quadratic error for training. Moreover, robust updating schemes can also be achieved by considering long-term tracking. If the target is lost or occluded, learning the appearance model is obviously harmful. To avoid learning the false positive samples, some studies have introduced long-term components with failure detection schemes \cite{hong2015multi, liu2015real, ma2015long, li2015reliable}. For instance, the tracker of \cite{liu2015real} stops updating if occlusions are detected, and the tracker of \cite{hong2015multi} refreshes the correlation filter if the prediction of long-term component is more confident. Experiments have shown that the detection of occlusions is extremely beneficial. 

\subsection{Comparisons of Different Training Schemes}
Training schemes discussed in this section include ASEF, MOSSE, Kernelized Correlation Filter (KCF) and STC. In general, they all follow the workflow described in Section \ref{sec:overview}, where computations based on Convolution Theorem are employed for detection and the filter is trained with a desired output. However, there are some differences among these CFTs.

For ASEF tracker, its filter is produced by averaging over all the learned filters, while MOSSE filter is trained by averaging over all the images. By introducing Ridge Regression problem and circulant matrices, kernelized correlation filters can be introduced for tracking. Theoretically, the linear kernel in KCF can be the same with MOSSE filter if multiple samples of single channel are used for training. If multiple channels are supported, the linear kernel, which is called Dual Correlation Filter (DCF) \cite{henriques2014high}, can be trained by only using a single sample. The general case which use several multi-channel samples to train filters requires expensive computation costs, it is inappropriate in online visual tracking.

The differences between STC and other introduced training schemes include the following aspects. First, STC is developed to model the relationships between the object and its local spatial contexts, while common CFTs model the input appearance with trained filters. Second, values of the confidence map in STC can be referred to as prior probabilities given the current object, while values in confidence maps of other CFTs are correlation scores. Third, the algorithm of STC has the ability of estimating scale variations, which is difficult for CFTs like MOSSE and KCF. More arguments can be found in \cite{zhang2014fast}.

\section{Further Improvements}
\label{sec:improve}
Instead of proposing a novel training scheme of correlation filter, there are various aspects that can improve the robustness of CFTs. Over the years, improvements have been mainly made on representing features, handling scale variations, applying part-based strategy and cooperating with long-term tracking.
 
\subsection{Feature Representation}
\label{sec:feature}
In earlier CFTs like MOSSE and CSK, raw pixels are directly used for tracking. However, noises brought by raw images extremely limit the tracking performance. Although shifting the input data to a zero-mean distribution or multiplying it with Hanning window can slightly resist these noises, more powerful features are still needed for further improvements.

Apparently, features with multiple channels can be more representative and informative. In KCF, integrating them is simple and efficient. For Gaussian kernel function, vectors from different channels can be simply added together:
\begin{equation}
	\mathbf{k}^{\mathbf{xx}'}  =  \exp\left(
-\frac{1}{\sigma^2}\left(\|\mathbf{x}\|^2 + \|\mathbf{x}'\|^2 - 2\mathcal{F}^{-1}\left(
\sum\nolimits_c\hat{\mathbf{x}}_c^* \odot\hat{\mathbf{x}}_c'
\right)
\right)
\right)
\label{eq:34}
\end{equation}
where $c$ denotes the number of channels. With the multi-channel kernel functions, the famous HOG feature \cite{dalal2005histograms} has been successfully applied in KCF trackers with superior performance. 

Besides HOG, color attributes are also believed to be beneficial \cite{danelljan2014adaptive}.  Color attributes \cite{van2009learning}, or Color Names (CN), are the names of different colors defined by humans. In English, it has been concluded that there are 11 basic color terms \cite{van2009learning}, which include white, black, blue and so on. A map between the RGB combinations and linguistic color attributes can be found in \cite{van2009learning},  which is trained with images retrieved from Google-image search. Using the map, RGB values can be associated with a probabilistic 11 dimensional color vector with unit length. By further proposing an adaptive dimensionality reduction technique, the resulted tracker can achieve state-of-art accuracy with a considerably high speed, which is over 100 FPS. 

To some extents, two features are complementary to each other. HOG feature is mainly applied for analyzing the image gradients, while CN feature focuses on color representations. Based on the efficiency of integrating multi-channel data in (\ref{eq:34}), both HOG feature and CN feature can be fused together to facilitate robust tracking \cite{li2014scale}.

%  Afterwards, a three-channel colorful pixel can be represented by a probabilistic 11 dimensional color vector with unit length. The color-naming method has already profited object detection \cite{shahbaz2012color} and recognition tasks \cite{khan2012modulating}.  To make it more applicable and efficient in visual tracking, the dimensions of color attributes vector are reduced through a novel adaptive dimensionality reduction mechanism proposed by \cite{danelljan2014adaptive}. Specifically, the technique shares a similar procedure with Principal Component Analysis (PCA) except that the projection matrix will be updated adaptively. Lastly, with a MOSSE-like updating scheme, this color-naming method was shown to more effective than just using raw pixels. Furthermore, SAMF tracker integrates both the HOG feature and color-naming feature for robust visual tracking. 

\subsection{Handling Scale Variations}
\label{sec:scale}
Conventional CFTs, such as MOSSE and KCF, mainly employ fixed-sized windows for tracking, and they are unable to deal with target changes. To handle the scale variations, numerous algorithms have been proposed.

In SAMF and DSST, a searching strategy is applied to estimate scales of the target. In specific, windows with different sizes are sampled around the target, and are correlated with the learned filter. Subsequently, the window with the highest correlation score can be predicted as the new state. 

\begin{figure}[!h]
  \begin{center}
  \footnotesize
  \includegraphics[width=0.99\linewidth]{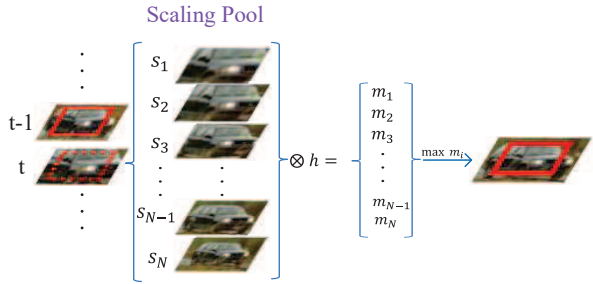}
  \end{center}
  \caption{Workflow of estimating scales using the searching strategy. Each time a new frame comes, windows with different sizes are cropped around previous position. By being correlated with a trained correlation filter $h$, corresponding confidence maps $m_i (i=1,\ldots,N)$ can be obtained. Then the window that can produce maximum confidence score is estimated as the new scale.}
  \label{fig:scale}
\end{figure}

Suppose the size of a window $i$ is denoted by a 2-dimensional vector $\mb{s}_i$. Let $\mb{s}_o$ denote a template window size, then we can have $\mb{s}_i=a_i\mb{s}_o$ where $a_i$ denotes a scale parameter given by a scaling pool $S=\{a_1,a_2, \ldots , a_N\}$ of $N$ positive numbers. As shown in Figure \ref{fig:scale}, the size of current target can be estimated by searching the window with a maximum correlation score among the sampled windows. In SAMF, $S$ is set by constant values ranged from 0.985 to 1.015, and in DSST, $S=\left\{ a^n|n=\lfloor -\frac{N-1}{2} \rfloor, \ldots , \lfloor \frac{N-1}{2} \rfloor \right\}$. The major difference between the two trackers is that SAMF processes one window at a time while a 3-dimensional correlation filter is employed to search the best scale in DSST. Some recent CFTs \cite{hong2015multi} and \cite{ma2015long} also apply the scaling pool method.

Different from using scaling pool, scale varying issue can also be solved by some part-based tracking methods. In \cite{liu2015real}, the whole target is determined through a Bayesian framework. In another part-based tracker \cite{li2015reliable}, a statistical method is employed, which records and sorts the variations of relative positions between different sub-patches to estimate scales. In practice, the performance of these methods is also promising.

In addition, it is worth mentioning that STC has its own scheme to deal with scale variations. According to the formulas used in Section \ref{sec:STC}, suppose the new estimated center of the target is $\bm{p}_o$, and $\ell(\bm{p}_o)$ is its computed confidence score. Then scales can be estimated by:
\begin{equation}
    \bm{s}_t' = \sqrt{\frac{\ell\left( (\bm{p}_o)_t\right)}
    {\ell\left( (\bm{p}_o)_{t-1}\right)}}
\label{eq:35}
\end{equation}
where $\bm{s}_t'$ is the predicted scale at time $t$. To smooth the predictions, estimated scales are averaged over $n$ consecutive frames, and linear interpolation is used for prediction:
\begin{equation}
\left\{
 \begin{array}{lll}
\overline{\bm{s}} &=& 
\frac{1}{n} \sum_{i=1}^n \bm{s}_{t-i}' \\
\bm{s}_{t+1} &=& (1-\lambda)\bm{s}_t + \lambda \overline{\bm{s}}
\end{array}
\right.  
\label{eq:36}
\end{equation}
where $\lambda$ is a fixed parameter. With the estimated size of the target, the parameter $\sigma$ of weight function (\ref{eq:25}) is also required to be updated\footnote{Detailed derivations of (\ref{eq:35}), (\ref{eq:36}) and (\ref{eq:37}) can be found at \url{http://www4.comp.polyu.edu.hk/~cslzhang/STC/STC.htm}}:
\begin{equation}
    \sigma_{t+1} = \bm{s}_t\sigma_t
\label{eq:37}
\end{equation}

While conducting the experiments, we found that the estimation in STC is sometimes unstable, because the computed $\bm{s}_t'$ can be extremely large if the denominator of (\ref{eq:35}) is close to zero. 

\subsection{Part-based Tracking}
Instead of learning a holistic appearance model, various part-based tracking algorithms \cite{cehovin2011adaptive, izadinia20122t, ross2008incremental, shu2012part, yang2012online, jia2012visual} have been proposed, in which the target is tracked by its local appearance.  If the target is partially occluded, its remaining visible parts can still represent the target and thus the tracker is able to continue tracking. In \cite{wu2013online, wu2015object}, experimental results have shown that the local representations are effective for object tracking. Therefore introducing part-based tracking strategy in CFTs is supposed to be advantageous.

Recently, \cite{liu2015real, li2015reliable} have made successful attempts to apply part-based tracking strategy to CFTs. In \cite{liu2015real}, 5 parts of the target are independently tracked by KCF trackers. When a new frame comes, confidence maps of these tracked parts are first computed. By assigning adaptive weights to these maps, a joint map can be constructed to predict new state using particle filter method. Another tracker, which is called Reliable Patch Trackers (RPT) \cite{li2015reliable}, also exploits the use of local contexts and treats KCF as its base tracker. However, the tracked parts in RPT are automatically selected by sampling, whose reliabilities are estimated on-the-fly. A reliable patch is defined as being trackable and sticking on the target. If a part is no longer reliable, it will be discarded and re-sampled around the target. After obtaining the tracking results of reliable patches, new state of the target is predicted by a Hough Voting-like scheme.   

In general, part-based tracking strategy can be helpful to gain robustness against partial occlusions. The main difficulty of developing a robust part-based CFT is how to design an appropriate mechanism to handle multiple results from different tracked parts. According to the introduced studies, particle filter method has proven to be an effective solution.

\subsection{Long-term Tracking}
One other vital challenge in visual tracking is the absence of the target. If the target partially or fully disappears from the view, conventional CFTs can be easily distracted by irrelevant objects because they do not contain a long-term component.  As a consequence, introducing long-term tracking methods is believed to be favorable for improving correlation filter-based tracking methods.

For long-term tracking, there exists several studies \cite{dinh2011context, kalal2012tracking, hua2014occlusion, supancic2013self, lebeda2013long, pernici2014object}, some of which introduce a re-detection module while the others attempt to learn conservative appearance of the target. For example, TLD tracker \cite{kalal2012tracking} trains a detector with an expert of false negative samples and an expert of false positive samples. If the tracking module in TLD fails, this trained detector can then re-initialize the tracker. On the other hand, the tracker of \cite{supancic2013self} conservatively learns the target appearance from reliable frames with a self-paced learning scheme. Regarding to CFTs, two recent studies \cite{hong2015multi, ma2015long} have successfully cooperated CFTs with long-term tracking.

Inspired by a biological memory model called Atkinson-Shiffrin Memory Model (ASMM) \cite{atkinson1968human}, a MUlti-Store Tracker (MUSTer) based on a cooperative tracking framework was proposed \cite{hong2015multi}. In ASMM, there are short-term memory and long-term memory in human brains. Short-term memory, which updates aggressively and forgets information quickly, stores local and temporal information, while long-term memory, which updates conservatively and maintains information for a long time, retains general and reliable information.  With short-term and long-term memory working together, both efficiency and robustness can be achieved. By considering CFTs as efficient short-term trackers, introducing long-term tracking methods is supposed to complement the shortage of CFTs. In MUSTer, the long-term part is a key points-based method.  In the course of tracking, key points of the target are maintained or discarded based on a forgetting curve, and then retrieved for locating the target if the CFT fails. Experiments have shown that MUSTer have surpassed various state-of-art trackers in different benchmarks.

Another method of introducing long-term tracking was proposed by Ma \textit{et al.} \cite{ma2015long}. In this method, a re-detection component is added into the tracking system. Similar to TLD tracker, the re-detection procedure is carried out based on an online random fern classifier, whose training samples are collected by a k-nearest neighbour (kNN) classifier. With this cooperation, the resulted tracker, namely LCT, has shown to be able to handle well with long term tracking. 

\section{Experiments}
\label{sec:exp}
In this section, both quantitative and qualitative experiments have been conducted on large scale benchmarks to evaluate the advantages brought by correlation filter-based tracking framework. To carry out comprehensive and fair comparisons, additional 29 popular trackers are evaluated as well, and parameters of all the evaluated trackers are set as default and fixed during the experiments. The hardware we have used in the evaluation is a cluster node (3.4GHz Intel Xeon CPU, 8 cores 32GB RAM).

\subsection{Experiment Setup}
\subsubsection{Compared Trackers}
For CFTs, trackers with currently available source codes are selected in our evaluations (except ASEF and MOSSE are implemented by ourselves), which are regularized ASEF \cite{bolme2010visual, bolme2009average}, MOSSE\footnote{Original implementation can be found in \url{http://www.cs.colostate.edu/~vision/ocof_toolset_2012/index.php}}\cite{bolme2010visual}, CSK (with raw pixels) \cite{henriques2012exploiting}, KCF\footnote{\url{http://home.isr.uc.pt/~henriques/circulant/}} (with HOG features) \cite{henriques2014high}, CN\footnote{http://liu.diva-portal.org/smash/record.jsf?pid=diva2\%3A711538\&dswid=-7492}\cite{danelljan2014adaptive}, DSST\footnote{\url{https://github.com/gnebehay/DSST}}\cite{danelljan2014accurate}, SAMF\footnote{\url{https://github.com/ihpdep/samf}}\cite{li2014scale}, STC\footnote{\url{http://www4.comp.polyu.edu.hk/~cslzhang/STC/STC.htm} (Note that STC may crash during the scale estimations, thus we simply fix the parameters before the detected errors)}\cite{zhang2014fast}, MUSTer\footnote{\url{https://sites.google.com/site/zhibinhong4131/Projects/muster}}\cite{hong2015multi}  and RPT\footnote{\url{https://github.com/ihpdep/rpt}}\cite{li2015reliable}. Other competing trackers used for comparisons include 28 trackers from the code library of Online Object Tracking Benchmark (OOTB)\footnote{\url{http://cvlab.hanyang.ac.kr/tracker_benchmark/}} \cite{wu2013online,wu2015object} and a recent state-of-art tracker MEEM\footnote{\url{http://cs-people.bu.edu/jmzhang/MEEM/MEEM.html}}\cite{zhang2014meem}. 

\subsubsection{Test Sequences}
All the test sequences in our evaluation come from OOTB \cite{wu2013online, wu2015object}. In original OOTB, there are 51 different tracking tasks with fully annotated attributes, which include scale variations, illumination variations, rotations and so on. Then in later OOTB, the number of tasks has been extended to 100. In our experiments, original OOTB is mainly used to compare trackers since it is more representative. The later OOTB is used as extended dataset to verify the performance of CFTs.

\begin{figure*}[!t]
\begin{center}
   \begin{tabular}{ccc}
   \includegraphics[width=0.3\linewidth]{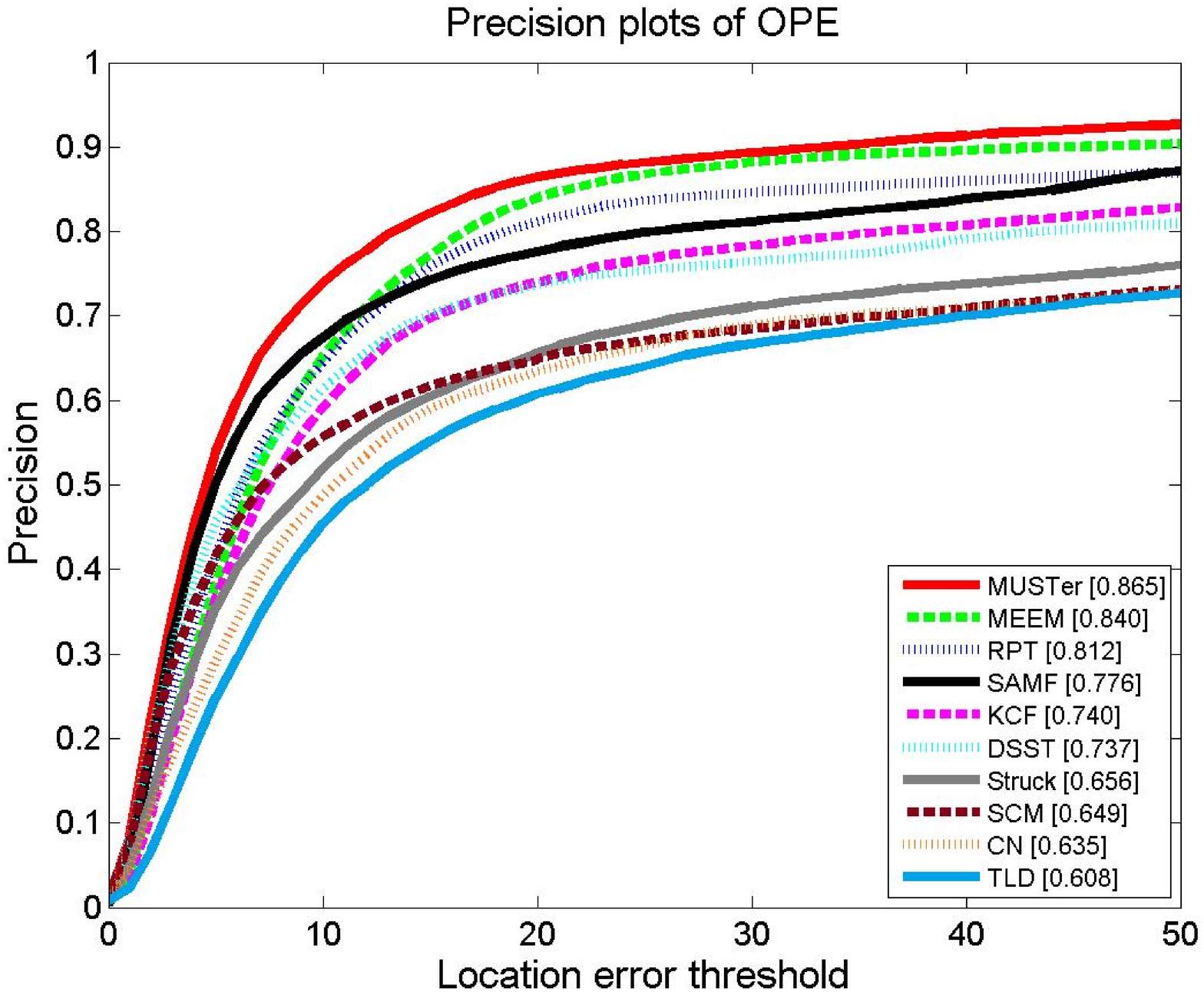} &
   \includegraphics[width=0.3\linewidth]{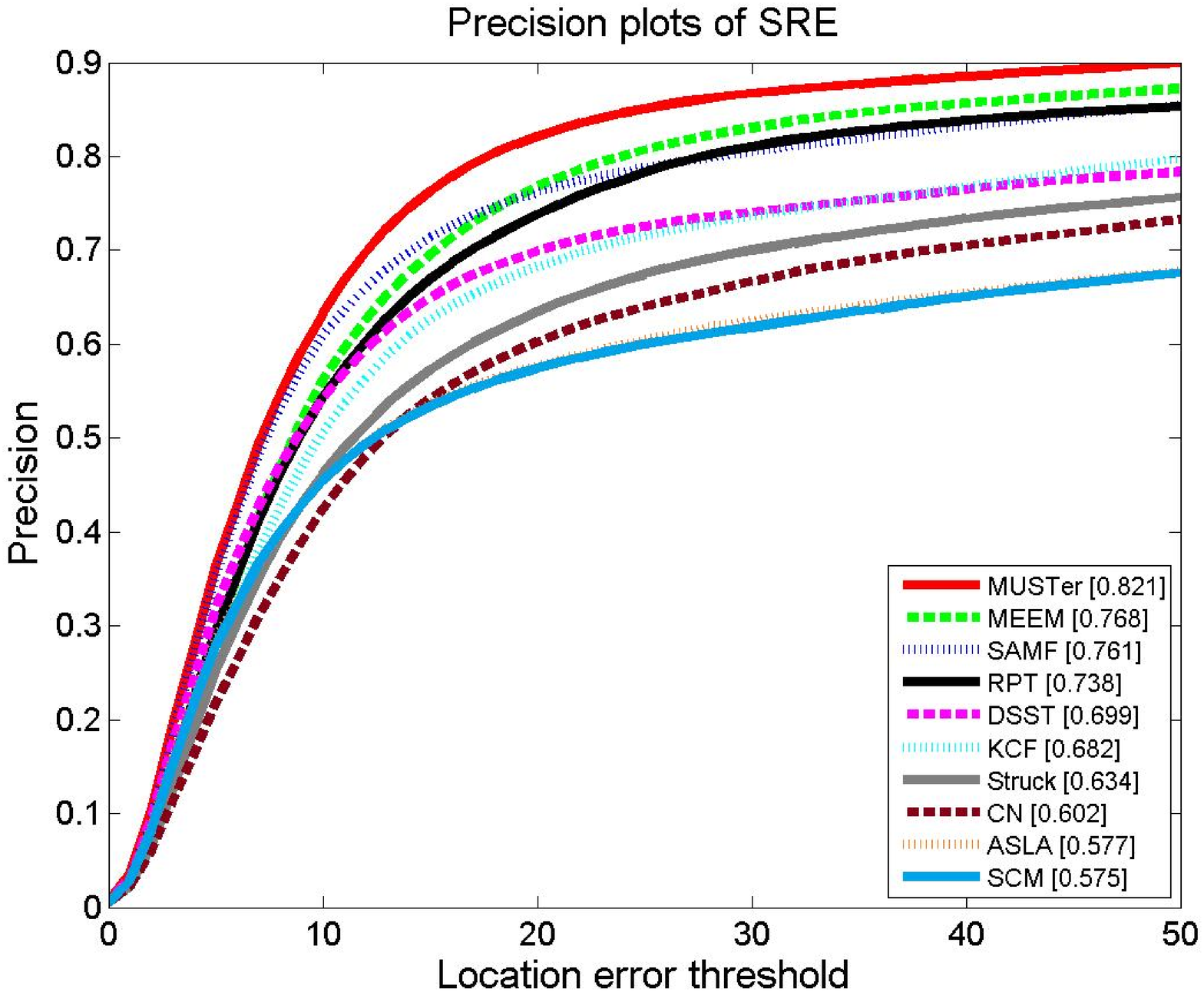} &
   \includegraphics[width=0.3\linewidth]{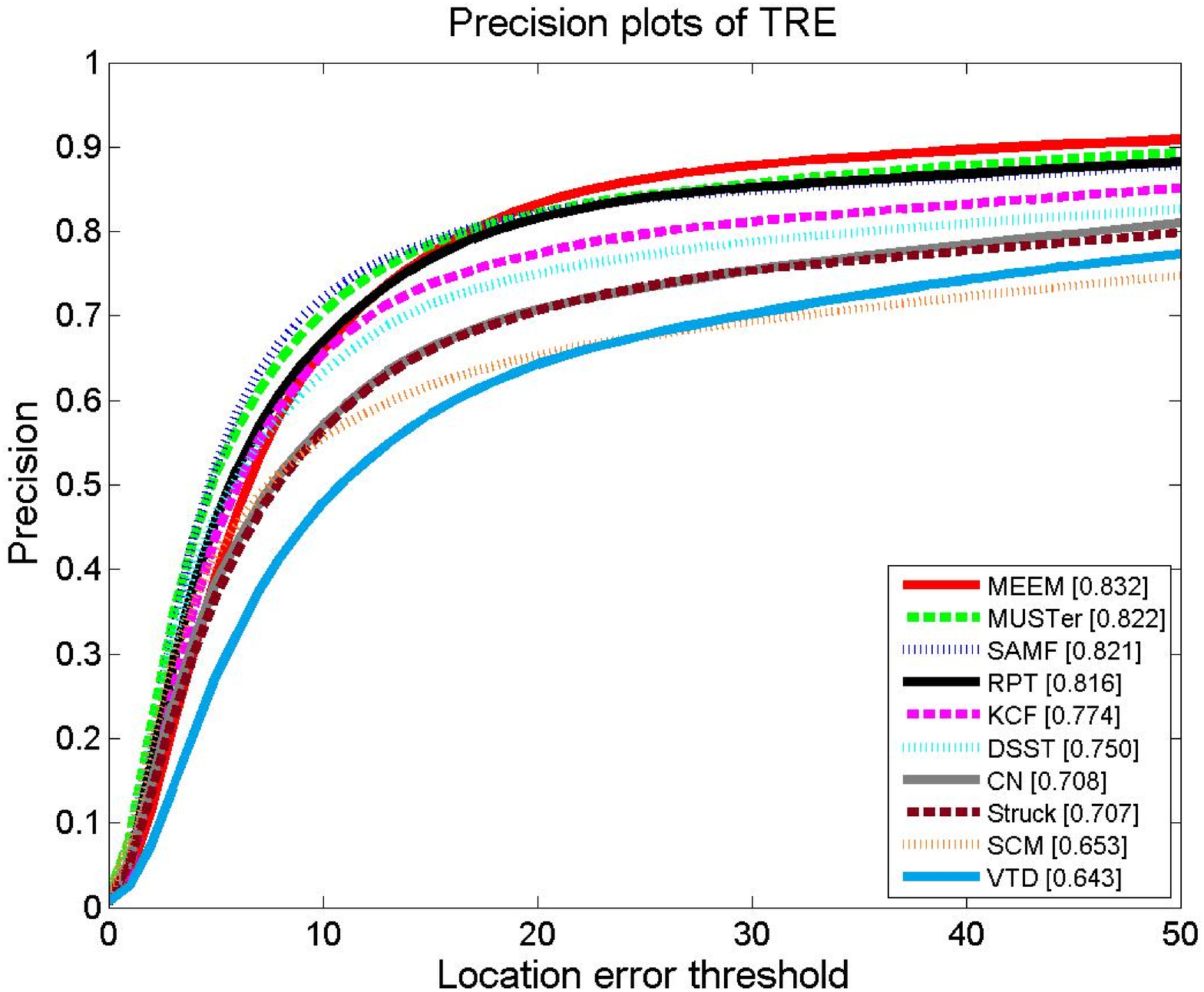}\\
   \includegraphics[width=0.3\linewidth]{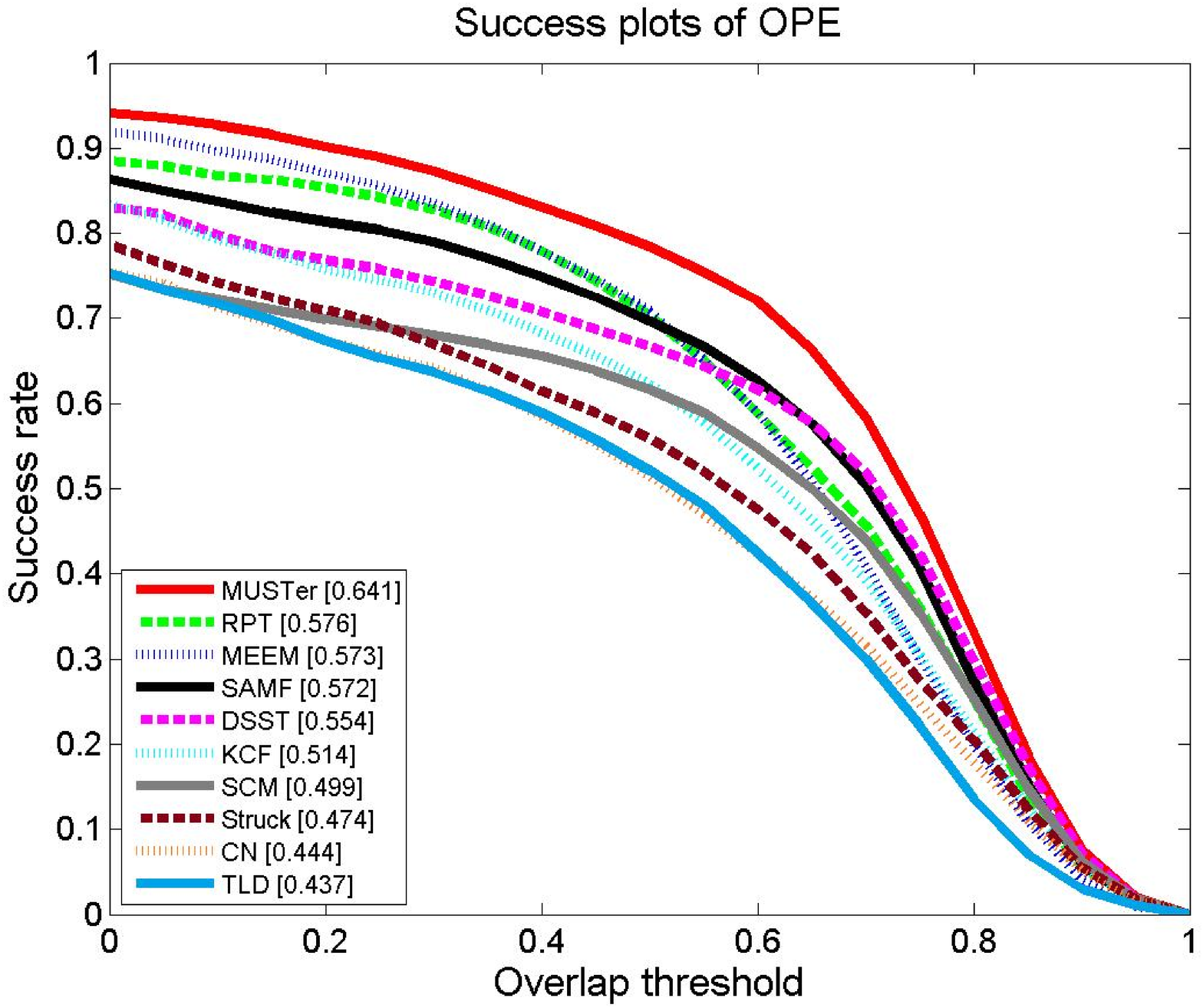} &
   \includegraphics[width=0.3\linewidth]{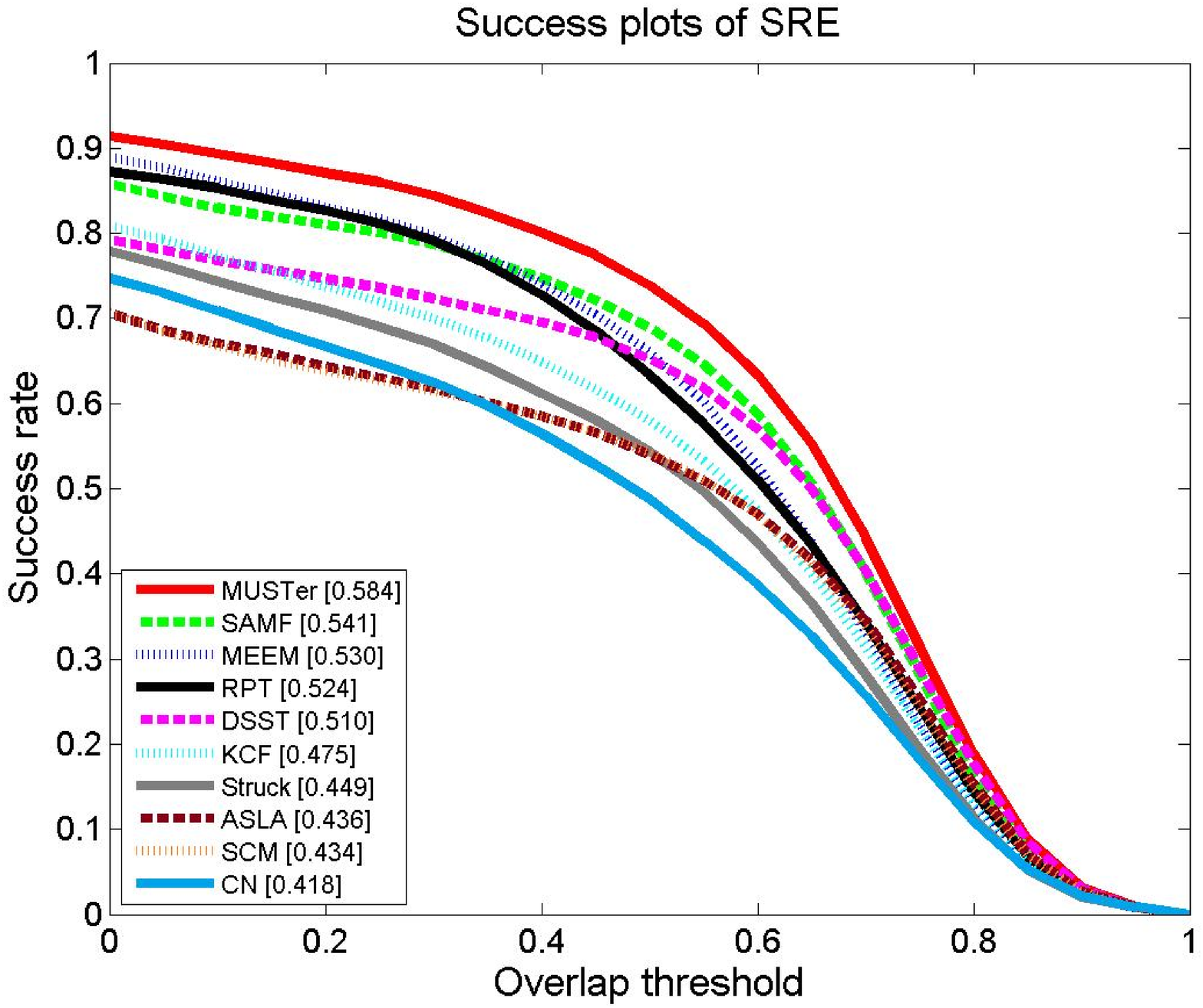} &
   \includegraphics[width=0.3\linewidth]{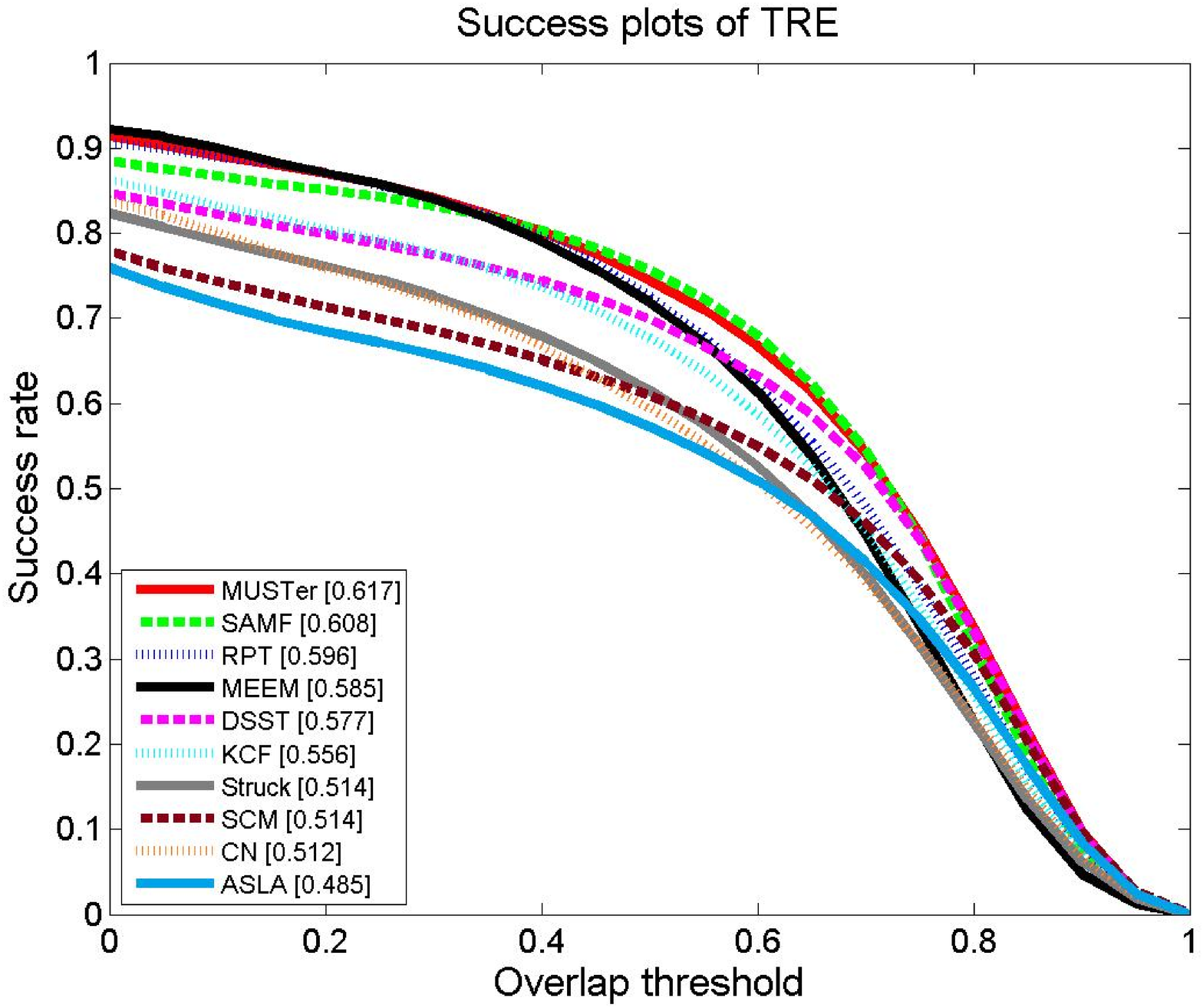}
   \end{tabular}
\end{center}
\caption{Plots of OPE, SRE and TRE in OOTB. Trackers with best 10 scores are presented in the legends.}
\label{fig:1}
\end{figure*}

\begin{figure*}[!h]
\begin{center}
   \begin{tabular}{cccc}
   %out of plane,in plane,  scale, deform, 
   \includegraphics[width=0.25\linewidth]{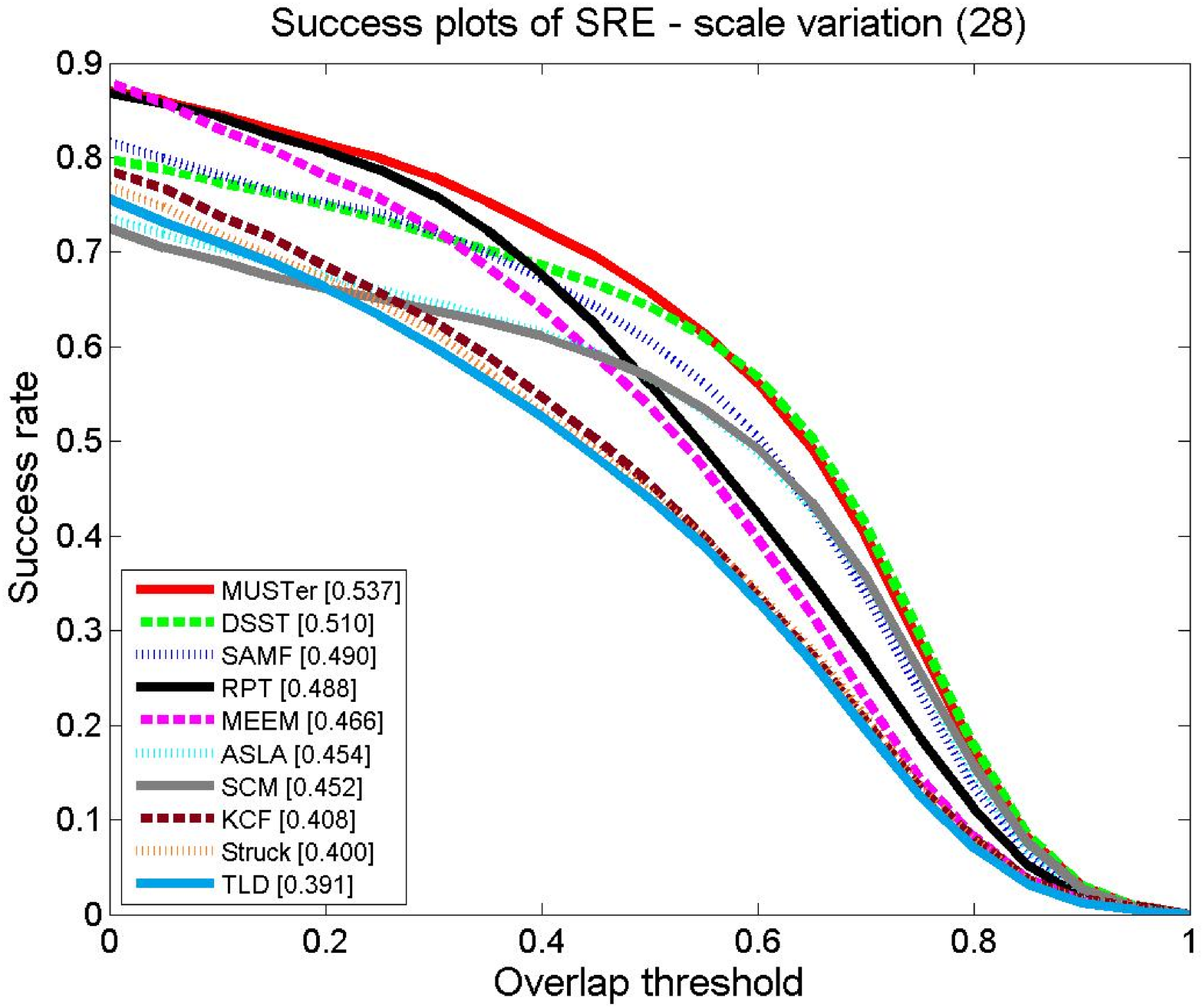} &
   
   \includegraphics[width=0.25\linewidth]{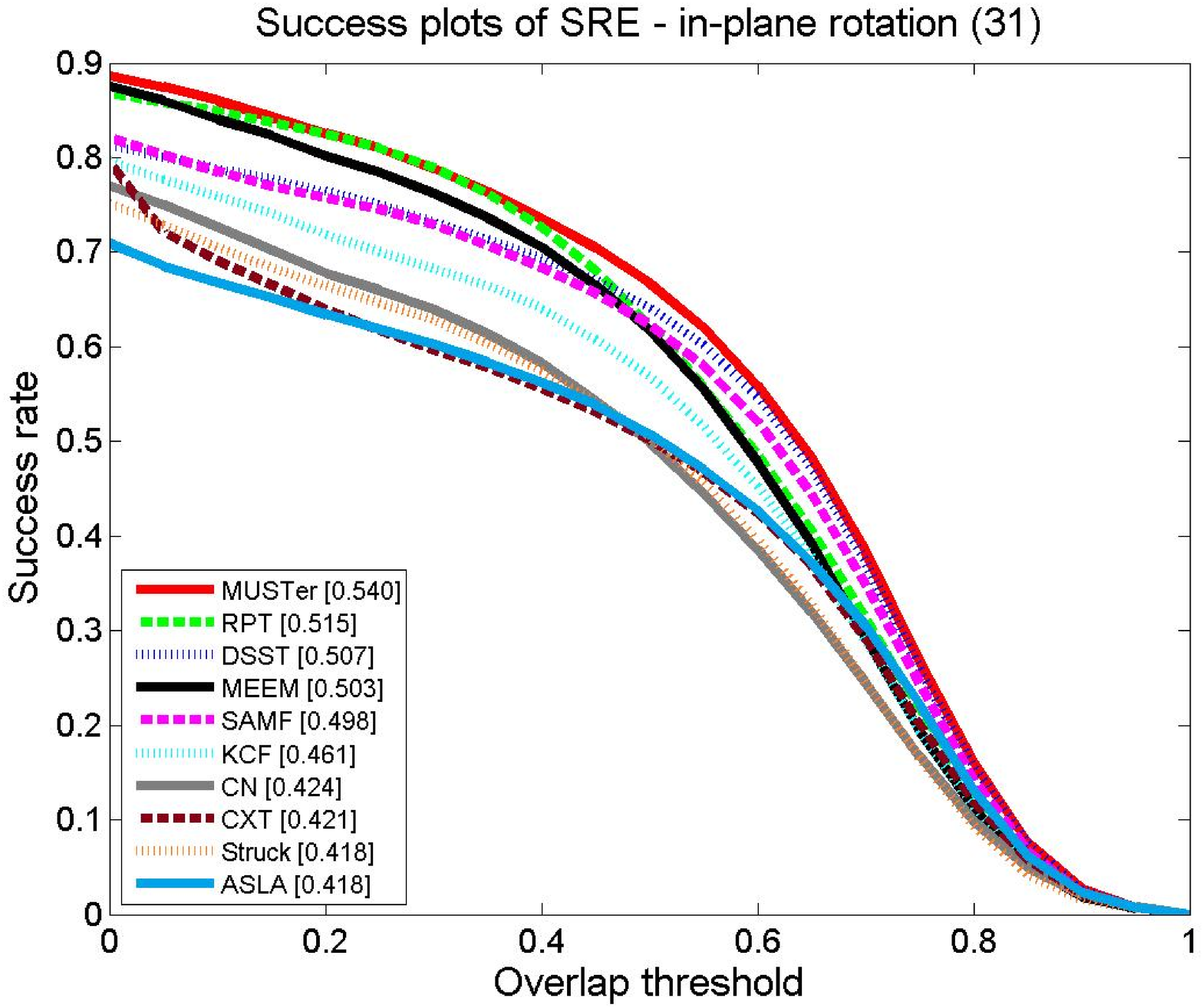} &
   
   \includegraphics[width=0.25\linewidth]{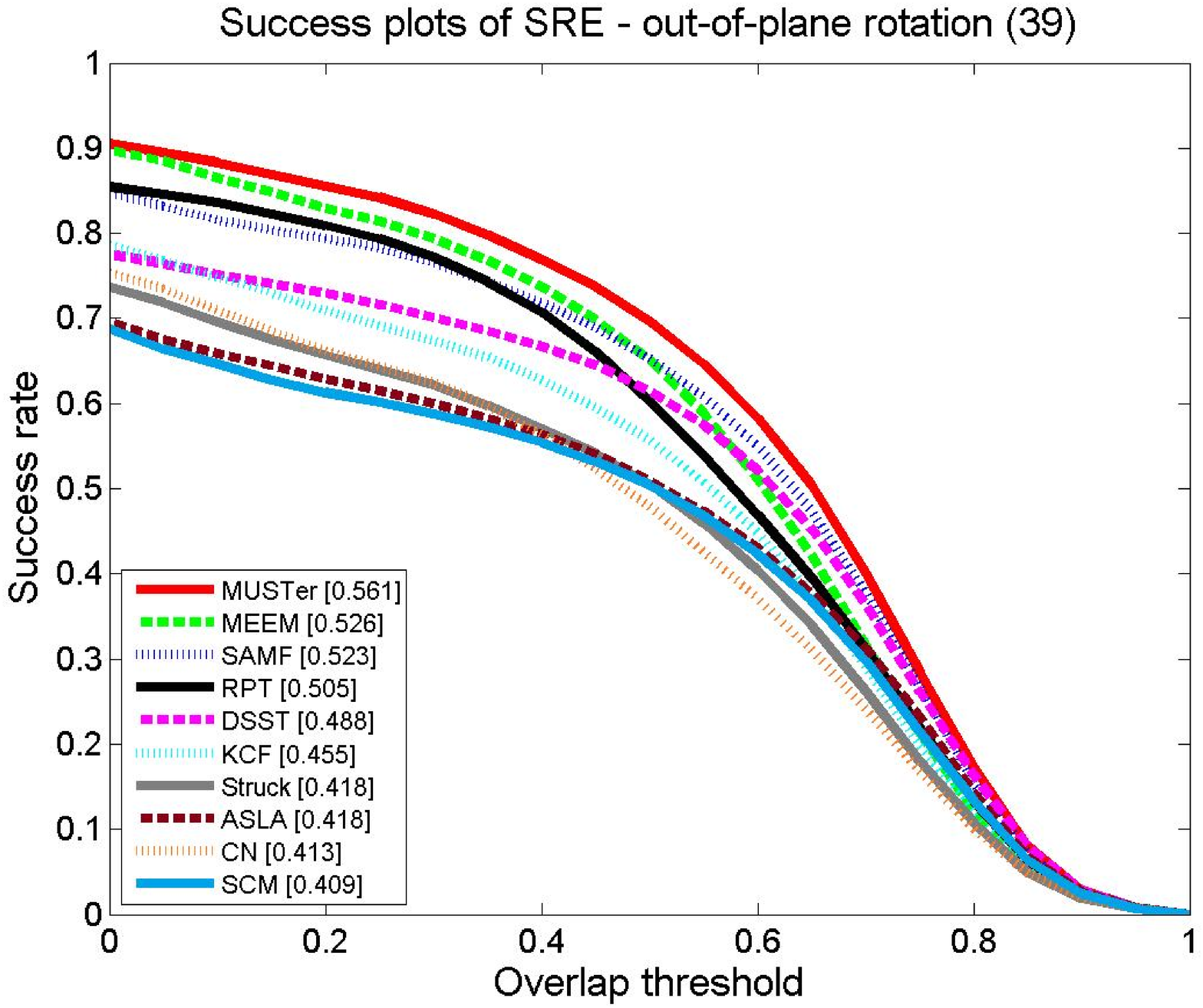} &
   
   \includegraphics[width=0.25\linewidth]{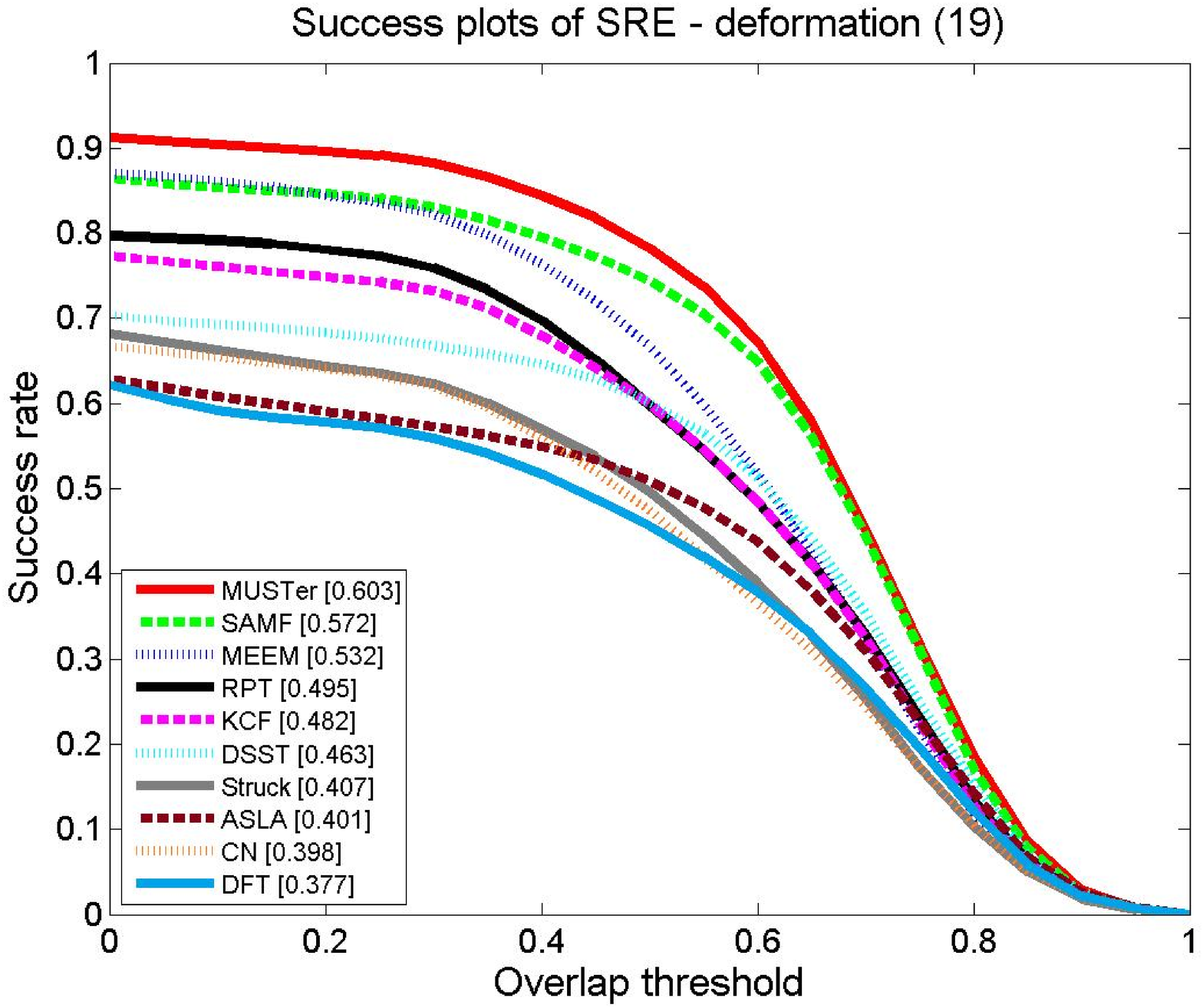} \\
    %fast motion, illum, occ, bac
   \includegraphics[width=0.25\linewidth]{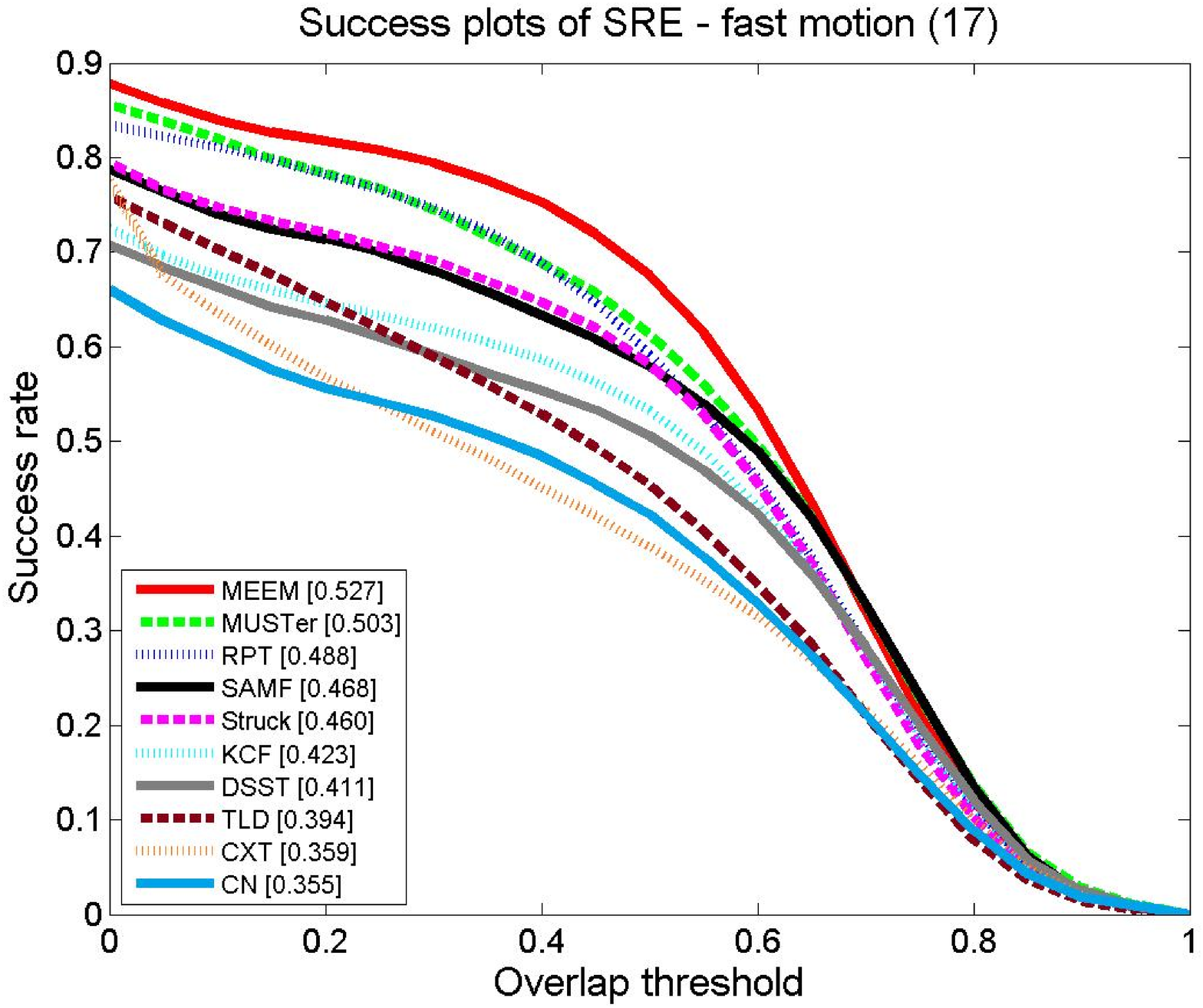} & 
   \includegraphics[width=0.25\linewidth]{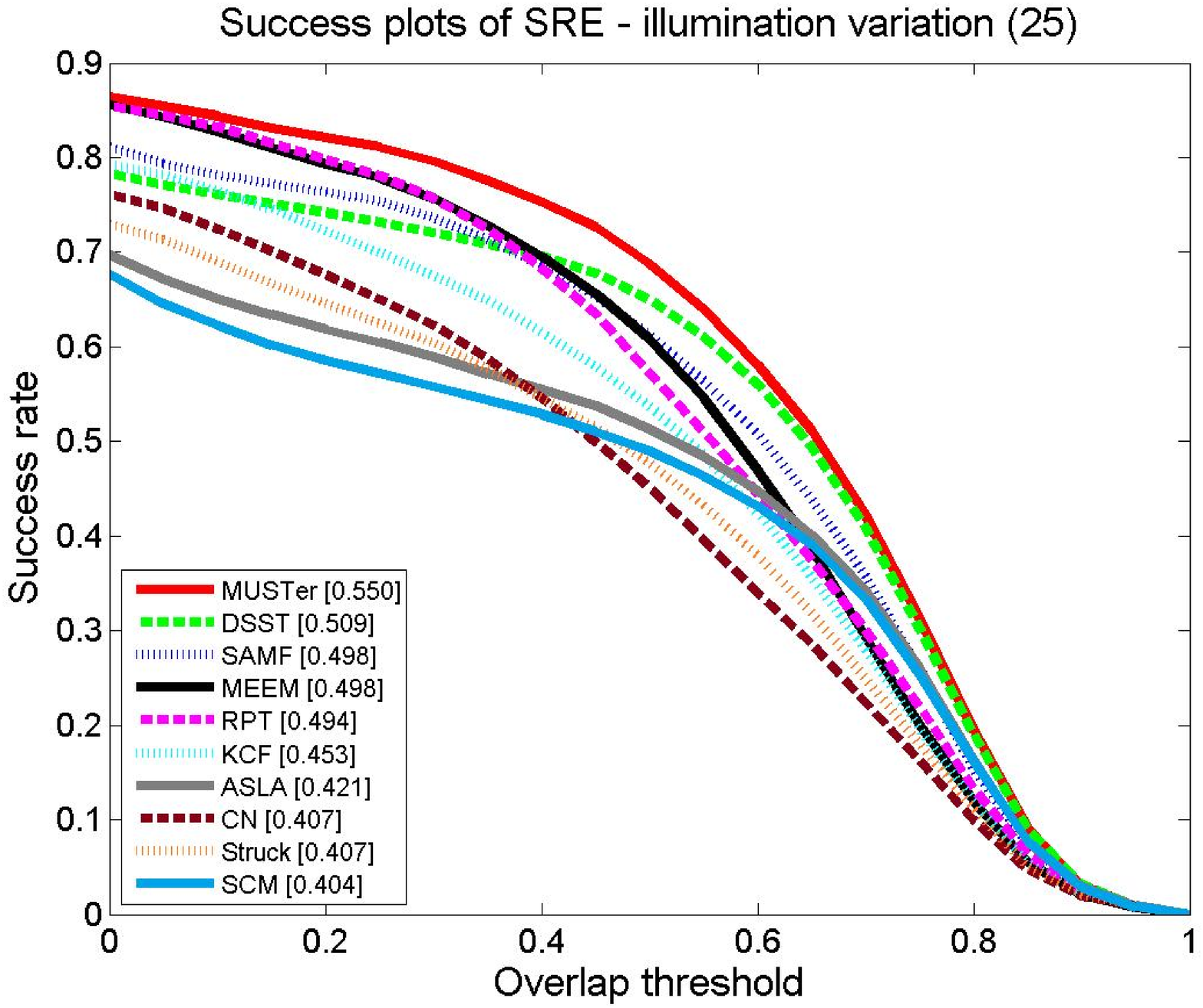} &
   \includegraphics[width=0.25\linewidth]{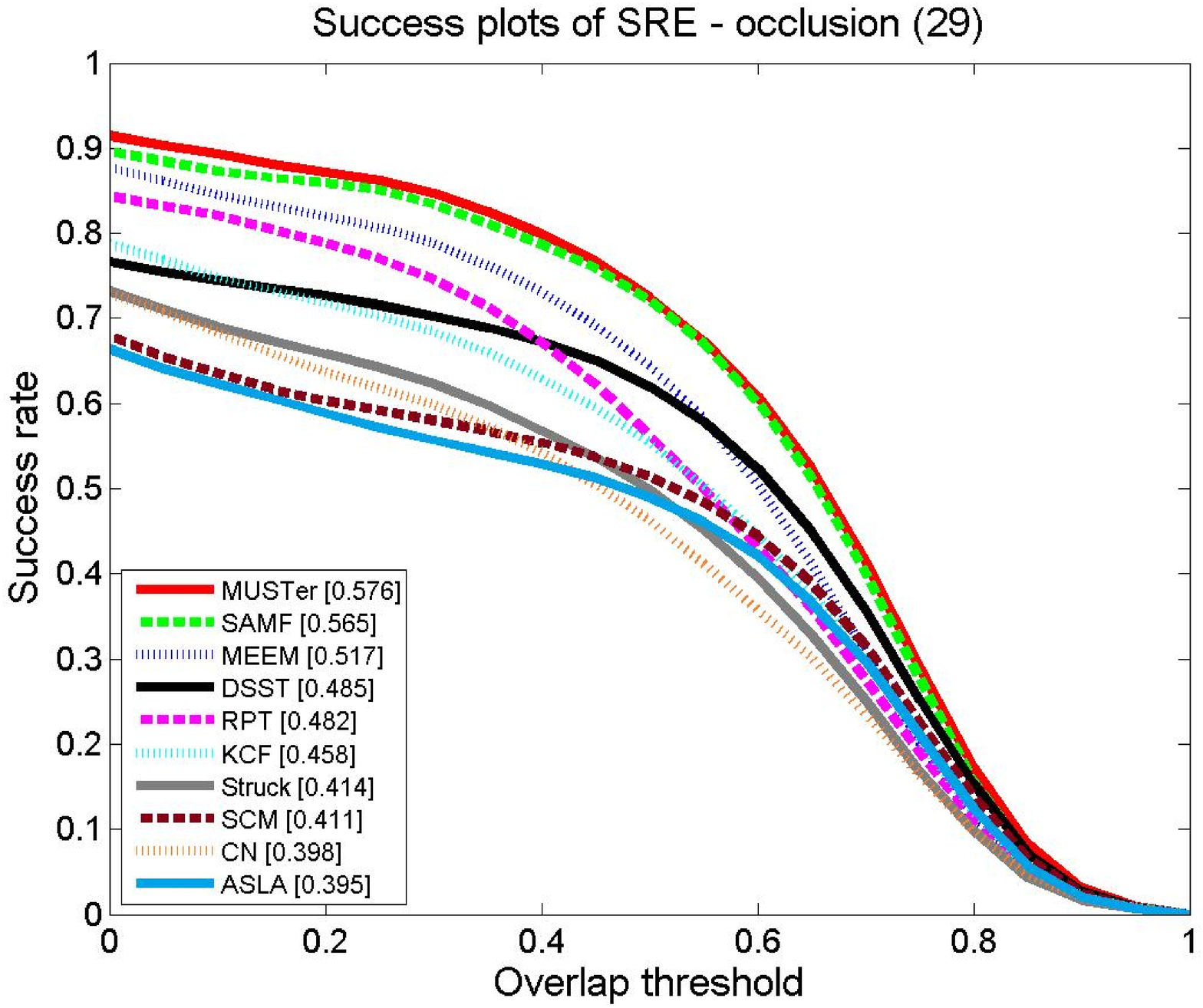} &
   \includegraphics[width=0.25\linewidth]{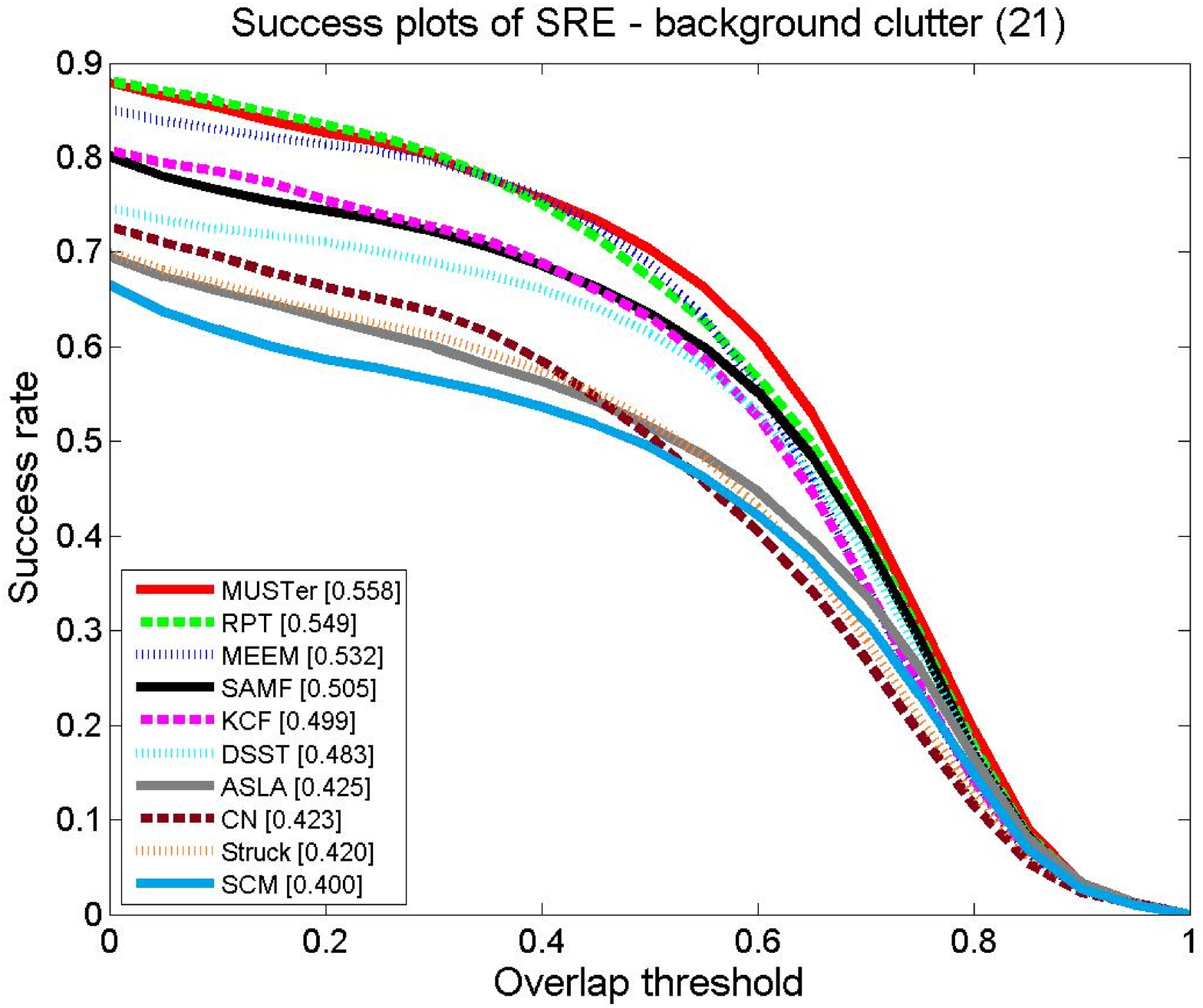}\\
   %blur, out of view, low res
   \includegraphics[width=0.25\linewidth]{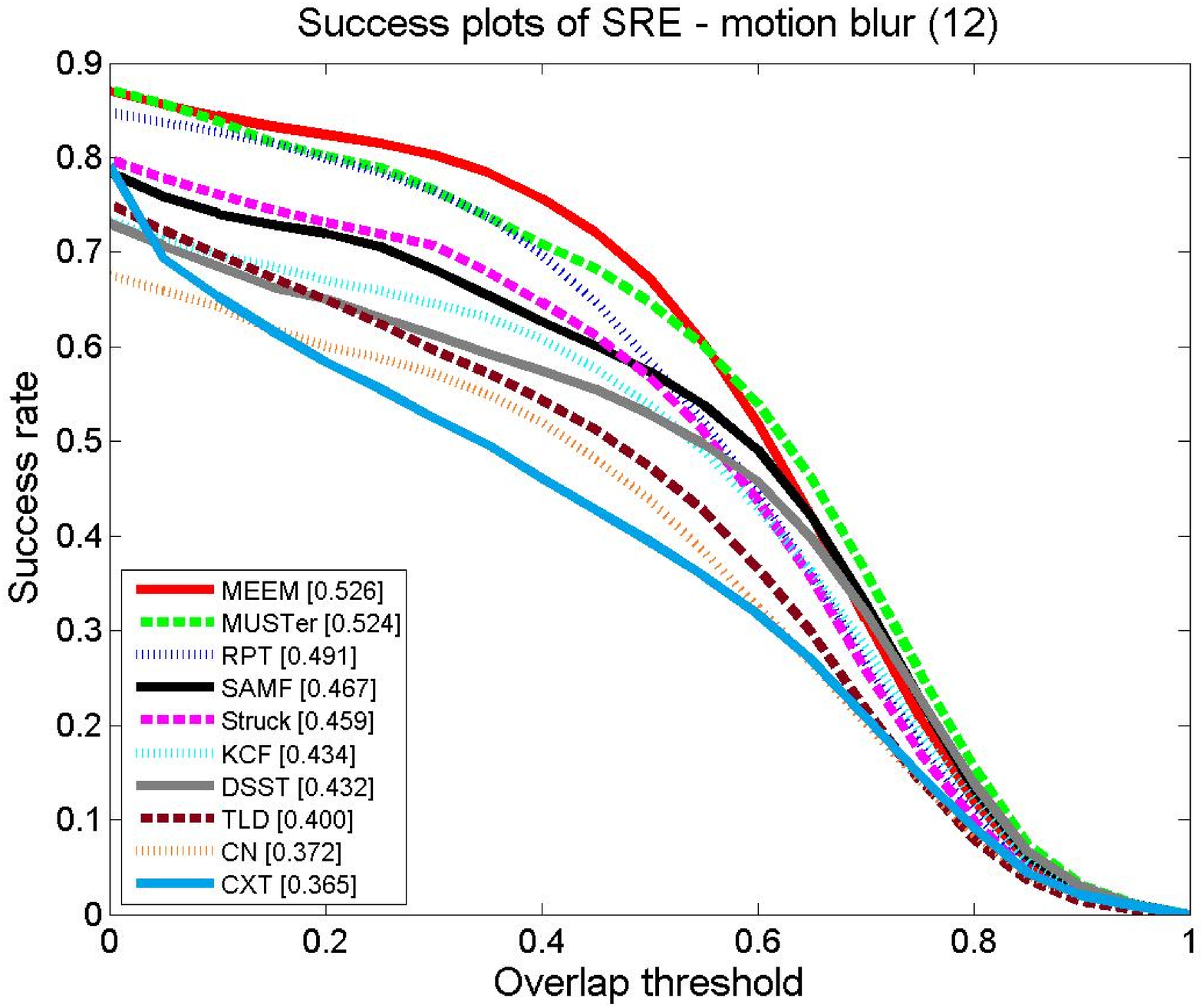} &
   \includegraphics[width=0.25\linewidth]{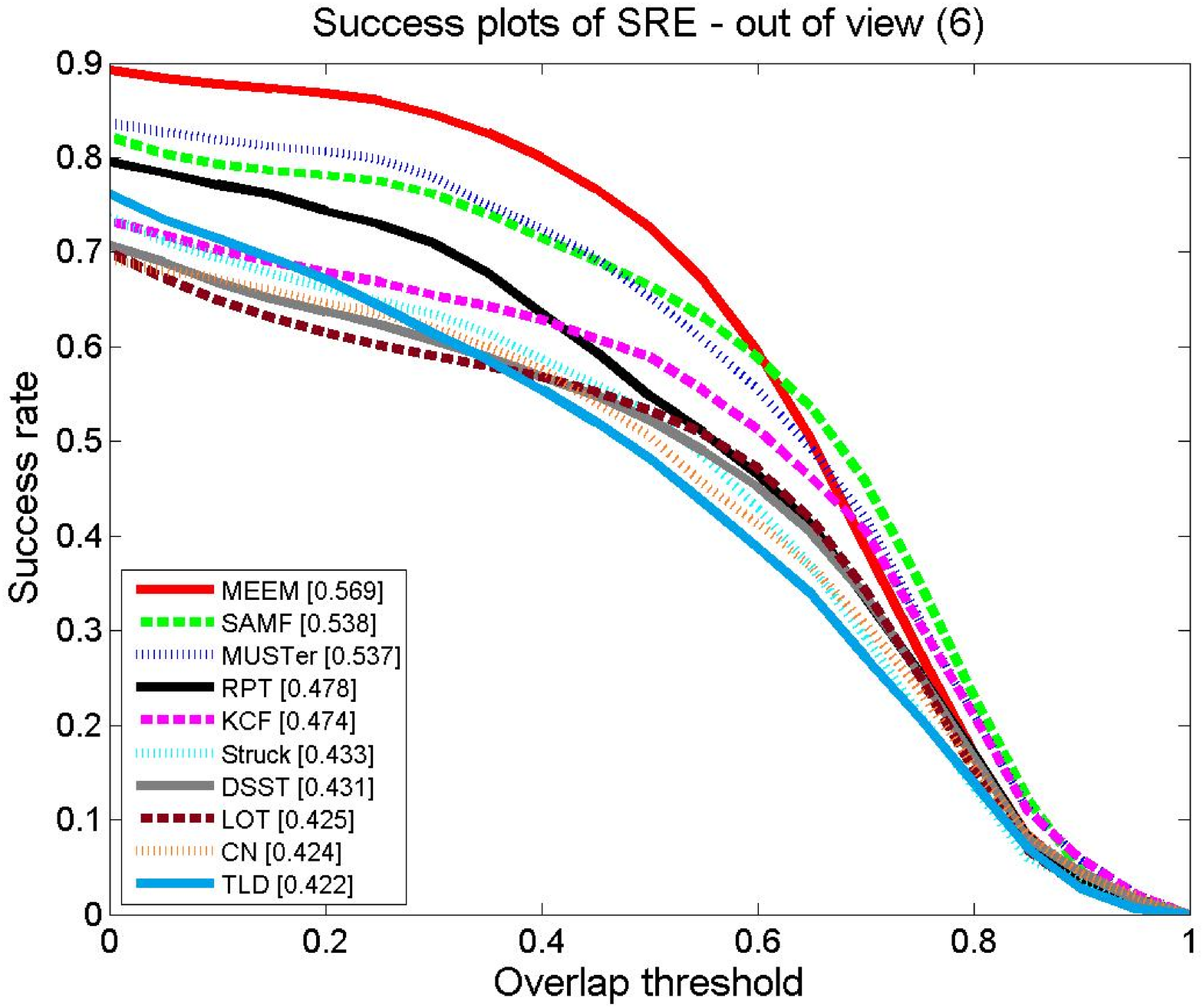} &
   \includegraphics[width=0.25\linewidth]{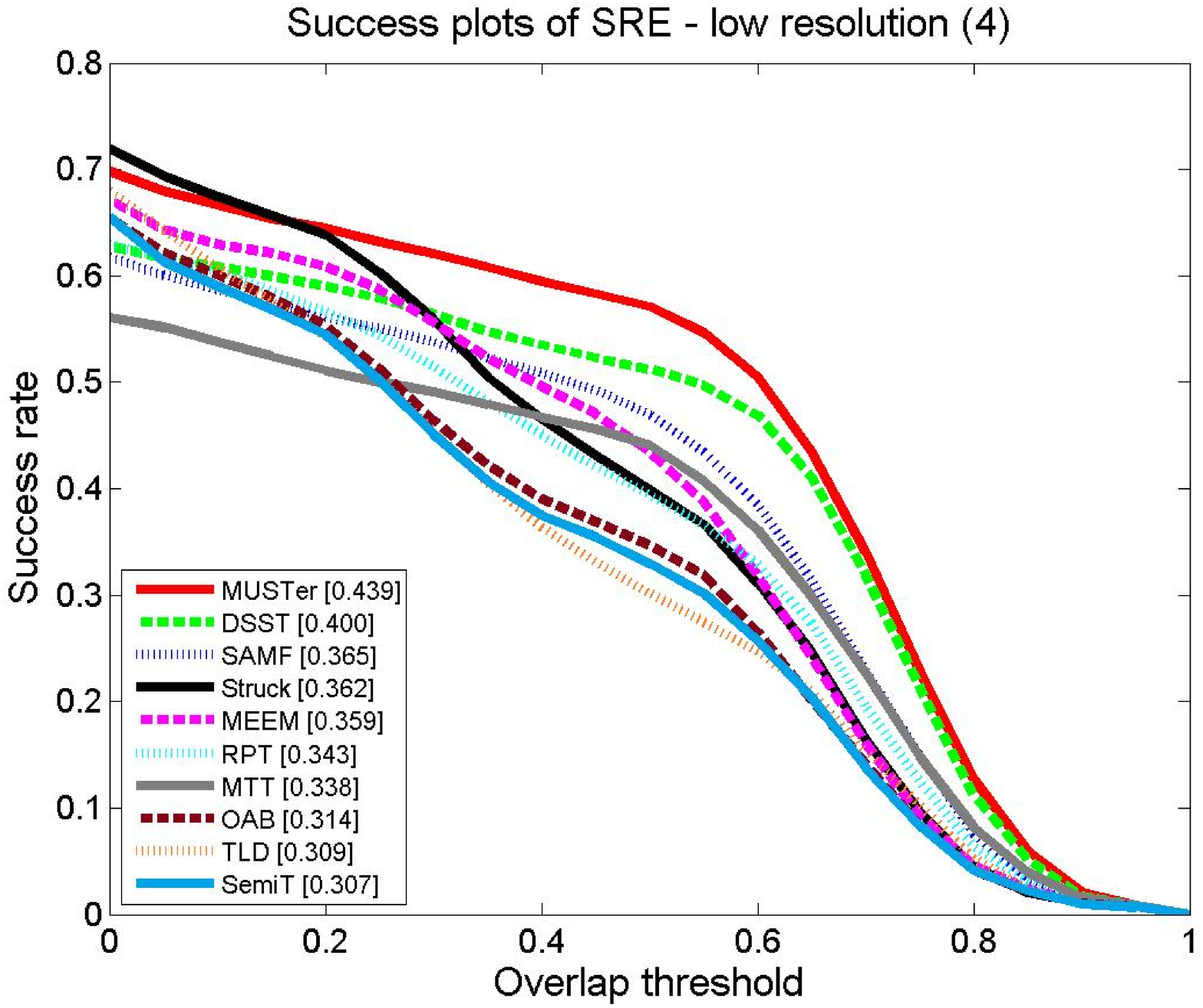} &
   \end{tabular}
\end{center}
\caption{Attribute-based success plots of SRE. The trackers with best 10 AUC scores are presented in the legends.
}
\label{fig:2}
\end{figure*} 
\begin{figure*}[!t]
\begin{center}
   \begin{tabular}{ccc}
   \includegraphics[width=0.3\linewidth]{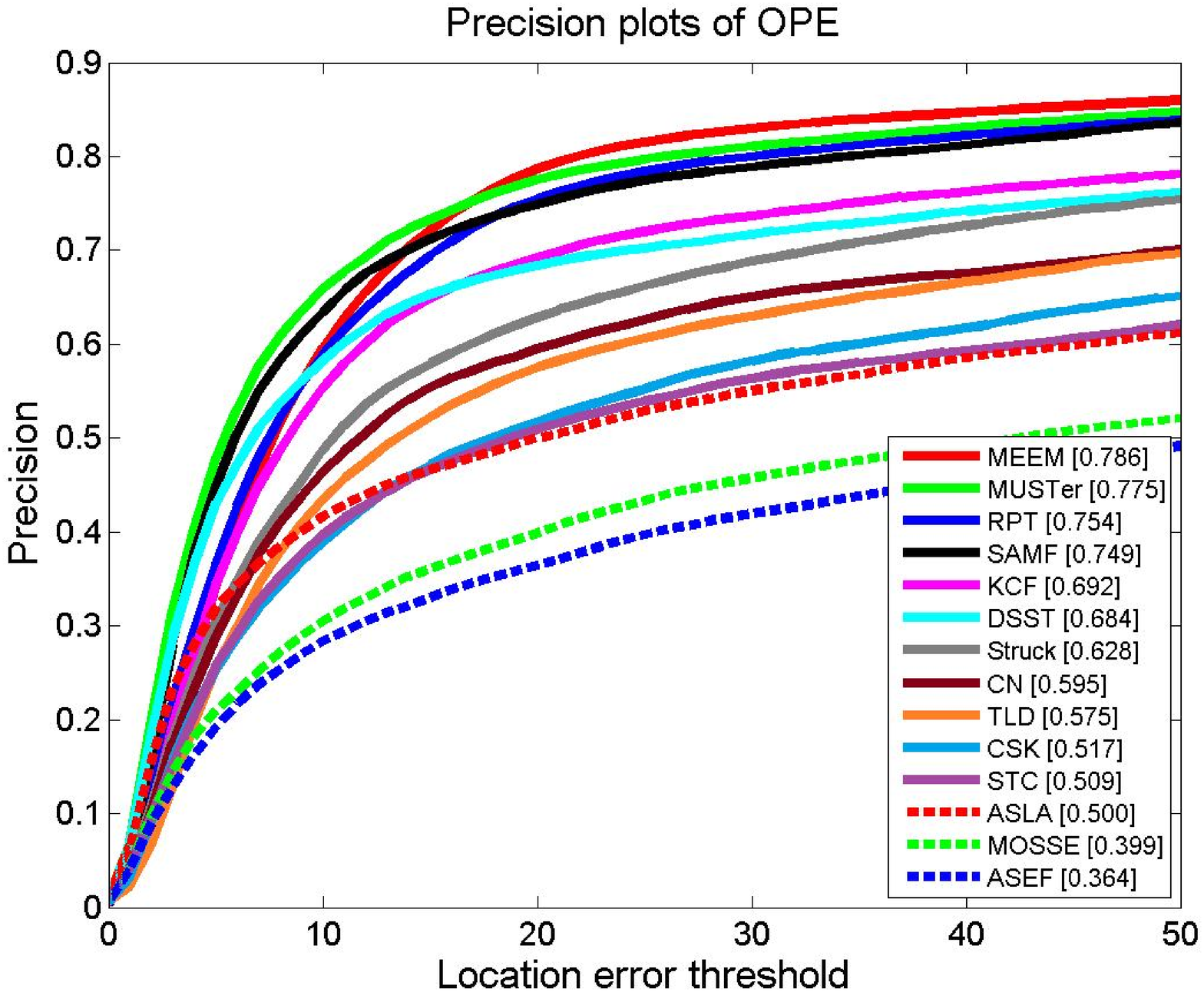} &
   \includegraphics[width=0.3\linewidth]{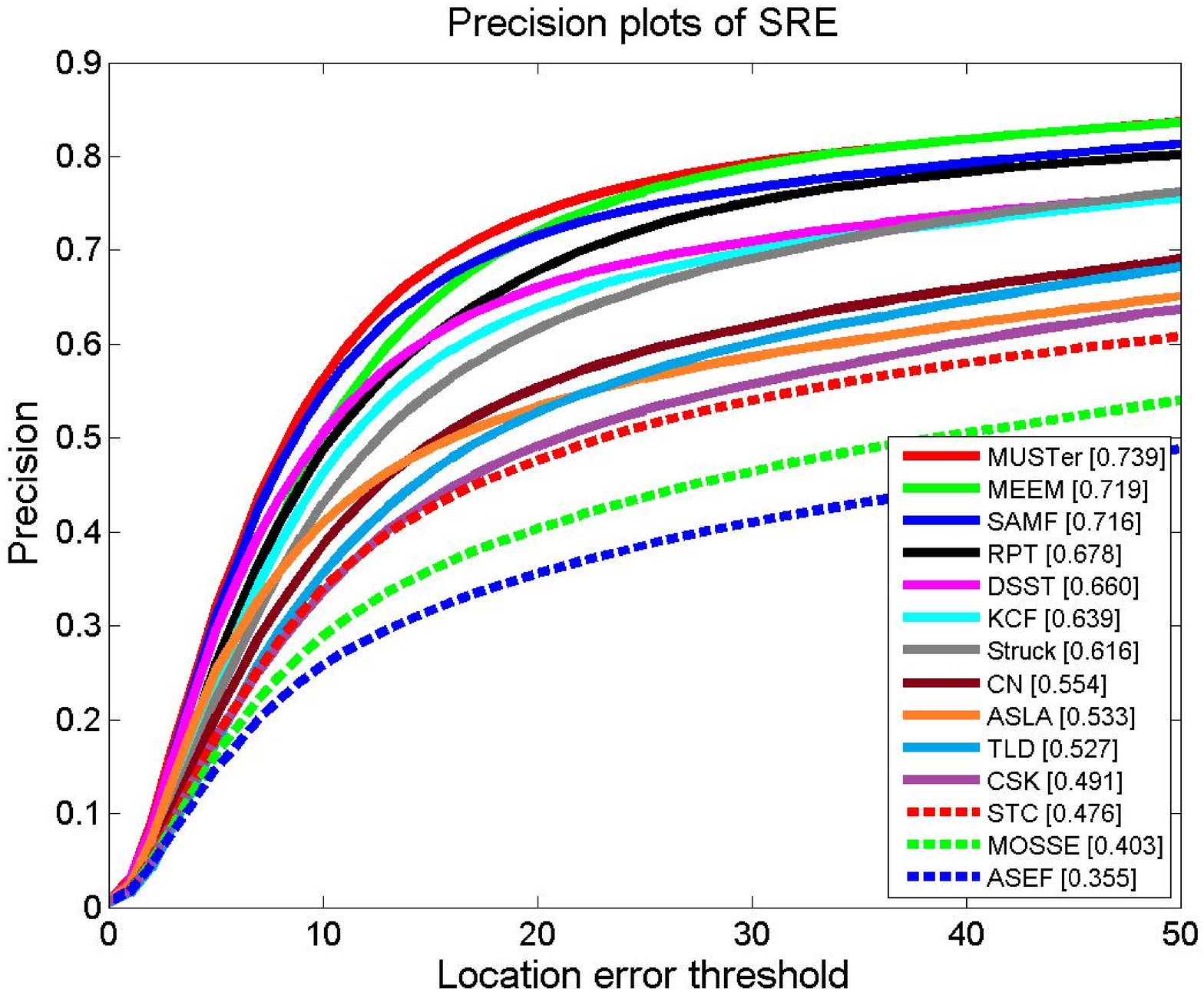} &
   \includegraphics[width=0.3\linewidth]{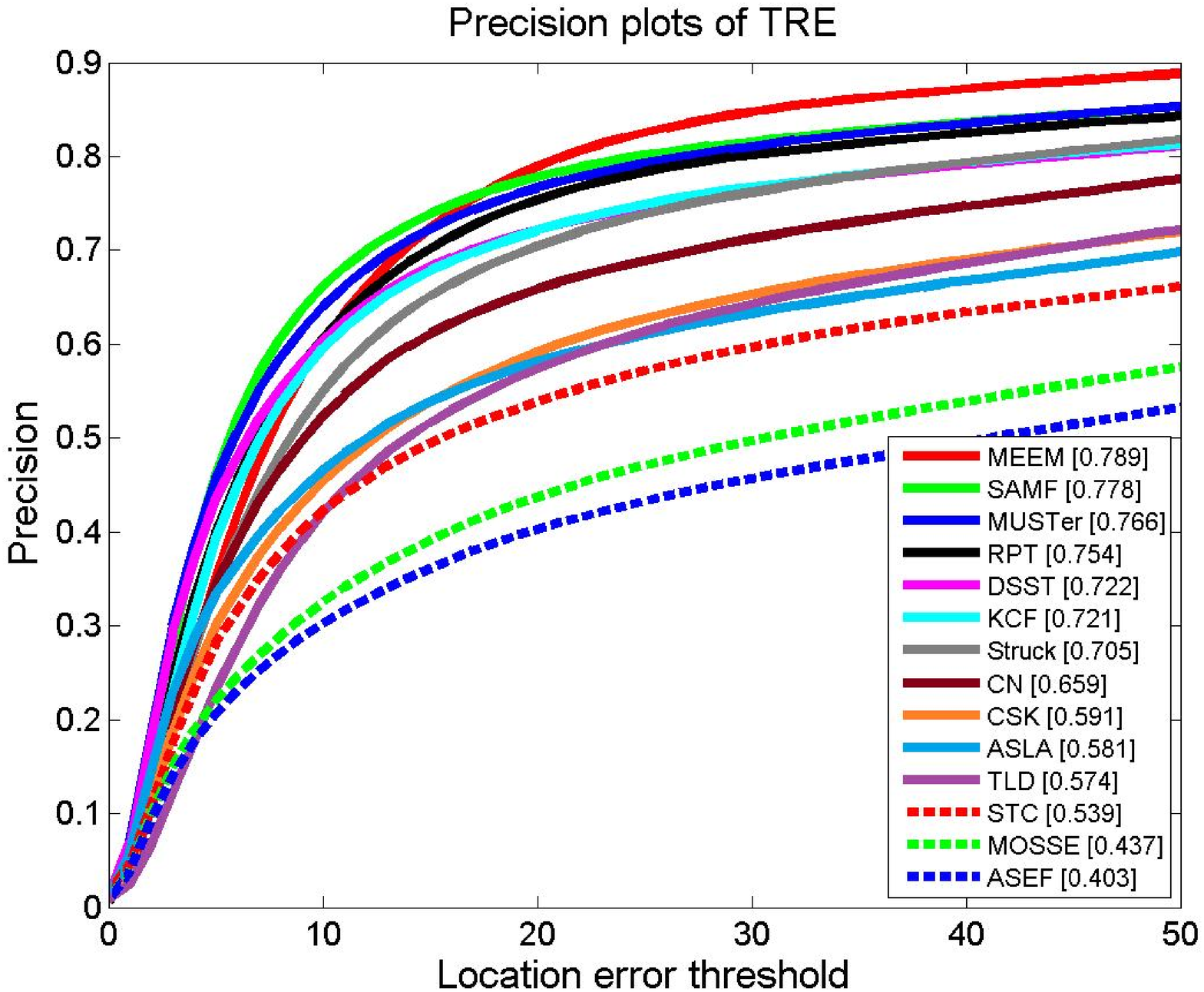}\\
   \includegraphics[width=0.3\linewidth]{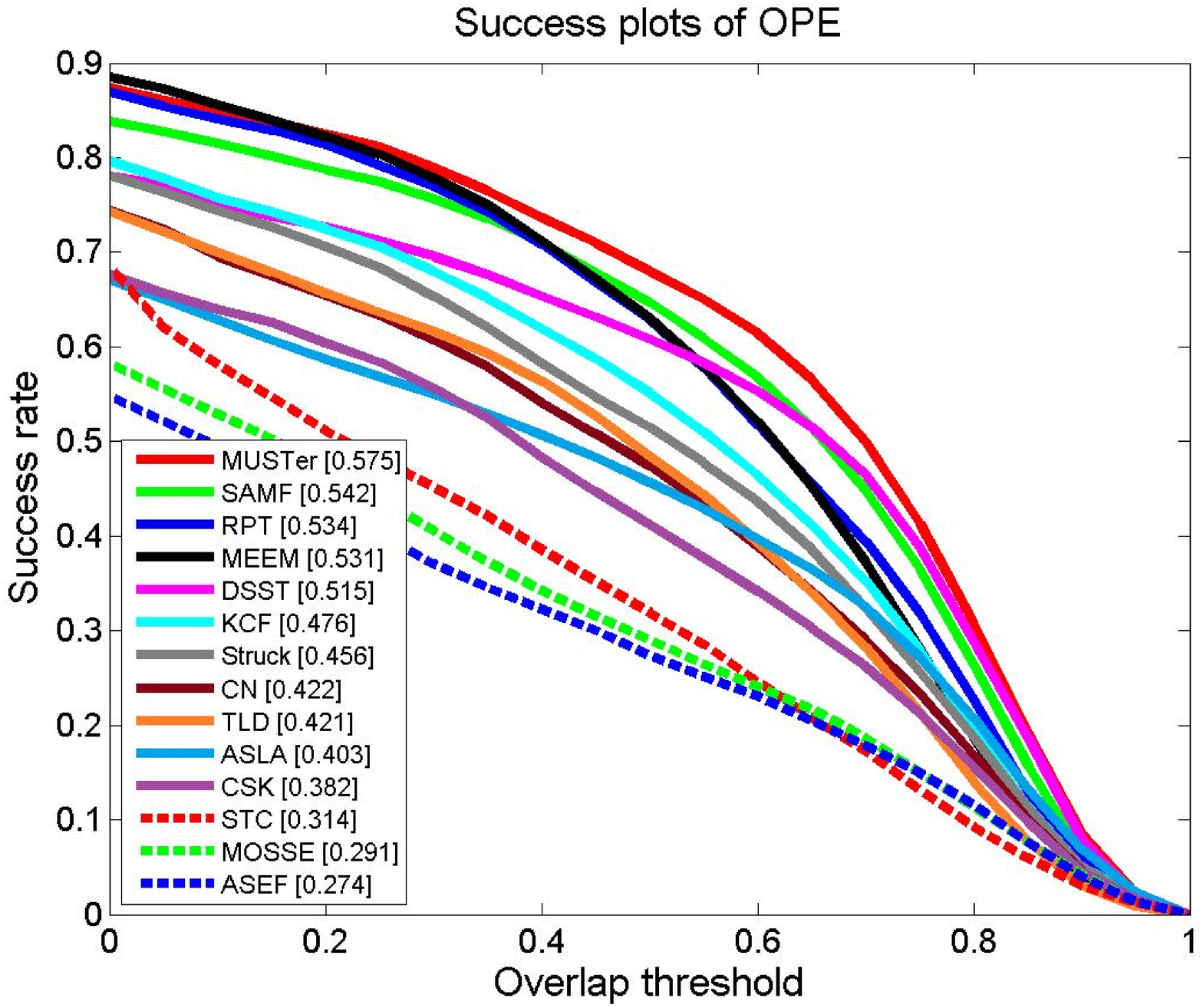} &
   \includegraphics[width=0.3\linewidth]{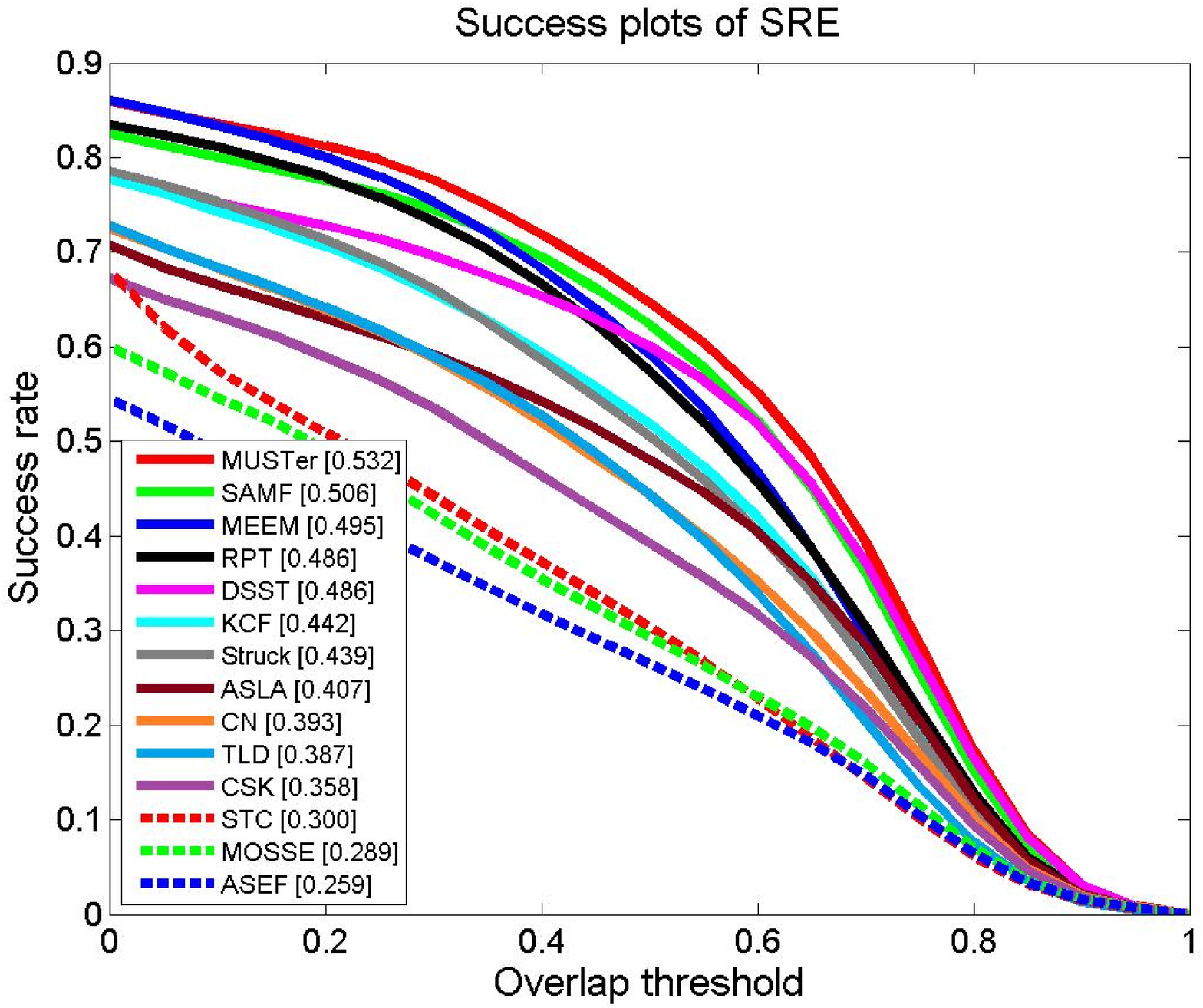} &
   \includegraphics[width=0.3\linewidth]{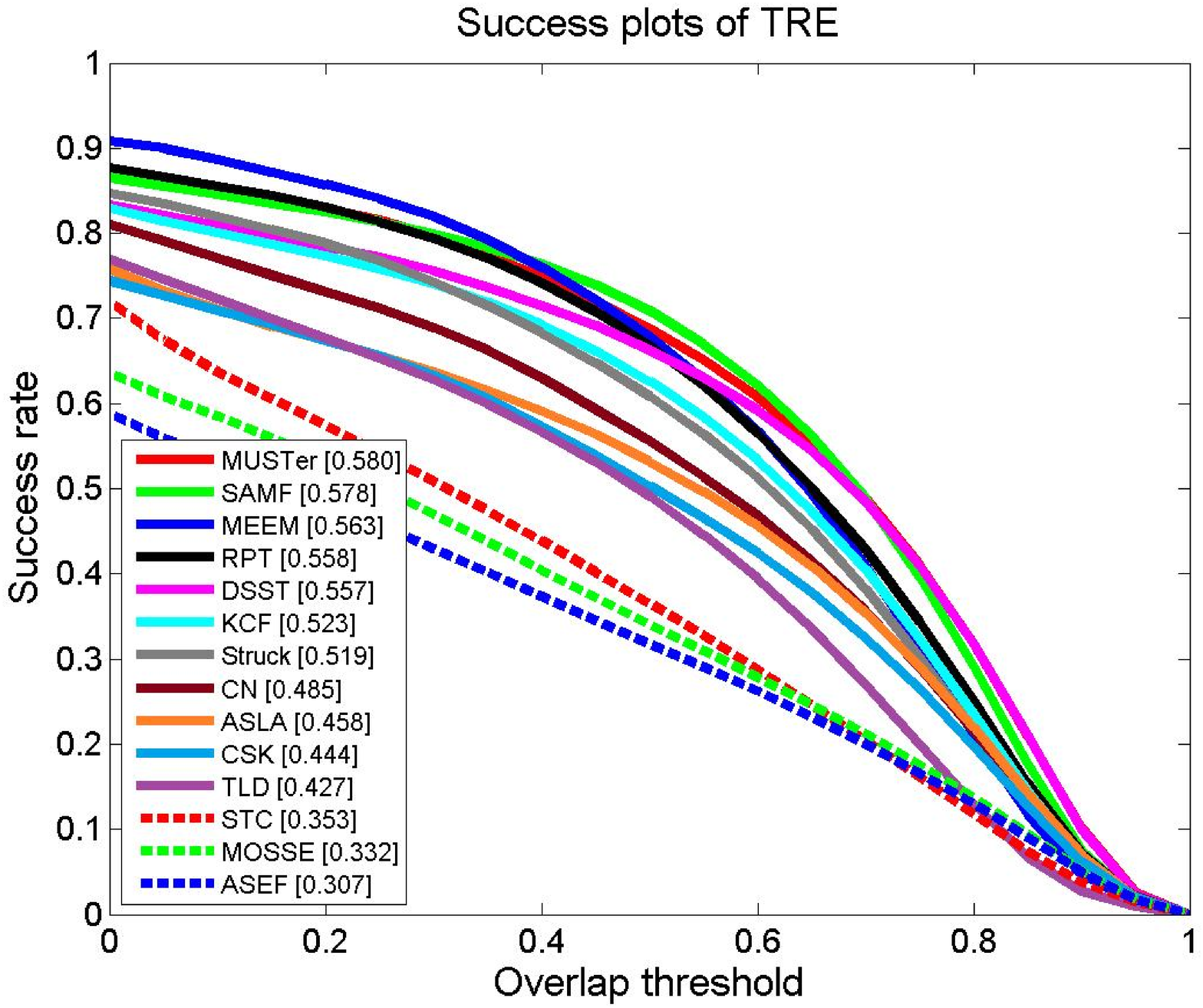}
   \end{tabular}
\end{center}
\caption{Extended results for 10 CFTs and 4 indicate trackers. Plots of OPE, SRE and TRE in OOTB100 are presented, as well as the scores and ranks of all the 14 trackers.
}
\label{fig:3}
\end{figure*}
\begin{table*}[!h]
    \begin{center}
       \caption{Additional results of OPE. Average center location errors (CLE), average overlap scores (OS) and average speeds (FPS) are presented. In each entry, score in the top comes from OOTB with 51 tasks, while the score in the bottom comes from OOTB100. The first and second best scores are highlighted by bold and underline.}
	  \label{tab:add}
	  \begin{tabularx}{\linewidth}{|X||*{14}{X<\centering|}}
	  \hline
	   Tracker &MUSTer &RPT &SAMF &MEEM &DSST &KCF &ASLA &Struck &TLD &CSK &MOSSE &STC &CN &ASEF \\
	  \hline
CLE &\textbf{17.3} \underline{31.7} &36.7 36.3 &28.9 35.9 &\underline{22.3} \textbf{28.8} &41.3 50.8 &35.5 45.3 & 73.1 75.5 &50.6 49.6 &48.1 54.9 &88.8 306 &82.8 99.6 &68.3 80.4 &64.8 81.8 &91.6 130 \\
	  \hline
OS(\%) & \textbf{65.0} \textbf{58.3} & \underline{58.2} 53.9 & 57.9 \underline{54.8} & 57.9 53.6 & 56.1 52.2 & 51.9 48.0 & 43.9 40.6 & 47.8 45.9 & 44.1 42.4 & 40.1 38.4 & 31.8 29.1 & 35.1 31.2 & 44.8 42.4 & 29.9 27.4\\
	  \hline
	   FPS&3.85 3.94&3.70 3.63&15.8 15.1&19.3 20.0 &32.7 31.5 &191 183 &7.48 4.75&20.4 20.7&33.3 40.9 &288 282 &281 284&\textbf{580} \textbf{578}&142 132&\underline{324} \underline{320}\\
	   \hline
	  \end{tabularx}
    \end{center}
\end{table*} 
\subsubsection{Evaluation Methods}
Following the protocol proposed in \cite{wu2013online}, One-Pass Evaluation (OPE), Temporal Robustness Evaluation (TRE) and Spatial Robustness Evaluation (SRE) are performed in our evaluation. OPE is a traditional evaluation method which runs trackers on each sequence for once. For TRE, it runs trackers on 20 sub-sequences segmented from the original sequence with different lengths, and SRE evaluates trackers by initializing them with slightly shifted or scaled ground truth bounding boxes. With TRE and SRE, the robustness of each evaluated trackers can be comprehensively interpreted.

After running the trackers, precision plots and success plots are applied to present results. Instead of using average Euclidean distances between the predicted centers to ground-truth centers, precision plots show percentages of frames whose estimated locations lie in a given threshold distance to ground-truth centers. Regarding to success plots, an overlap score is introduced to represent performance. Let $r_t$ denote the area of tracked bounding box and $r_a$ denote the ground truth. An Overlap Score (OS) can be defined by $S=\frac{|r_t\cap r_a |}{|r_t\cup r_a |}$  where $\cap$ and $\cap$ are the intersection and union of two regions, and $|\cdot|$ counts the number of pixels in the corresponding area.  Afterwards, a frame whose OS is larger than a threshold is termed as a successful frame, and the ratios of successful frames at the thresholds ranged from $0$ to $1$ are plotted in success plots.

Furthermore, trackers are ranked in both plots, and first 10 are presented. In precision plots, the precisions at the threshold of 20 pixels are used for ranking, while Area Under Curve (AUC) is used for ranking in success plots. Since AUS calculates overall performance, it is more representative for estimating the robustness of trackers. 

\subsection{Quantitative Evaluation}
\subsubsection{Overall Performance}

Using 51 sequences from OOTB, the overall performance of all the 39 trackers are obtained and presented in Figure \ref{fig:1}. 

According to the presented results, it can be found that CFTs, such as MUSTer, RPT and SAMF, perform considerably well in these plots. Particularly, the recent proposed tracker MUSTer has outperformed other trackers in all the success plots (64.1\% in OPE, 56.4\% in SRE and 61.7\% in TRE). In precision plots, MUSTer can also achieve state-of-art results despite that MEEM has a 1\% higher score in TRE.

Moreover, there are six improved CFTs that can always carry out top-10 performance, which indicates that these CFTs are considerably robust in tracking.

\subsubsection{Attribute-based Evaluation}

With the annotated attributes of each sequence, the performance of evaluated trackers with respect to different challenges is revealed. These involved challenges are mainly caused by three factors, which are varying appearance of the target, severe surrounding environments and the limitations of the cameras. The SRE results of success plots are presented in Figure \ref{fig:2}.   

The challenges brought by varying appearances of the target are scale variation, out-of-plane rotation, in-plane rotation, deformation, and fast motion. In scale variations, MUSTer, DSST and SAMF perform the best with around 50\% overlap score, which suggests that the searching strategy introduced in Section \ref{sec:scale} is effective. In the meantime, the score of RPT (48.8\%) proves that its scale estimation scheme is also applicable. For rotations and deformations, MUSTer is shown to be the most robust tracker, and RPT, SAMF, MEEM also perform well. While in the fast motion evaluation, the best score is carried out by MEEM (52.7\%), which is 2.4\% higher than the score of MUSTer. As a conclusion, the improved CFTs, which are MUSTer, RPT, DSST and SAMF, can learn a considerably robust appearance model, while MEEM is shown to be better at locating the fast moving targets.

Another group of challenges are caused by surrounding environments, including illumination variations, occlusions and background clutters. It can be found that CFTs still produce better results and can exclude other trackers from the first two ranks, which implies that background contexts can be efficiently identified to help avoid distractions using correlation filter-based tracking algorithms.

The rest of the evaluated attributes are motion blur, out of view and low resolution, which are mainly brought by the limitations of cameras. In motion blur and out of view challenges, the multi-expert restoration scheme has made MEEM the most robust among tested trackers, while MUSTer maintains the first rank in low resolution. It is worth mentioning that RPT has dropped a lot in low resolution tracking, which only achieves 34.3\% success rates and 6th rank in top 10. This may suggest that tracking with local appearance need high resolution for better performance.

To sum up, recent CFTs such as MUSTer, RPT, SAMF, DSST and KCF can all perform well in various challenging tasks, and the strengths of using correlation filters for tracking are shown to be significant. In particular, MUSTer becomes the most robust tracker in our evaluation since it wins 8 challenges out of 11 challenges.

\subsubsection{Extended Experiments}

For a better interpretation of tested CFTs, a larger OOTB with 100 sequences, which can be denoted as OOTB100, is used for extended experiments. The overall performance can be found in Figure \ref{fig:3}. To better represent the performance of CFTs, 4 other competing trackers are additionally selected as indicators. These selected trackers are Struck and ASLA, which use a single online classifier for tracking, and MEEM and TLD, which apply multiple online classifiers. 

According to the presented results, some aspects of successful improvements on CFTs can be revealed. First, the training scheme of MOSSE, which averages over all the samples, is shown to be better than training scheme of ASEF, which averages over learned filters. Second, spatial contexts used in STC are also shown to be beneficial. Furthermore, introducing kernel methods has made CSK and KCF much more competing, and has helped CFTs like MUSTer and SAMF become state-of-art trackers. On the other hand, color attributes used in CN tracker and HOG feature used in KCF are also shown to be advantageous. By integrating both features, SAMF further improves the overall performance. Moreover, improvements based on relieving the scaling issue, applying part-based strategy and introducing long-term tracking are proven to be effective as well.

In comparison with other four competing trackers, improved CFTs like MUSTer and SAMF still perform well in OOTB100, despite that the best results in precision plots of OPE and TRE are achieved by MEEM. This may because its restoration method can help MEEM quickly re-locate the target after drifting. 

Besides the plots, additional statistical data of trackers under OPE can be found in Table \ref{tab:add}, where average Center Location Error (CLE), average Overlap Score (OS)  and average speeds of the tested trackers are presented. According to the results, MUSTer and MEEM deliver the most precise results, and higher overlap scores are achieved by MUSTer, RPT and SAMF. By further observing the speeds of presented trackers, it can be found that better scores are often carried out by slower trackers, which suggests that acceleration is still an undergoing topic. 

\subsection{Qualitative Evaluation}
\begin{figure*}[!t]
\begin{center}
\begin{tabular}{@{\hspace{2mm}}c@{\hspace{1mm}}c@{\hspace{1mm}}c@{\hspace{1mm}}c@{\hspace{1mm}}c}
	\textit{Basketball} & \textit{Board} & \textit{Bolt} & \textit{Car1} & \textit{Car4} \\
   	\includegraphics[width=0.188\linewidth,height=0.10\textheight]{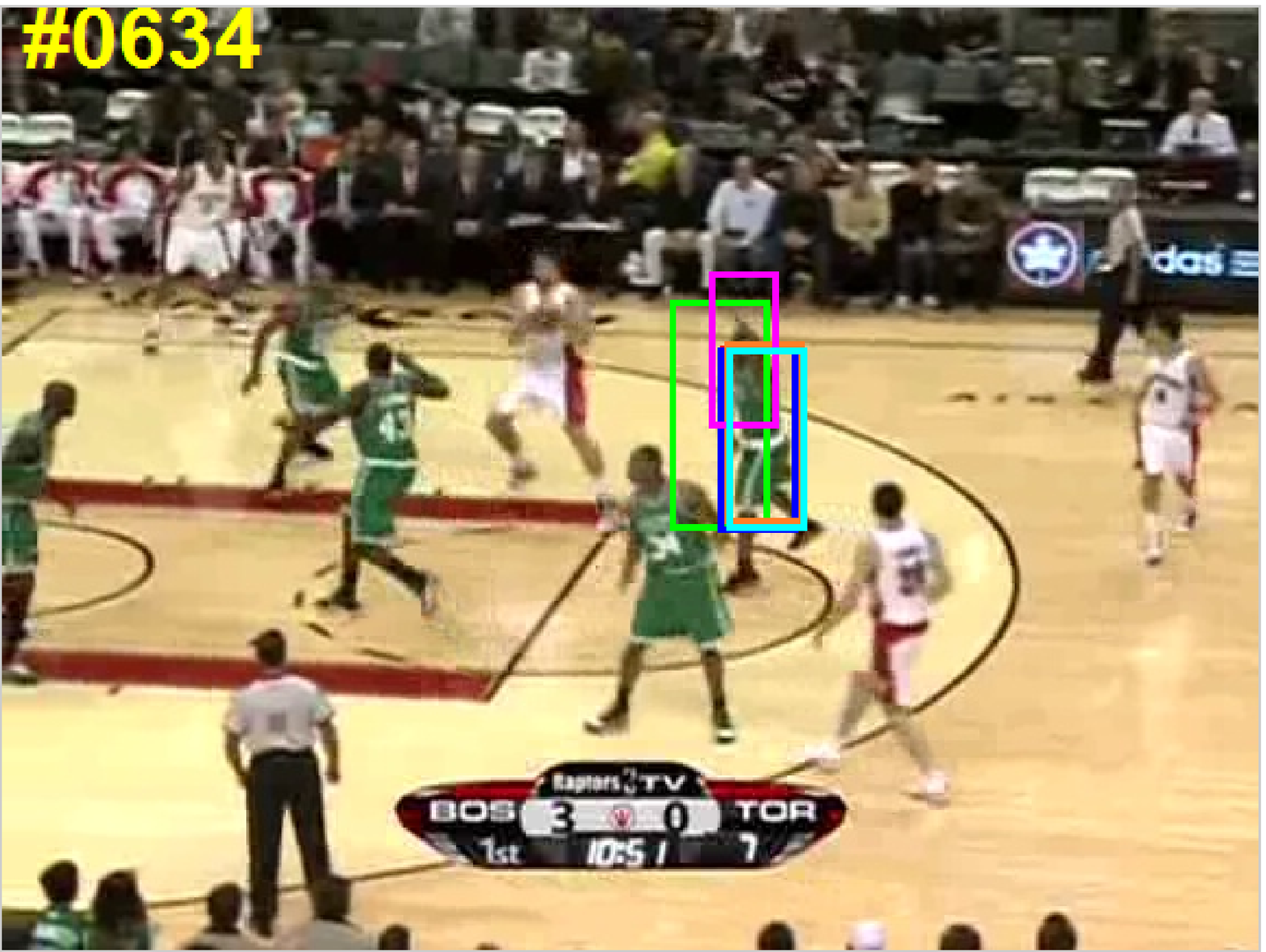}  &
	\includegraphics[width=0.188\linewidth,height=0.10\textheight]{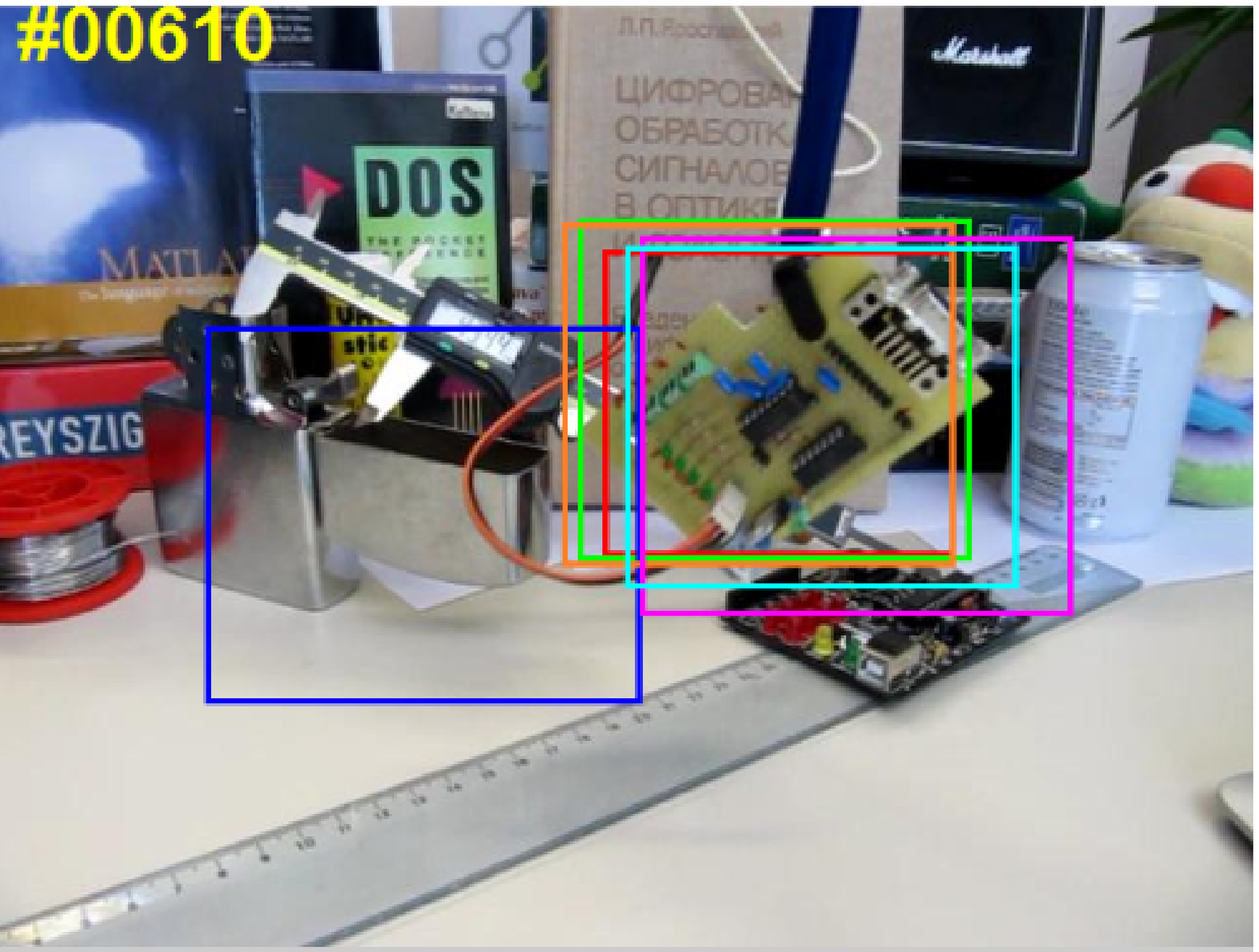}  &
	\includegraphics[width=0.188\linewidth,height=0.10\textheight]{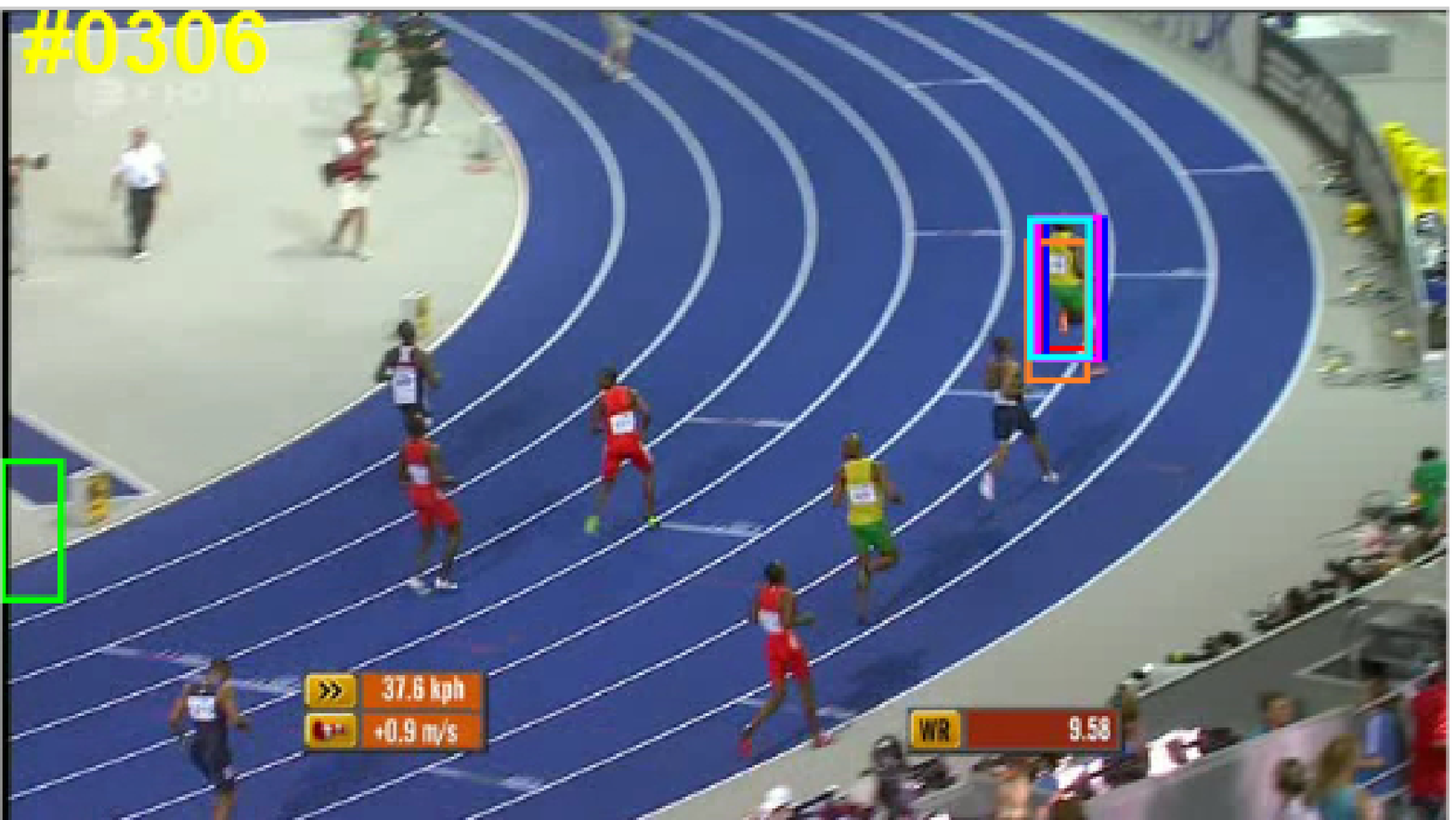}  &
	\includegraphics[width=0.188\linewidth,height=0.10\textheight]{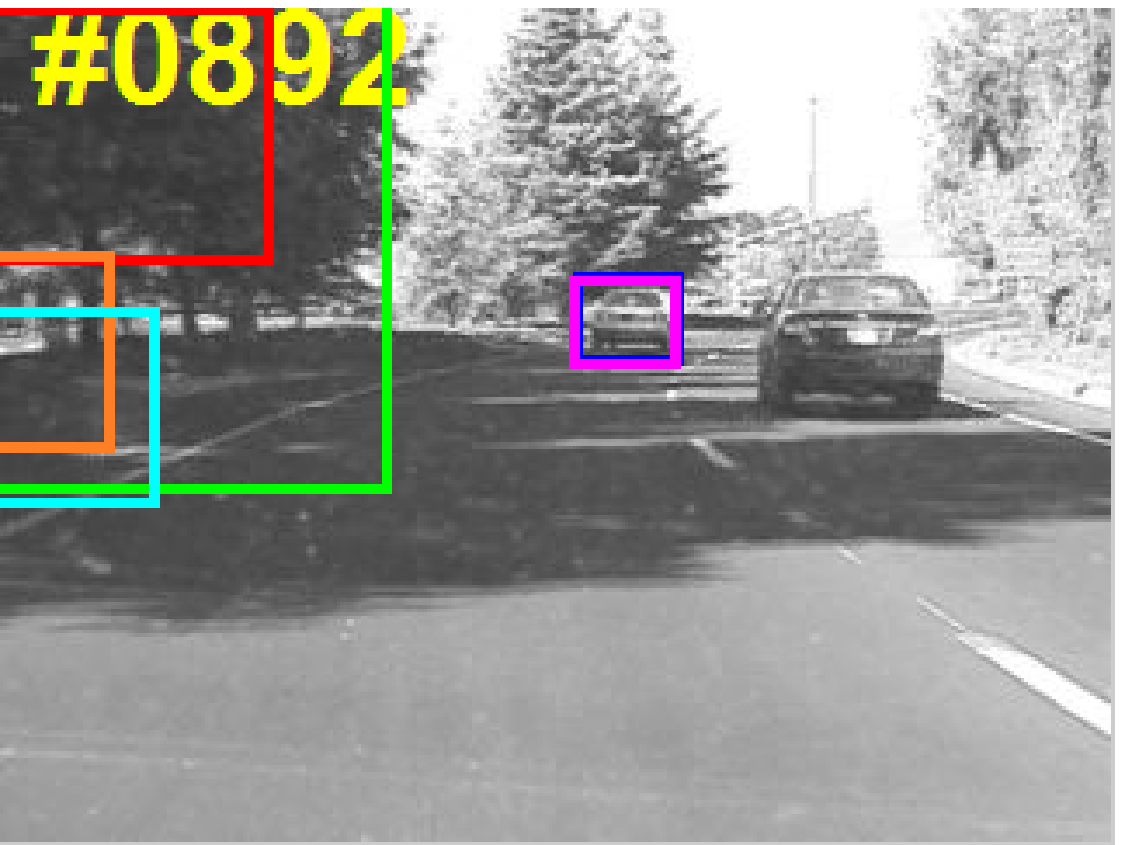}  &
	\includegraphics[width=0.188\linewidth,height=0.10\textheight]{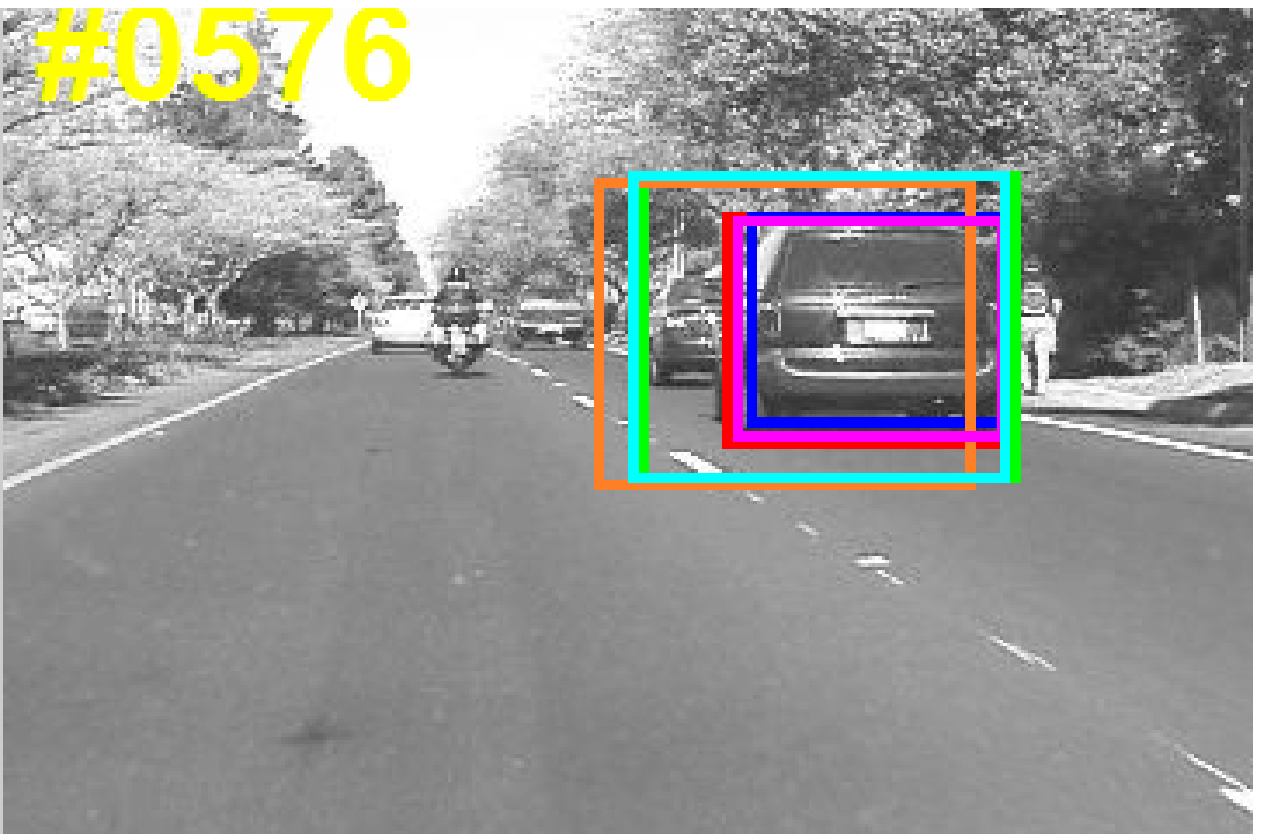}  \\
\\
	
	\textit{CarScale} & \textit{Couple} & \textit{David} & \textit{Deer} & \textit{FaceOcc2} \\
	\includegraphics[width=0.188\linewidth,height=0.10\textheight]{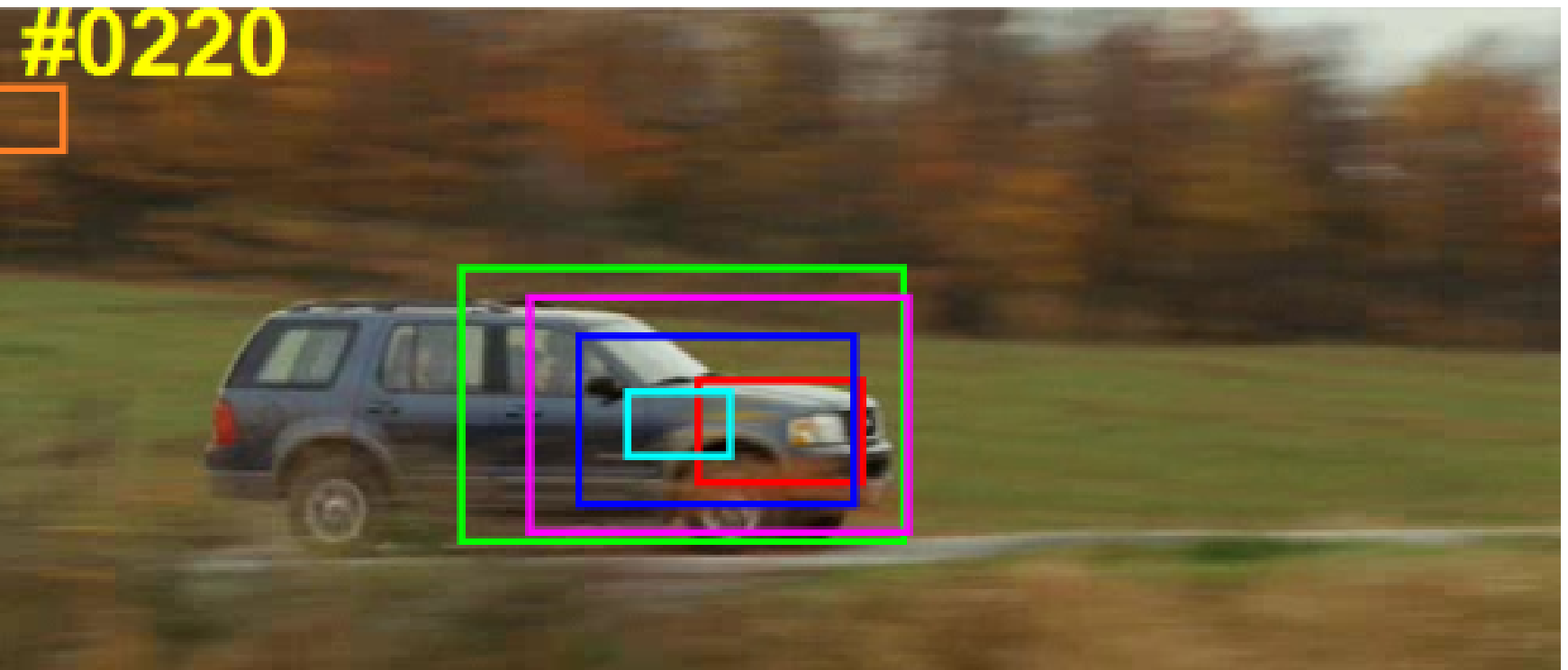}  &
	\includegraphics[width=0.188\linewidth,height=0.10\textheight]{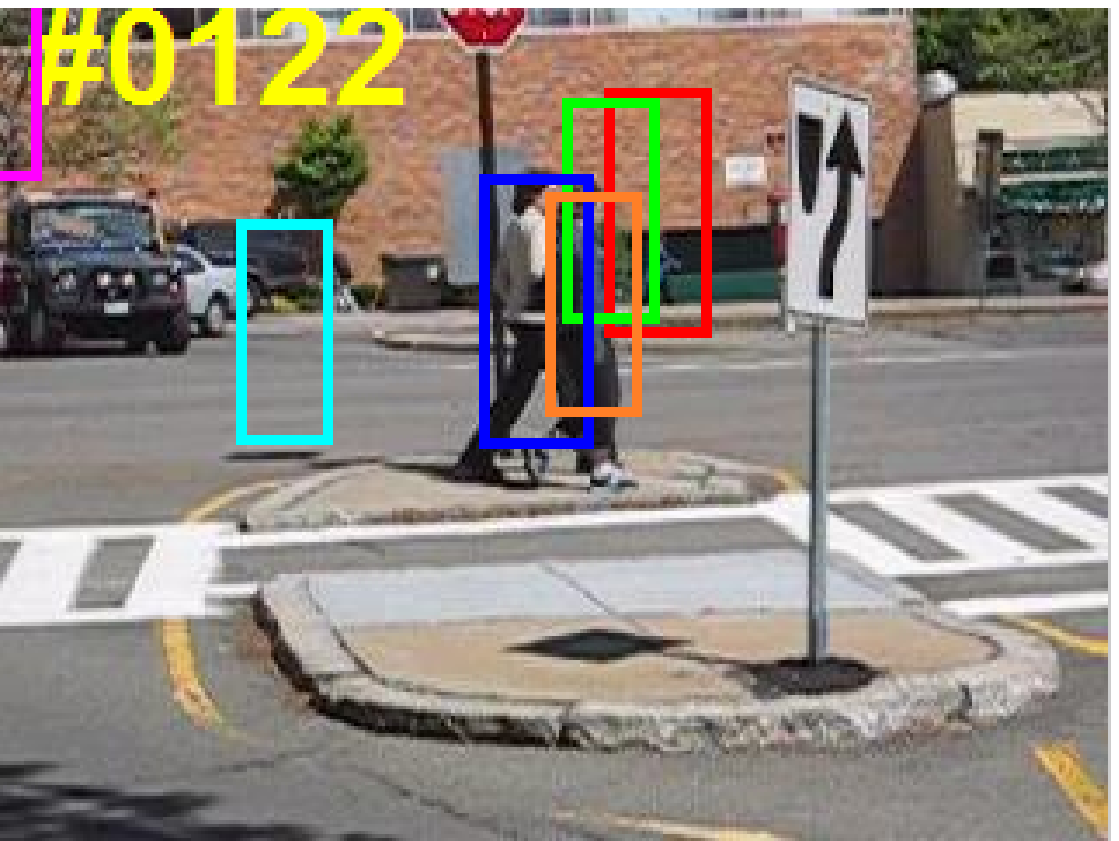}  &
	\includegraphics[width=0.188\linewidth,height=0.10\textheight]{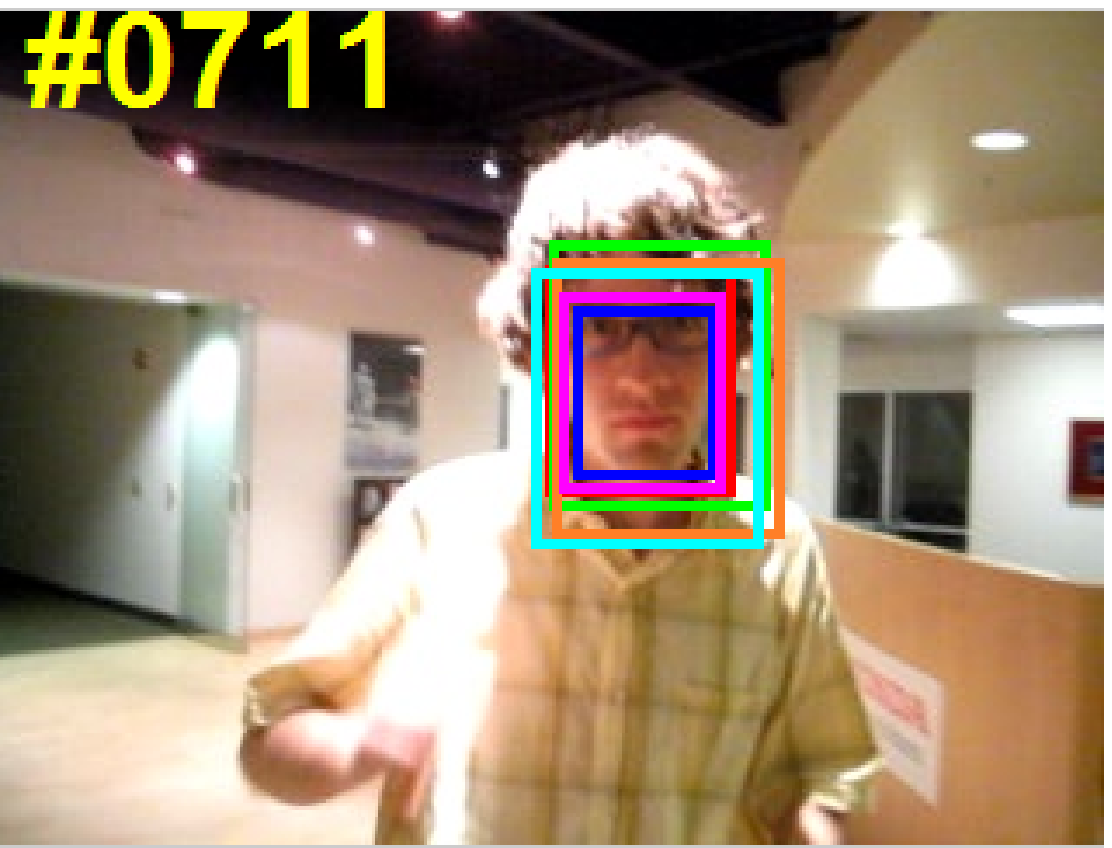}  &
	\includegraphics[width=0.188\linewidth,height=0.10\textheight]{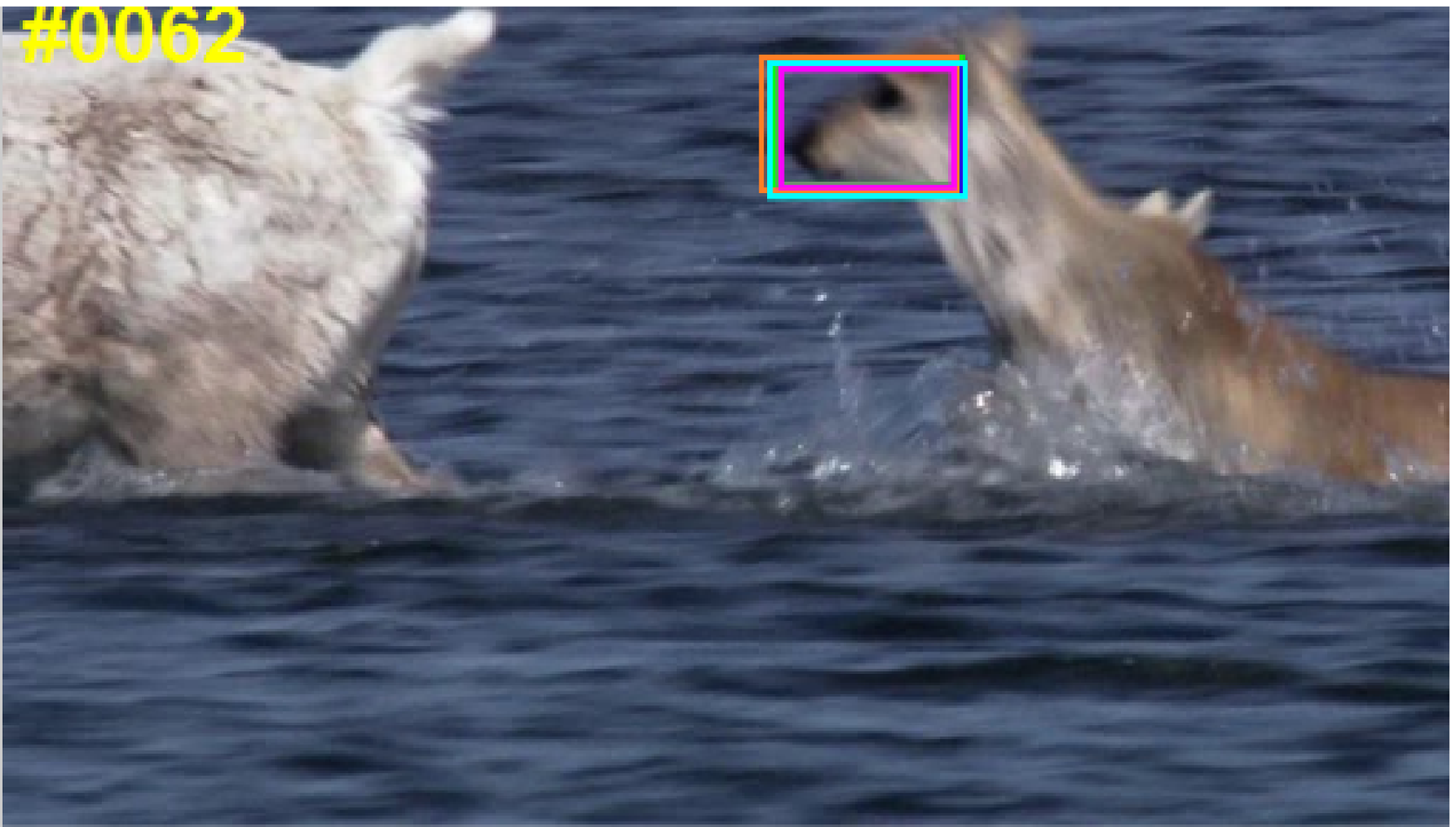}  &
	\includegraphics[width=0.188\linewidth,height=0.10\textheight]{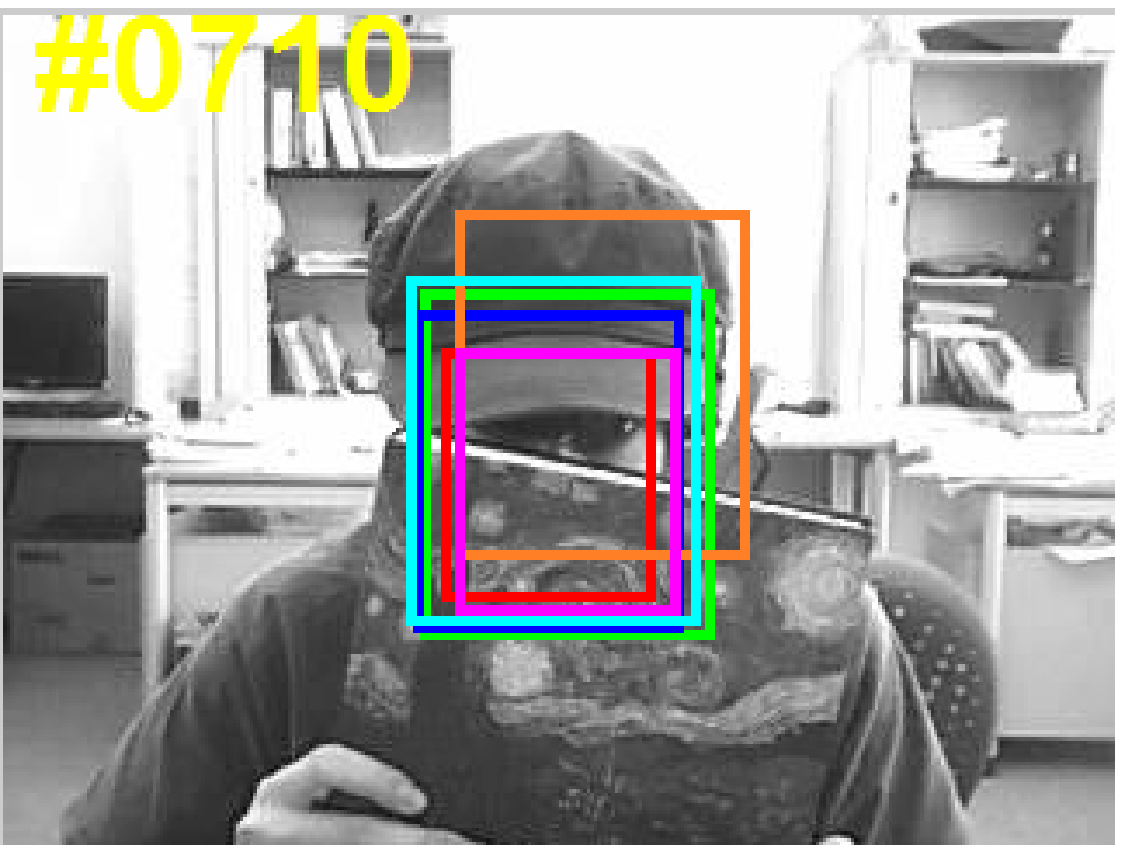}
\\
\\
	
	\textit{Freeman1} & \textit{Girl2} & \textit{Jogging-2} & \textit{Liquor} & \textit{MotorRolling} \\
	\includegraphics[width=0.188\linewidth,height=0.10\textheight]{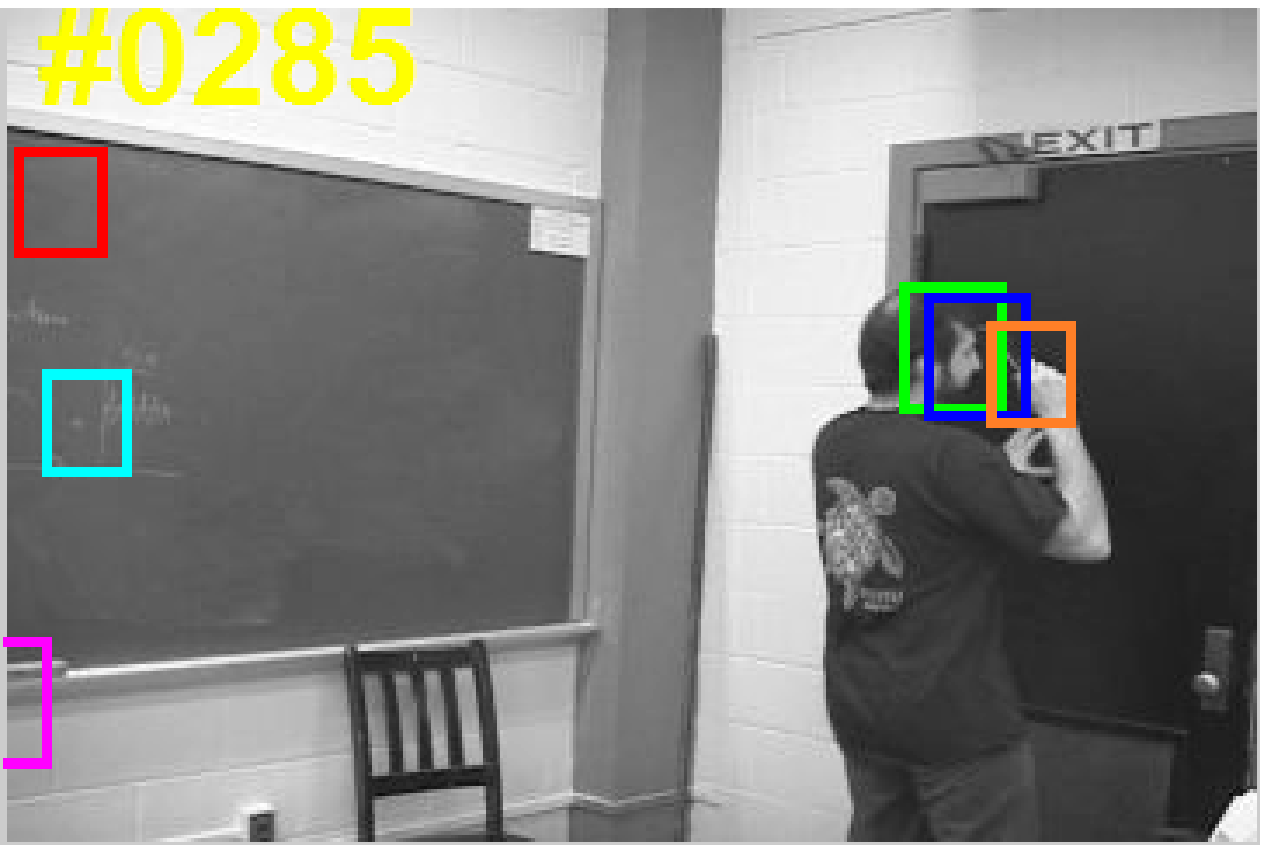}  &
	\includegraphics[width=0.188\linewidth,height=0.10\textheight]{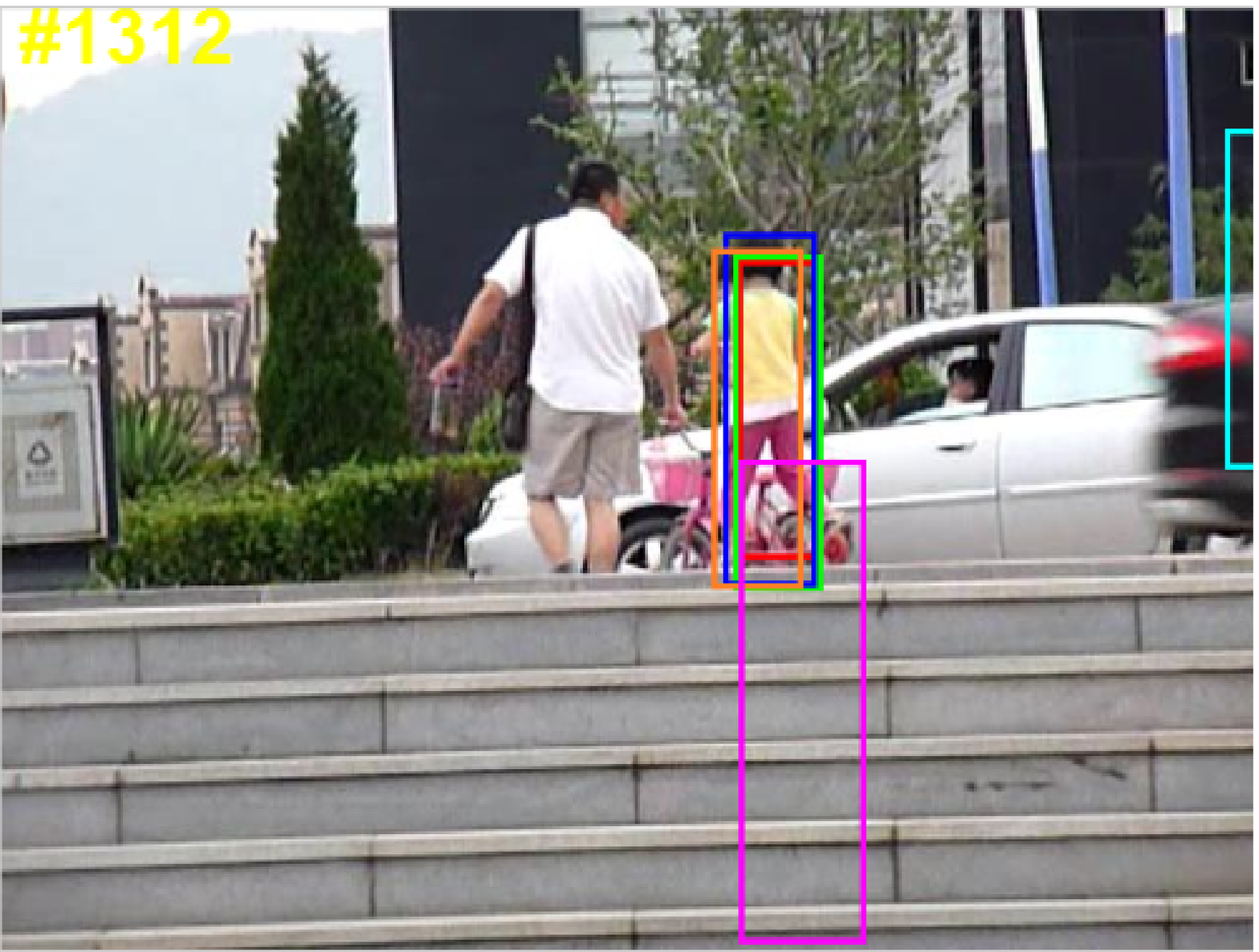}  &
	\includegraphics[width=0.188\linewidth,height=0.10\textheight]{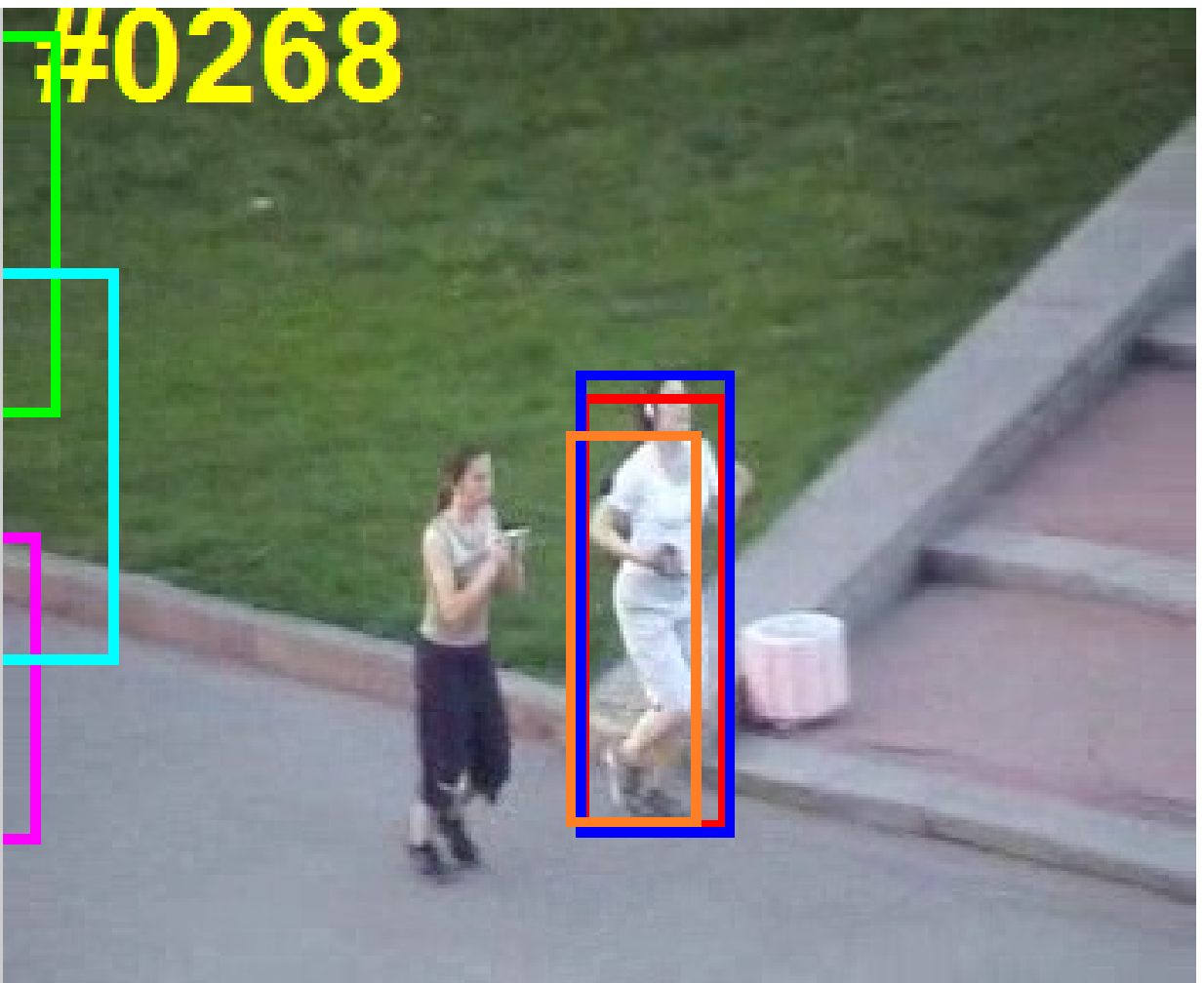}  &
	\includegraphics[width=0.188\linewidth,height=0.10\textheight]{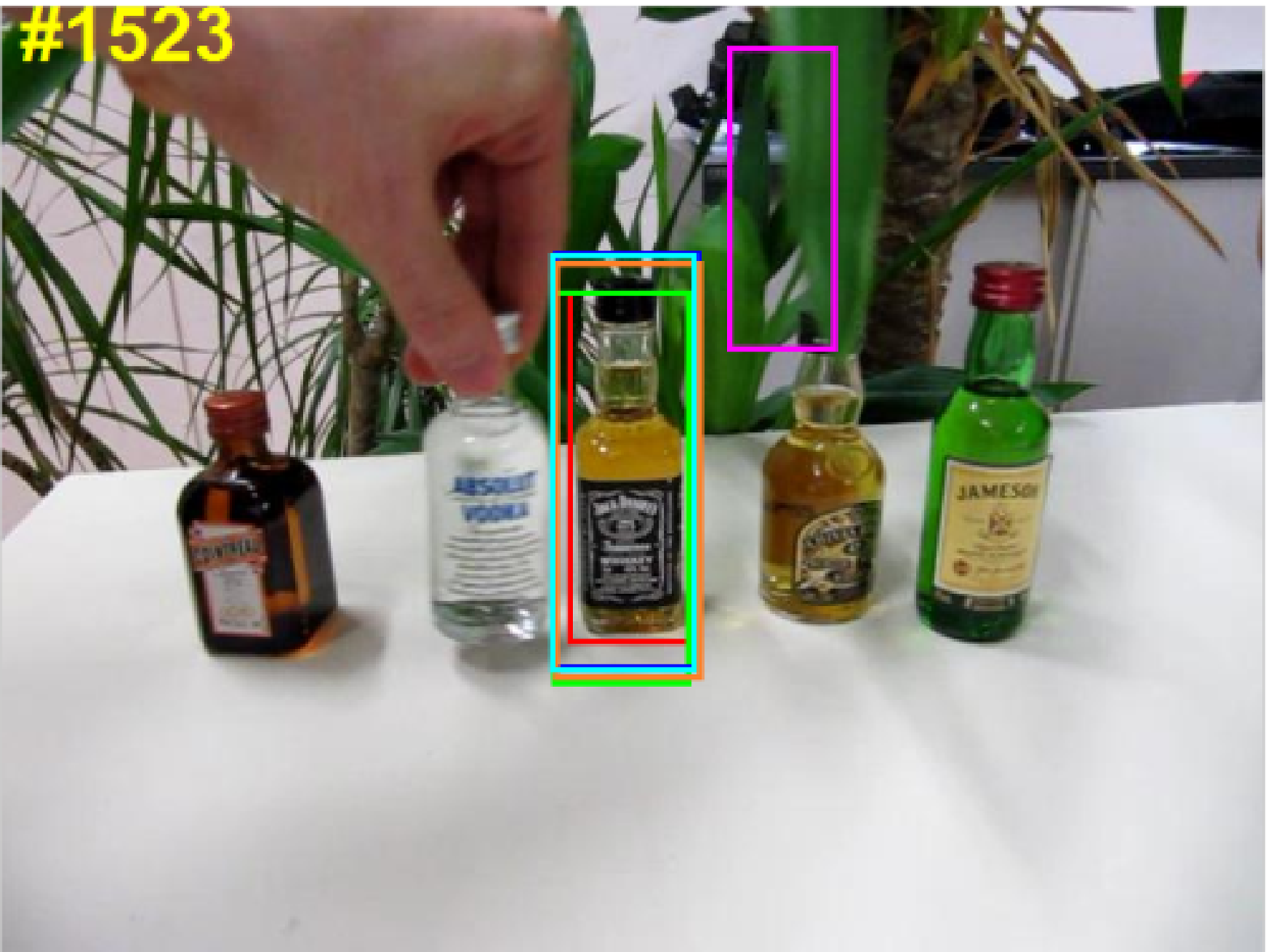}  &
	\includegraphics[width=0.188\linewidth,height=0.10\textheight]{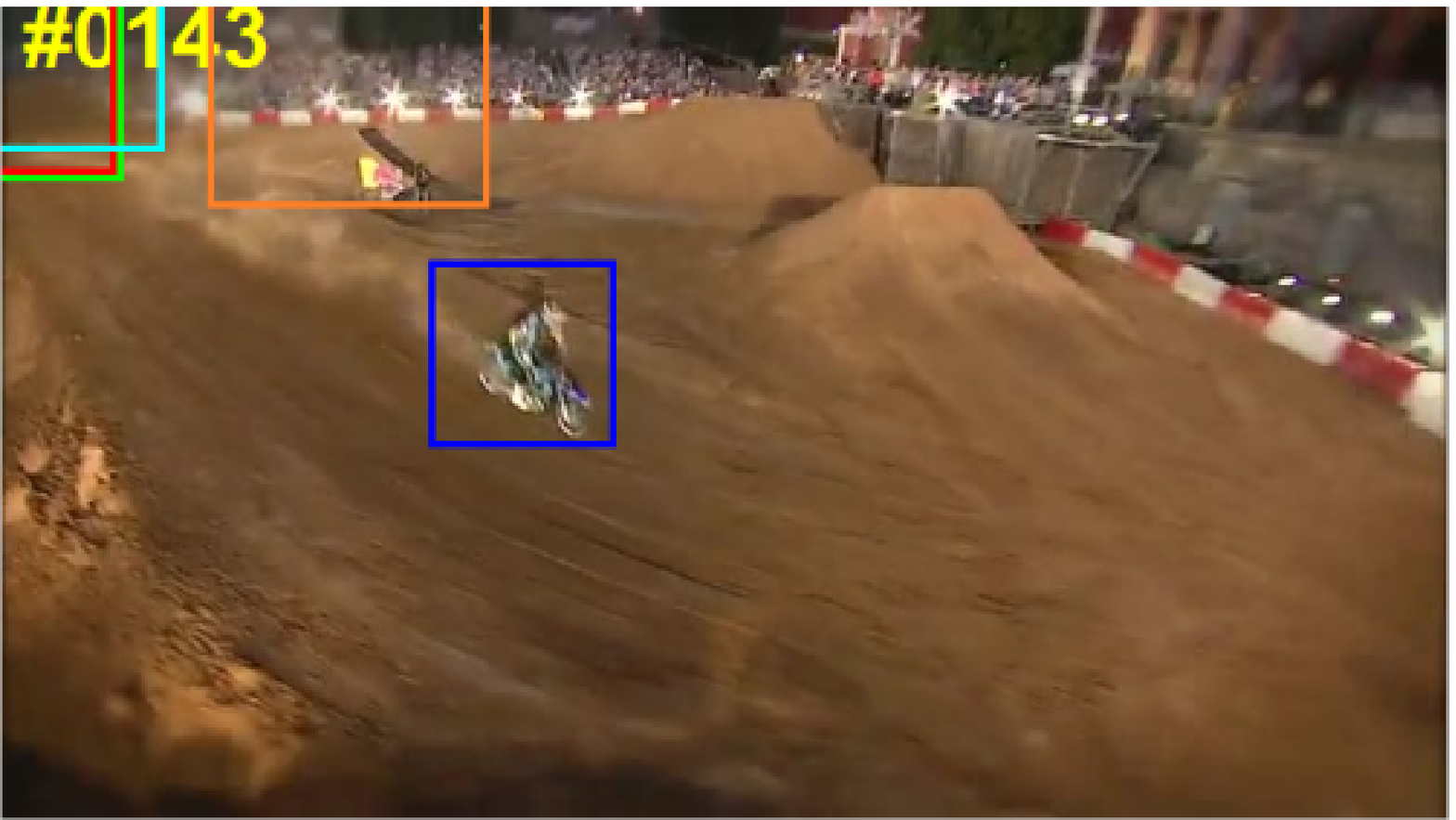}  
\\
\\

	\textit{Rubik} & \textit{Singer1} & \textit{Soccer} & \textit{Sylvester} & \textit{Walking2} \\
	\includegraphics[width=0.188\linewidth,height=0.10\textheight]{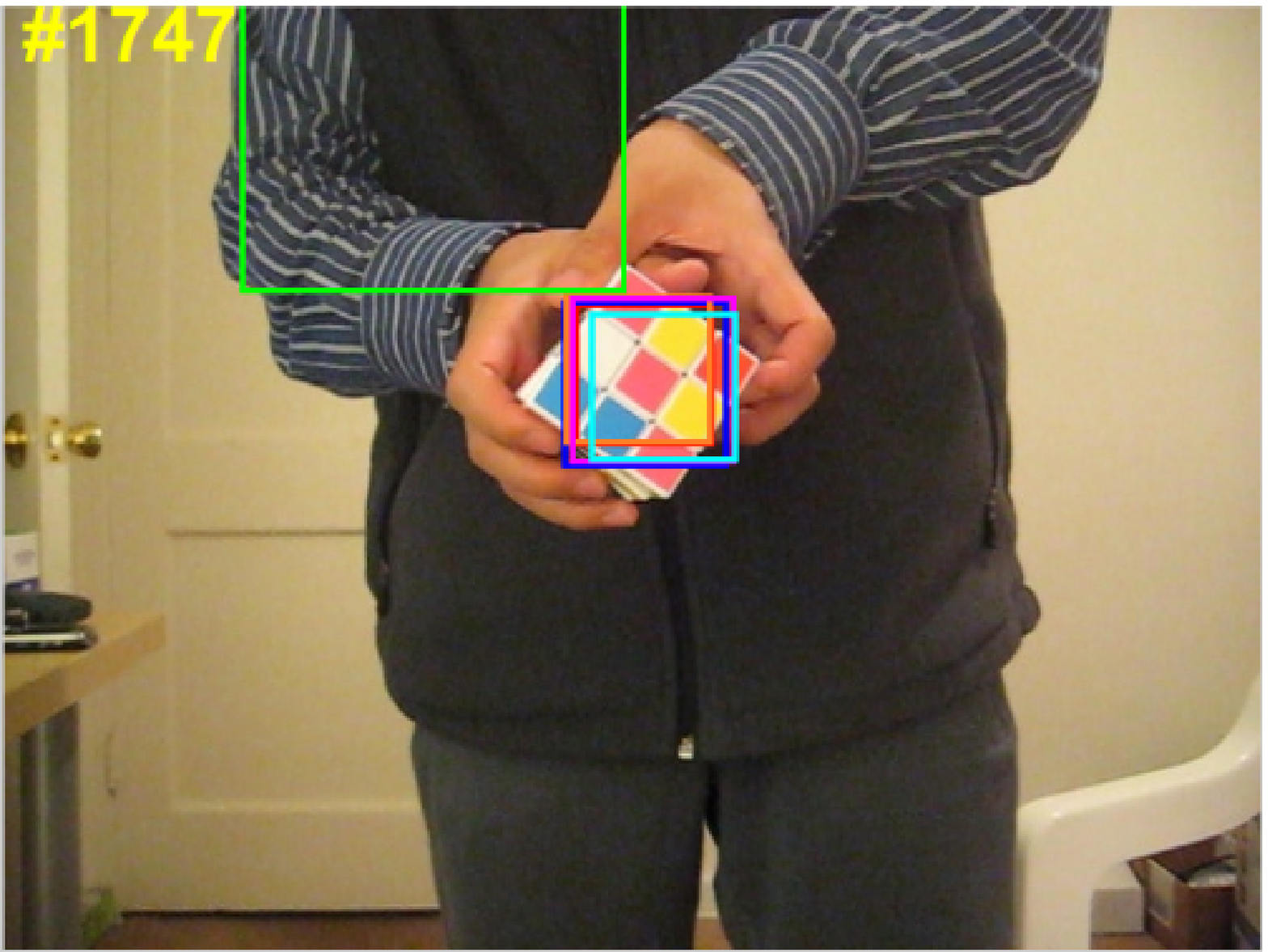}  &
	\includegraphics[width=0.188\linewidth,height=0.10\textheight]{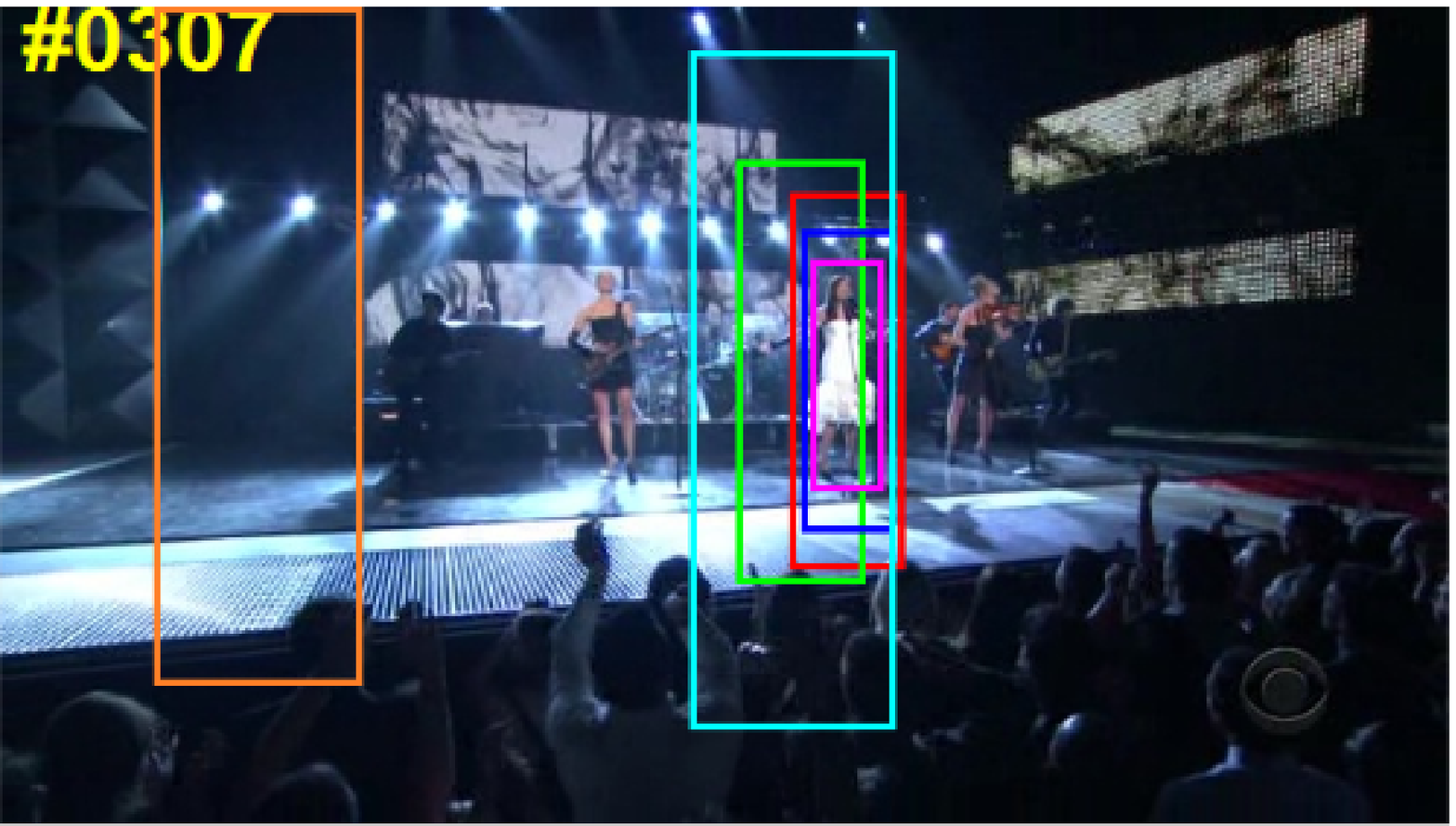}  &
	\includegraphics[width=0.188\linewidth,height=0.10\textheight]{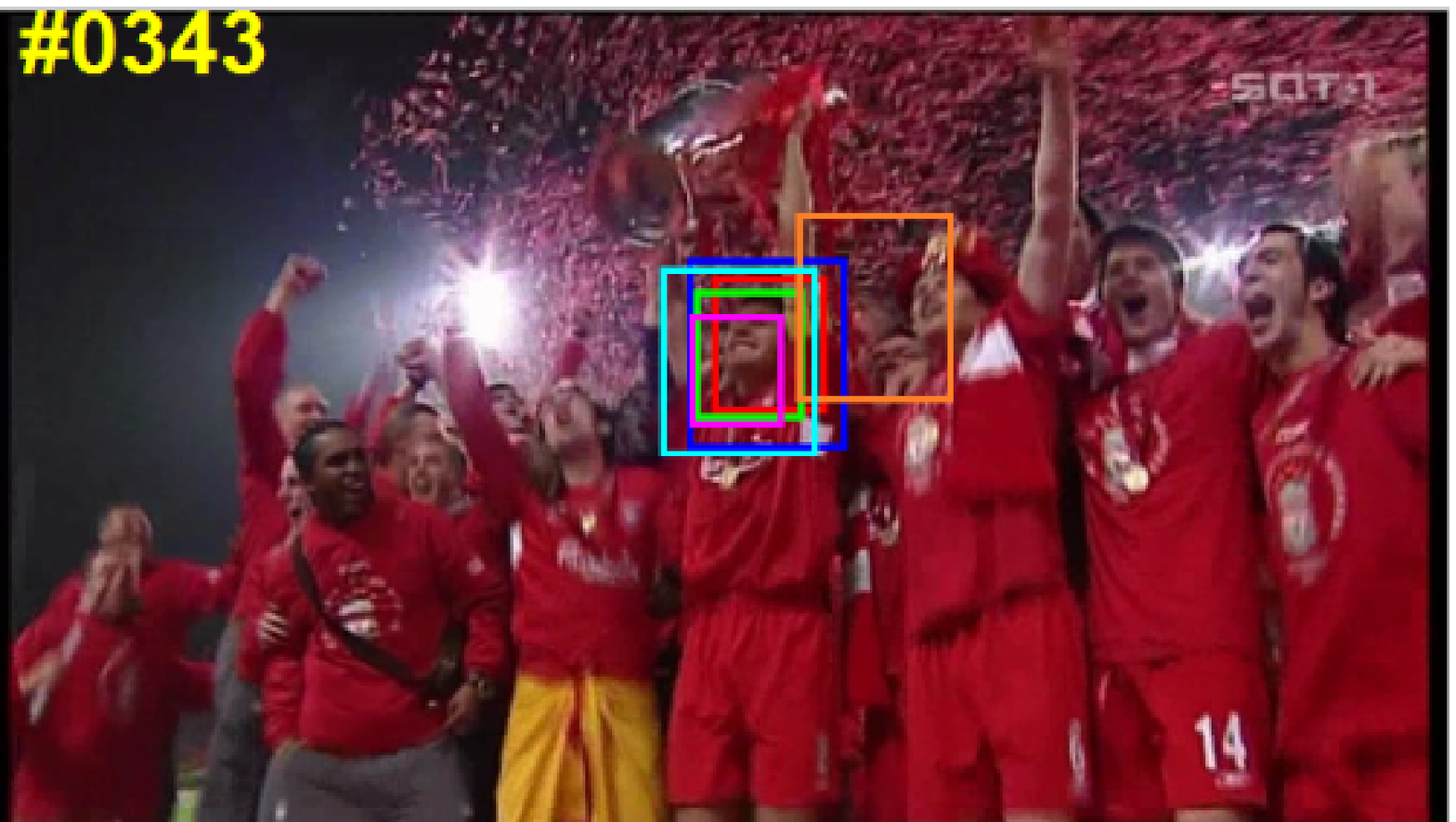}  &
	\includegraphics[width=0.188\linewidth,height=0.10\textheight]{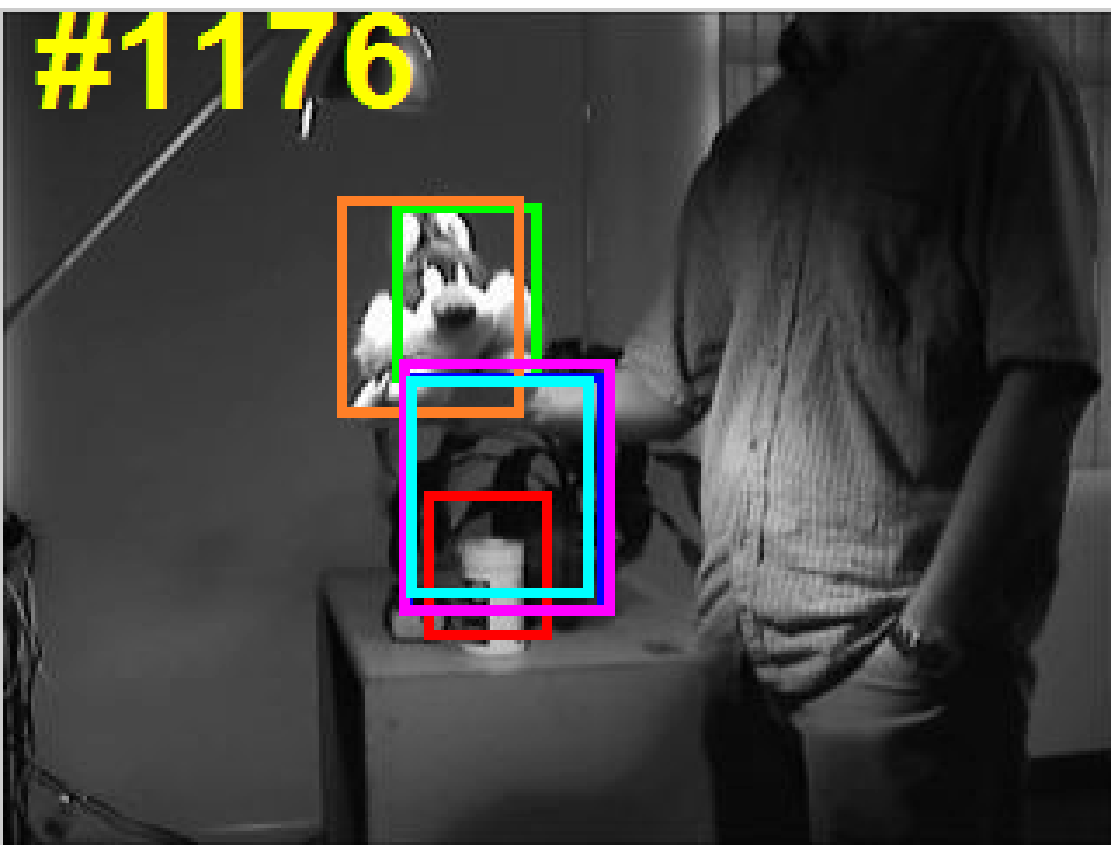}  &
	\includegraphics[width=0.188\linewidth,height=0.10\textheight]{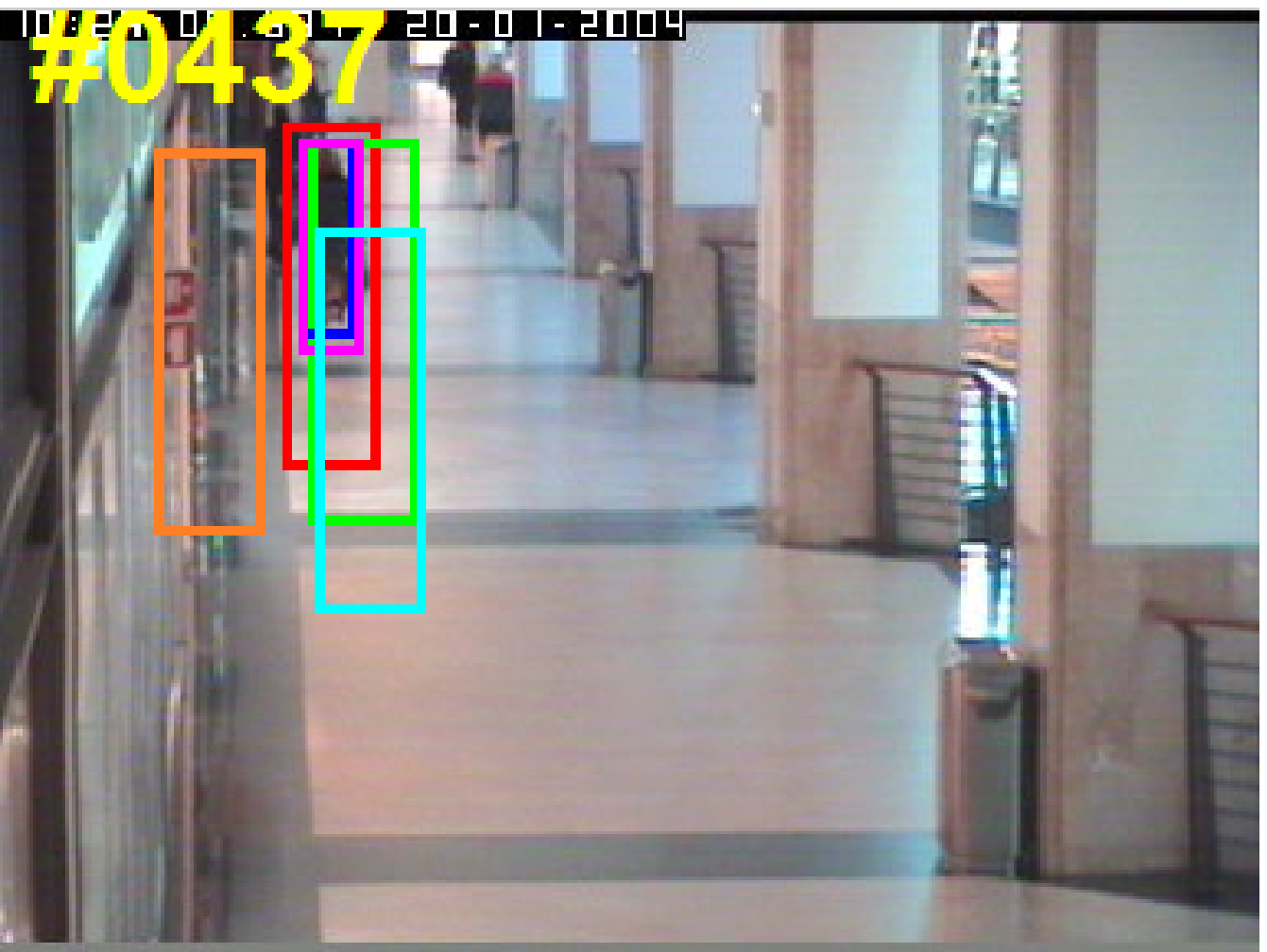} \\
\multicolumn{5}{c}{\includegraphics[width=0.6\linewidth]{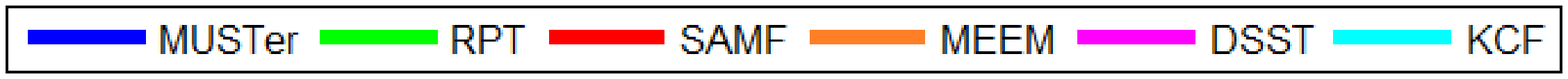}}
\end{tabular}
\end{center}
\caption{Qualitative results of 6 selected trackers in 20 sequences. The name of each sequence is located on top of the corresponding figure. Frames are collected from seven-eighth of the sequences.}
\label{fig:qualitative} 
\end{figure*}

To evaluate the actual tracking results, we have randomly selected 20 video sequences from OOTB-100 and used them for qualitative evaluation. These videos include \textit{Couple, CarScale, Bolt, Car4, Liquor, Board, Walking2, Singer1, Soccer, Car1, Girl2, Sylvester, MotorRolling, Jogging2, Rubik, Freeman1, FaceOcc2, Deer, Basketball}. In these videos, all the challenge attributes are properly included. Top-6 trackers in OOTB100, which are MUSTer, RPT, SAMF, MEEM, DSST and KCF, are selected for comparison, whose predicted boxes at the seven-eighth of the tested sequences are presented in Figure \ref{fig:qualitative}.

For KCF tracker, its limitation of using fixed windows can be revealed in \textit{CarScale} and \textit{Singer1}, where the predicted boxes are obviously incompatible with the target.  By relieving the scaling issue, DSST and SAMF track better in sequences like \textit{Car4} and \textit{CarScale}. However, lacking of long-term component has made them unable to re-locate missed targets, such as the failures in \textit{Freeman1}. With long-term consideration, the results of MUSTer are much better. For example, while other tested trackers lose the target in \textit{MotorRolling}, MUSTer is able to find the target and restart tracking. The only failed sequence for MUSTer is the \textit{Board}, in which background objects are quite similar to the target. In addition to MUSTer, MEEM also has the ability to re-locate the target, but its performance is limited when dealing with light variations regarding to the failure in \textit{Singer1}. For RPT, as discussed in the attribute-based evaluations, its tracking results become worse if the video quality is rather low, and RPT fails in \textit{Bolt} from the very beginning, which suggests that the initialization may be crucial for RPT.

Overall, although the qualitative performance of evaluated trackers is promising, there is still plenty of room to improve the robustness of CFTs as well as other trackers. The implementation of MUSTer may provide a promising direction for further research.

\section{Conclusion}
\label{sec:conclusion}

In this paper, we have reviewed numerous correlation filter-based tracking algorithms, and have conducted comprehensive experiments have been conducted. According to the experimental results, the efficiency and robustness of tracking with correlation filters have been verified. In specific, state-of-art performance can be achieved by improved CFTs like MUSTer, RPT, SAMF and DSST. Furthermore, MUSTer has shown to be the most robust tracker in the experiments. 

To further improve CFTs, some valuable points can be concluded based on the obtained results. First, it can be found that the possibilities of drifting are significantly reduced with powerful features, as shown by improved performance of CN and KCF. It is worthy of trying more types of features. Next, solving the scaling issue is also a vital direction for improving CFTs. Although experimental results have shown that both scaling pool and part-based methods are helpful for handling scale variations, approaches with less processing time are favorable. Regarding to the part-based tracking strategy, its advantages have not been fully exploited in RPT according to its performance. A more effective mechanism to fuse tracking results of different parts is desired. Lastly, introducing efficient long-term tracking methods is another promising direction. With the outstanding performance brought by MUSTer, long-term tracking is shown to be the most helpful complement of correlation filter-based tracking. Further studies can be made on improving the accuracy of long-term tracking component, developing more appropriate architectures and accelerating the overall system.

\ifCLASSOPTIONcaptionsoff
  \newpage
\fi

\bibliographystyle{IEEEtran}
\bibliography{IEEEabrv,Ref}

\end{document}